%% file: Thesis.tex
\documentclass[10pt,a4paper,twoside]{report}

\input{./meta/imports.tex}

\input{./meta/commands.tex}

\input{./meta/hardware.tex}
\author{Maciej Sakowicz}
\title{Detailed saliency maps generation using model-agnostic methods}

\begin{document}

\includepdf{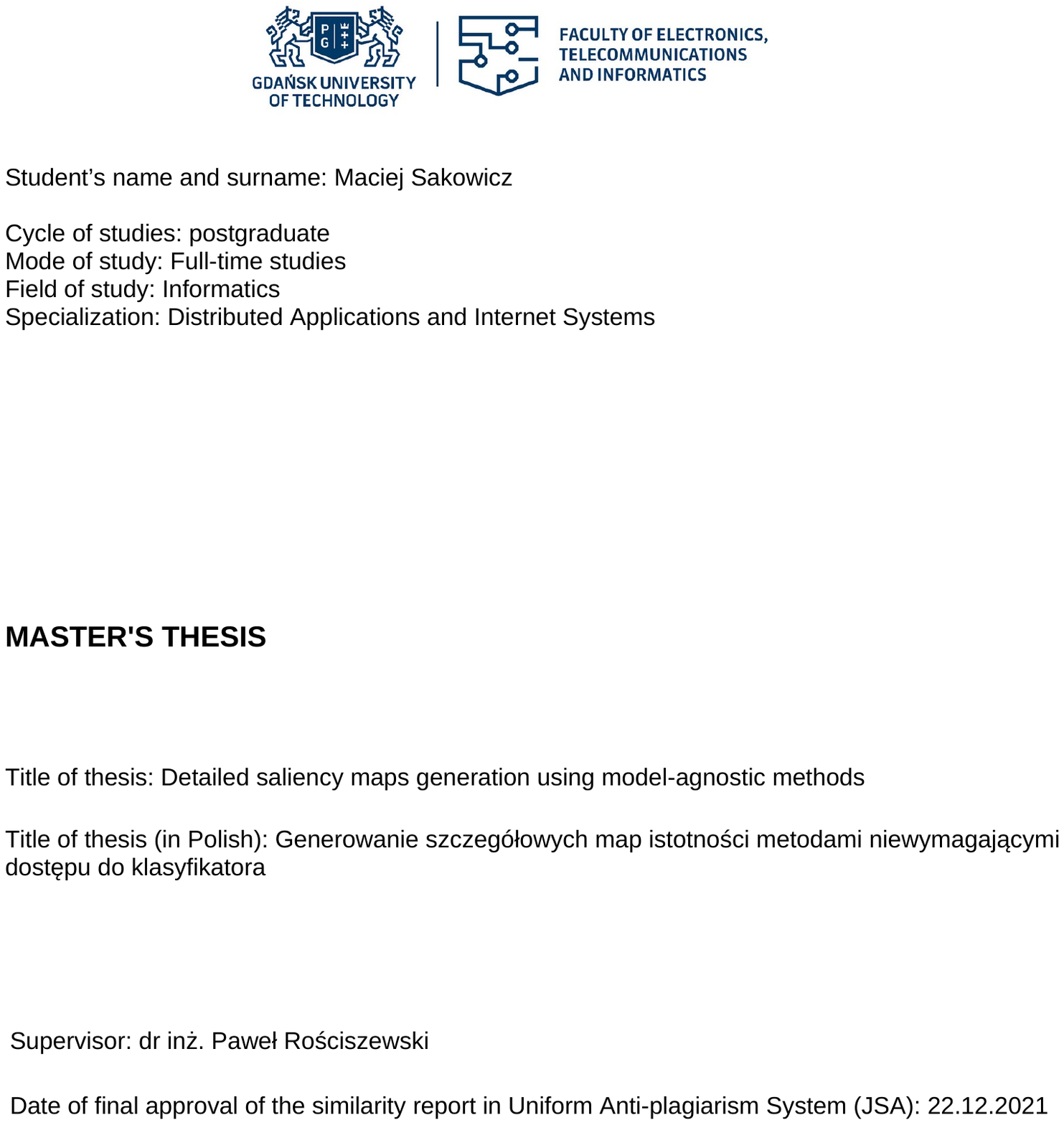}

\chapter*{Abstract}
\input{./tex/fluff/abstract.tex}

\singlespacing
\tableofcontents
\onehalfspacing

\input{./meta/symbols.tex}

\chapter{Introduction}
\label{ch:introduction}
\input{./tex/fluff/introduction.tex}

\chapter{Related Work}
\label{ch:related_work}
\input{./tex/fluff/related_work.tex}

\chapter{Proposed solution}
\label{ch:solution}
\input{./tex/solution/_intro.tex}
\input{./tex/solution/vrise_description.tex}
\input{./tex/improvements/guaranteed_grid_gens.tex}

\chapter{Methodology}
\label{ch:methodology}
\input{./tex/methodology/vrise_evaluation.tex}

\input{./tex/methodology/blur_sigma_selection.tex}

\input{./tex/methodology/fp16.tex}

\chapter{Evaluation}
\label{ch:evaluation}
\input{./tex/evaluation/_intro.tex}
\input{./tex/evaluation/grid_search.tex}

\input{./tex/evaluation/targeted_s.tex}

\input{./tex/evaluation/imgnet.tex}

\input{./tex/evaluation/ssim_eval.tex}
  \section{Observed side-effects}
  \label{sec:side_effects}
  \input{tex/theory/feature_slicing.tex}
  \input{tex/theory/saliency_misattribution.tex}

\input{./tex/evaluation/_outro.tex}

\chapter{Summary}
\label{ch:summary}
\input{./tex/fluff/summary.tex}

\input{./tex/fluff/future_work.tex}

\input{./tex/fluff/acknowledgements.tex}

\singlespacing
\bibliography{bib/bibliography}
\bibliographystyle{bib/IEEEtranN}
\addcontentsline{toc}{chapter}{Bibliography}
\listoffigures
\addcontentsline{toc}{chapter}{List of Figures}
\listoftables
\addcontentsline{toc}{chapter}{List of Tables}
\onehalfspacing

\begin{appendices}
\input{./appendix/B_gridsearch_map_samples.tex}
\input{./appendix/C_grid_search_heatmaps.tex}
\input{./appendix/D_grid_search_lineplots.tex}

\end{appendices}

\end{document}

%% file: meta/imports.tex
\usepackage{gutthesis}
\usepackage{tikz}
\usetikzlibrary{positioning}
\usetikzlibrary{spy}
\usepackage{siunitx}
\usepackage{mathtools}
\usepackage{amsmath}
\usepackage{subfig}
\usepackage{placeins}
\usepackage{pdflscape}
\usepackage{lscape}

\usepackage{multirow}
\usepackage{booktabs}

\usepackage{float}

\usepackage{pgfplots}
\DeclareUnicodeCharacter{2212}{−}
\usepgfplotslibrary{groupplots,dateplot}
\usetikzlibrary{patterns,shapes.arrows}
\pgfplotsset{compat=newest}
\usepackage{adjustbox}

%% file: meta/commands.tex
\newcommand\figref{Figure~\ref}
\newcommand\chref{Chapter~\ref}

\newcommand\tblref{Table~\ref}

%% file: tex/fluff/abstract.tex
The emerging field of Explainable Artificial Intelligence focuses on researching methods of explaining the decision making processes of complex machine learning models.
In the field of explainability for Computer Vision, explanations are provided as saliency maps, which visualize the importance of individual pixels of the input to the model's prediction.

In this work we focus on a perturbation-based, model-agnostic explainability method called RISE and propose two modifications - replacement of square occlusions with polygonal occlusions based on cells of a Voronoi mesh, and addition of an informativeness guarantee to the occlusion mask generator. These modifications, collectively called VRISE (Voronoi-RISE), are meant to, respectively, improve the accuracy of maps generated using large occlusions, and accelerate convergence of saliency maps in cases where sampling density is set to extreme values.

We perform a quantitative comparison of accuracy of saliency maps produced by VRISE and RISE on the validation split of ILSVRC2012 using a saliency-guided content insertion/deletion metric and a pointing metric based on bounding boxes.
Our results show a modest but universal improvement of accuracy measured by insertion/deletion metric, but performance in pointing game is improved only in configurations using large occlusions. We also notice greater improvement for images containing multiple object instances.

Additionally, we explore the space of configurable parameters for RISE and VRISE on a small dataset and make observations regarding relationships between configuration of the algorithms and accuracy of results.

We also describe and demonstrate two effects observed over the course of experimentation, stemming from the random sampling approach of RISE: "feature slicing" and "saliency misattribution"

\textbf{Keywords:} Explainable Artificial Intelligence, Computer Vision, Neural Networks, Machine Learning

\textbf{Field of science and technology in accordance with OECD requirements:} natural sciences, Computer and information sciences, Computer sciences, information science and bioinformatics

%% file: meta/symbols.tex
\chapter*{List of important symbols and abbreviations}
\addcontentsline{toc}{chapter}{List of important symbols and abbreviations}

\noindent
\begin{tabularx}{\textwidth}{p{3cm}p{0.75cm}X}

	\texttt{\#X} & --- & "Number of instances of X". i.e. \texttt{\#objects} - "number of objects" \\

	$a\Delta$/abs$\Delta$/absdiff & --- & absolute difference: $abs\Delta(X, Y) = X - Y$ \\
	$r\Delta$/rel$\Delta$/reldiff	& --- & relative difference: $rel\Delta(X, Y) = \frac{X - Y}{Y}$ \\

	AltGame		& --- & Alteration Game \\
	AuC				& --- & Area Under Curve \\

	blur $\sigma$ & --- & Strength of Gaussian Blur applied to occlusion masks by VRISE (arbitrary unit) \\

	cell 		  & --- & A single cell of an occlusion mesh (alt. "\texttt{polygon}") \\
	\textit{checkpoint}	& --- & A snapshot of a Saliency Map's state captured during map generation process \\
	CGG				& --- & Coordinate-based Grid Generator	\\

	FR				& ---	& Fill Rate	\\
	FP16 			& --- & 16-bit floating point number \\
	FP32 			& --- & 32-bit floating point number \\

	HOG		    & --- & Histogram of Oriented Gradients \\
	HGG				& --- & Hybrid Grid Generator \\

	inter-run/intra-set & --- & all results for a single unique set of parameters, gathered across independent evaluations. The counterparts of these terms ("intra-run" and "inter-set") are never used in this work \\

	\textit{mask} 			& --- & An occlusion pattern applied to the input image to obtain information about location of salient features. (alt. \textit{occlusion pattern}) \\
	MC				& --- & Abbreviation of \texttt{meshcount} \\
	\texttt{meshcount} & --- & Number of unique Voronoi meshes used to generate occlusions. \\

	$N_{masks}$ & --- & Number of masks (occlusion patterns) used to generate a Saliency Map. \\

	obj & --- & Abbreviation for "objects" \\
	observed class 	& --- & Object class being explained by a saliency map. \\
	occlusion selector & --- & A GridGen used to select subsets of Voronoi cells to be rendered as \textit{masks}. \\

	$P_U$			& ---	& Likelihood of occurrence of an uninformative occlusion mask \\
	$paramset$ & --- & A set of parameters configuring (V)RISE. Influences Saliency Map generation process and its result. \\
	\textit{polygon}   & --- & A single cell of an occlusion mesh (a Voronoi mesh in the context of VRISE or a grid of squares in the context of RISE) \\
	PG				& --- & Pointing Game \\
	PGG				& --- & Permutation-based Grid Generator	\\

	$\sigma$ 	& --- & see "blur $\sigma$" \\
	SMap 			& --- & Saliency Map \\
	SSIM		  & --- & Structural Similarity Index Measure \\

	TGG				& --- & Threshold-based Grid Generator	\\
\end{tabularx}

%% file: tex/fluff/introduction.tex
The tremendous advances in the field of Machine Learning over the past decade and advancing adoption of its discoveries in commercial and industrial applications has created a growing interest in methods which interpret and explain decisions made by complex machine learning models, such as deep neural networks.

In some domains, like finance, explainability has already become required by law \cite{credit_scoring}, as credit scoring systems must be able to provide explanations of their decisions.
Another example is the recent GDPR regulation, which contains a clause granting individuals the right to explanation of algorithmic decisions pertaining to them \cite{right_to_explanation}.
Errors in systems involved in making critical decisions, like control systems for autonomous vehicles, can have catastrophic results \cite{volvocrash}, and the capacity to accurately and confidently determine the causes of erroneous decisions will be crucial to further adoption of these systems.

Examples of situations in which ML models provide the right predictions for the wrong reasons, like an image classifier identifying wolves through presence of snow in the background of an image \cite{lime} or a similar case involving recognition of horses solely by presence of a watermark \cite{horse_watermark}, are not uncommon. Even though such comical outcomes are often caused by composition of the training dataset, without methods of inquiry into the decision-making process of a machine learning model, correct identification of causes of these issues would be difficult.

Even though the first attempts to inspect decision-making rules learned by neural networks were published in 1990s \cite{1998nnExplain}, the field of Explainable Artificial Intelligence (XAI), has only begun to properly establish itself over the course of the past 5 years.

As industry demand for explainability of machine learning models has arisen only recently and the field is still very young, the definition of "explainability" remains fuzzy and the goals of XAI are an object of ongoing debate \cite{arrieta_overview}.
Usually, explainability is understood as \textit{interpretability} - a possibility of comprehending the machine learning model and identifying the basis for its decision-making process in a way understandable to humans \cite{mythos}.
Other postulated qualities include: \textit{fairness}~\cite{arrieta_overview}, which demands the learned decision making rules to abide ethics; \textit{privacy}, often understood as differential privacy~\cite{privacy}; or \textit{trustworthiness}.
These concepts are difficult to define in a rigorous manner and span a broad range of fields, from law through social sciences to philosophy.

Within the scope of this work we will define explainability as \textit{interpretability} of machine learning models.

\section{Explainability methods for Computer Vision}

In the domain of Computer Vision, explanations are usually provided as \textit{saliency maps}, which are heatmaps visualizing the importance of each pixel of input image w.r.t. prediction of a particular output class. It is essentially a visual answer to "which parts of this image cause the inspected model to recognize an object of class X?"

\begin{samepage}
  The majority of existing methods in this domain fall into one of two categories:
  \begin{itemize}
    \item \textit{gradient-based}, \textit{model-specific} techniques, which create explanations using the internal state of the model, like activation maps or connection weights.
    \item \textit{perturbation-based}, \textit{model-agnostic} methods, which only manipulate inputs and analyze resulting predictions to infer the saliency of individual input features.
  \end{itemize}
  These categories, their traits and example techniques are discussed in more detail in Chapter~\ref{ch:related_work}.
\end{samepage}

In this work we focus specifically on a single, \textit{perturbation-based} explainability algorithm named RISE - which is currently state-of-the-art in its class~\cite{survey2020} - and propose two modifications meant to improve the accuracy of its results in specific cases.

\section{The RISE algorithm}

We will now briefly describe the RISE algorithm to provide context for the introduction of our proposed modifications and goals.

RISE has three configurable \textit{parameters}:
\begin{itemize}
  \item the number of occlusion patterns used to generate a map ($N$)
  \item percentage of cells visible after occlusion ($p_1$)
  \item mesh density parameter ($s$)
\end{itemize}

These \textit{parameters} can be thought of as hyperparameters, since they can be configured arbitrarily by the user, and certain configurations might be more suitable to specific types of images.
For instance, small occlusions (high $s$) tend to yield more precise maps for multi-instance images.
Various parameter configurations also lead to varied saliency maps (see fig.~\ref{fig:gs_appendix.goldfish_demo}) and the visual difference is usually proportional to "distance" between vectors of configuration values.

\begin{figure}[H]
  \centering
  \includegraphics[width=\textwidth]{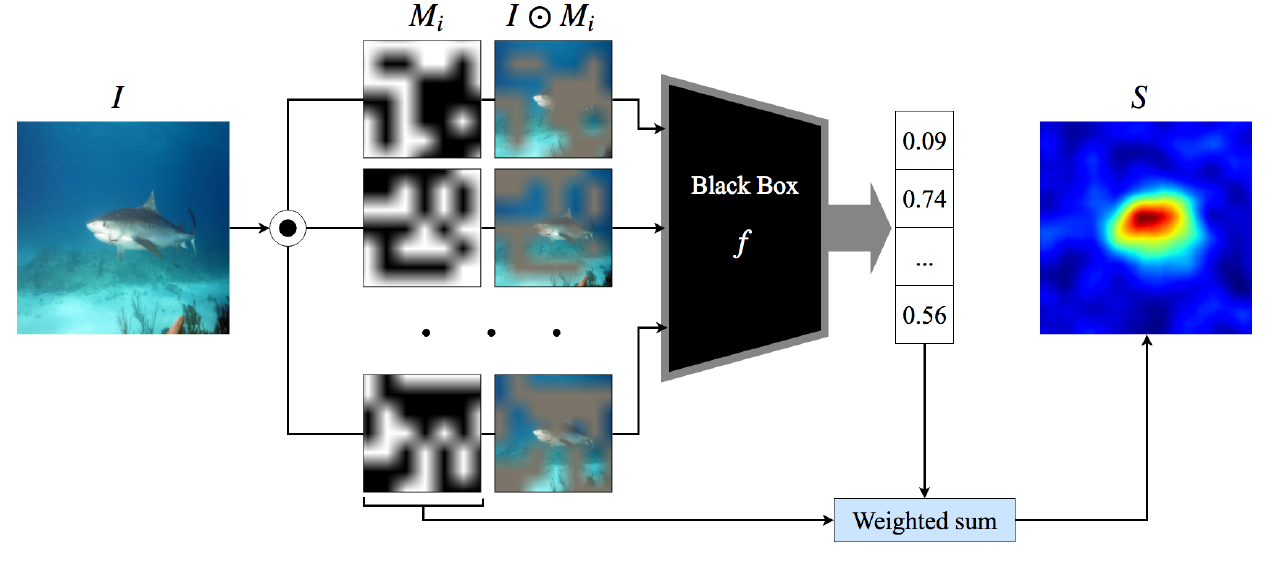}
  \caption[A diagram of RISE]{A diagram of RISE. Source: Petsiuk et al.~\cite{rise}}
  \label{fig:voronoi_mesh_example}
\end{figure}

\begin{samepage}
  The RISE algorithm works in the following manner:
  \begin{enumerate}
    \item Generate $N$ occlusion masks (steps quoted from RISE \cite{rise})
    \begin{enumerate}
      \item "\textit{Sample $N$ binary masks of size $h \times w$ (smaller than image size $H \times W$) by setting each element independently to 1 with probability $p$ and to 0 with the remaining probability}""
      \item "\textit{Upsample all masks to size $(h+1)C_H \times (w+1)C_W$ using bilinear interpolation, where $C_H \times C_W = \lfloor H/h \rfloor \times \lfloor W/w \rfloor$ is the size of the cell in the upsampled mask.}"
      \item "\textit{Crop areas $H \times W$ with uniformly random indents from $(0,0)$ up to $(C_H, C_W)$}"
    \end{enumerate}
    \item For each mask - combine the mask with input image via pointwise multiplication and resize to match the input layer of the evaluated model
    \item Multiply each mask by weighting factor $W = \frac{\text{S}}{p_1}$, where $S$ is the confidence score for explained object class
    \item Produce the saliency map as a mean of all weighted masks
  \end{enumerate}

  In essence, it's an application of the Monte Carlo method to saliency attribution.
\end{samepage}

\section{Observed issues}

\begin{enumerate}
  \item We have observed that maps generated by RISE configured with a low value of $s$ (typically lower than 6) stabilise quickly, but large occlusions and strong blur (the strength of which is effectively inversely proportional to occlusion size), make the resulting maps accurate, but \textit{coarse} and imprecise. Reducing occlusion size in order to bring out fine detail is only effective up to a certain point, as very small occlusions ($s > 12$) yield noisy maps, unless $N > 10000$,
  and converge slower.
  Furthermore, random sampling of small occlusions at low $p_1$ decreases the likelihood of capturing a salient feature in its entirety by a single occlusion mask.

  \item The original occlusion selection step (a) used by RISE leaves a possibility of occurrence of grids in which all values are set to either 0 or 1, in particular when $p_1 < \frac{1}{s^2}$ or $p_1 > 1-\frac{1}{s^2}$, respectively. Such grids would either occlude the entire image or leave all of it visible, carrying no discriminative information and, in consequence, increasing the number of evaluations required to confidently locate salient regions.
\end{enumerate}

\section{Proposed modifications}

\begin{minipage}[l]{0.55\textwidth}
  To address issue 1., in sec.~\ref{ch:vrise_description} we propose occlusions based on random Voronoi meshes, rendered in resolution matching the input of the evaluated model. Since bilinear upsampling can no longer be used to blur these occlusions, we replace it with Gaussian Blur.

  Our intuition is that random Voronoi meshes, due to irregular occlusion shapes, should be capable of capturing salient features with greater precision, especially when large occlusions are used. Use of multiple Voronoi meshes, instead of a single one, is meant to address the expected increase in variance of results caused by random layout of individual meshes.

\end{minipage}
\hspace{0.05\textwidth}
\begin{minipage}[l]{0.40\textwidth}
  \begin{figure}[H]
    \centering
    \includegraphics[width=0.85\textwidth]{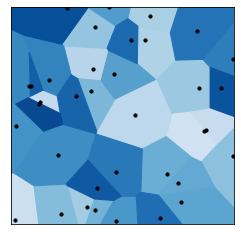}
    \caption[An example of a Voronoi Mesh]{An example of a Voronoi mesh. Black dots mark the origin points of each cell.}
    \label{fig:voronoi_mesh_example}
  \end{figure}

\end{minipage}

  To address issue 2., in sec.~\ref{sec:guaranteed_grid_gens} we propose various approaches to \textit{mask informativeness guarantee}, which would ensure that occlusion of the input is always partial. We also propose a hybrid grid generator which offers precise control over the tradeoff between likelihood of uninformative masks and extra computational cost associated with enforcement of this guarantee.

  The resulting method will still be a model-agnostic, perturbation-based saliency attribution technique.
  It will share both the positive traits inherent to this group, like applicability to all image classification models; as well as their shortcomings, like the necessity of gathering a significant number of predictions for perturbed inputs, which, in case of image classification, is costly.

  \section{Goals}
  The goals we expect our modifications to achieve are listed below. We assess them from the perspective of gathered results in section~\ref{sec:reflection_on_goals} and chapter~\ref{ch:summary}.
  \begin{enumerate}
    \item We expect the use of Voronoi occlusion meshes to achieve the following results:
    \begin{itemize}
      \item Improved map quality for low-density meshes (low \texttt{polygons}) - in these configurations RISE is forced to use large, strongly blurred occlusions and is incapable of producing sharp, well defined saliency maps. Since Voronoi occlusion shapes are random and irregular, and therefore likely to be capable of marking salient areas in a more precise way, we expect an increase of map accuracy even without reduction of blur strength. Specifically, the scores of Alteration Game metric should improve in these scenarios.

      \item Improved performance for images which contain multiple object instances - this goal partially overlaps with the previous one. RISE can perform well on multi-instance images, as shown both in the original article \cite{rise} and our reproduced results (fig.~\ref{fig:gs_appendix.goldfish_demo}), but it is not able to do so when meshes are sparse and individual occlusions are large. Again, irregular shapes and non-uniform sizes of Voronoi occlusions should improve the ability to create maps which distinguish between individual instances. This improvement will be captured primarily by the Alteration Game metric.

      \item No decrease in inter-run saliency map consistency when each mask is generated from an unique Voronoi mesh - the decision to replace random mask shifts used by RISE with a multi-mesh approach might cause our results to become less consistent. However, we expect that a large pool of Voronoi meshes will achieve a similar regularization effect to that provided by random occlusion shifts in RISE.

    \end{itemize}
    \item We expect the addition of informativeness guarantee to achieve the following results:
    \begin{itemize}
      \item Increased map convergence rate (sec.~\ref{sec:similarity_metrics}) during early stages of the map generation process, in cases where values of both \texttt{polygons} and $p_1$ are low, and therefore, the likelihood of occurrence of uninformative masks is non-negligible (fig.~\ref{fig:uninformative_prob}).
      The impact of uninformative masks declines as the total number of evaluations increases, so this modification will have the greatest impact at the very beginning of the evaluation.
      The difference in convergence rate will be measured between two sets of results generated by RISE: one instance equipped with the original RISE occlusion selector and the other configured to use an occlusion selector with guarantee.
    \end{itemize}
  \end{enumerate}

%% file: tex/fluff/related_work.tex
The literature relevant to this work consists primarily of XAI methods applicable to Convolutional Neural Networks, and metrics used to evaluate them.

\section{Methods}

The class of explainability methods characteristic to the domain of Computer Vision are \textit{gradient-based} methods. The most basic of which, and also the earliest, is a method proposed by Simonyan et al. \cite{gradientMethod}. The intuition behind it assumes that when error information for a specific class is backpropagated through the network, the resulting input map will show which inputs would have to change the least in order to affect the confidence score the most, implying that these inputs are the most relevant to the explained class.
Saliency maps produced by this method were distinctively noisy and a set of proposed refinements followed.

\begin{itemize}
  \item SmoothGrad \cite{smoothgrad} showed that averaging of saliency maps generated for several copies of the input image, each copy combined with random noise, actually increases the clarity of resulting map.

  \item Guided Backpropagation \cite{guided_backprop} produced high-resolution maps which reflected parts - usually edges - of the explained object instance. It altered the approach of original backprop by propagating only through positively-valued connections and positively activated neurons. This approach was later shown to have unintended side-effects by \cite{sanityChecks}, which we will discuss later.

  \item Layerwise Relevance Propagation (LRP) \cite{lrp}, while based on the principle of backpropagation, propagates the confidence score - instead of gradient - back through the network, using neuron activations and weights of connections between layers to trace paths leading to the most relevant inputs.
\end{itemize}

All of these methods rely on access to the internal state of the model - such as gradient information, layer activations and connection weights - and hence they belong to the group of \textit{model-specific} or \textit{white-box} methods. These techniques usually offer very good performance \cite{table_of_perf}, making them suitable for use in real-time applications \cite{zhang2016topdown}. However, they are difficult to assess in a quantitative manner and are vulnerable to adversarially modified input \cite{gradient_adversarial}.

Another noteworthy technique is the Class Activation Map \cite{cam}, which could be considered an example of \textit{explainable architectures} - models designed to provide both predictions and their explanations (e.g. SENN \cite{senn}). This particular solution is not an explainable architecture in itself but rather a "conversion kit" which inserts an additional fully-connected layer right after the last convolutional layer of the target model.
This additional layer performs avg-pooling of final feature maps, and connects to the output layer. After retraining, these connection weights reflect the relevance of each feature map to each recognized class. To obtain a saliency map, feature maps are weighted by values of connection between their corresponding output of the CAM layer and the network's output for explained class.

This technique is rather unwieldy since it not only requires retraining of the explained network but also needs to be manually adapted to each model architecture. It was quickly superseded by GradCAM \cite{gradcam}, which combines feature maps with gradient information to achieve similar results without the added inconvenience of rewiring the model.

Our target class of solutions are \textit{perturbation-based}, \textit{black-box} techniques, the first of which, referred to simply as "Occlusions" \cite{slidingWindow} was proposed in 2013 and builds explanations by moving a rectangular occlusion across an image and analysing its influence on confidence score of the explained class.
If the score decreases when a particular region is occluded - that region must be salient. This logic is the basis of all perturbation-based approaches to explanations of Convolutional Neural Networks.
Because these techniques do not access the internal state of the model, they belong to the class of \textit{model-agnostic} methods.
Other examples of perturbation-based methods are:
\begin{itemize}
  \item \textbf{RISE} \cite{rise} - samples the input with random occlusion grids. As of 2020, it is state-of-the-art within the group of perturbation-based methods~\cite{survey2020}. This method is the primary focus of our work and will be described in more detail later on.
  \item GLAS \cite{glas} - inspired by RISE, samples the input with Gaussian kernels laid out on a grid. Authors also propose a recursive variant which gradually excludes salient areas from the image, allowing the process to identify secondary areas of importance. The latter approach is particularly useful for images featuring multiple instances of a single class, where a single prominent instance may "monopolize" the map.
  \item Extremal Perturbations \cite{extremalPerturbations} - using an iterative optimization process designed to maximize the reduction of confidence score with the smallest occlusion possible. Since this approach is very similar to an adversarial attack, a regularisation term is used to avoid meaningless, pixel-space perturbations.
  \item LIME \cite{lime} - although not developed with Computer Vision in mind, LIME is a model-agnostic method and can be applied to models in this domain. When operating on image data, it uses occlusion of \textit{super-pixels} to generate perturbed examples. Currently, LIME is one of the most popular general-purpose XAI methods, along with SHAP~\cite{shap}.
  \item D-RISE \cite{d_rise} - an extension of RISE dedicated to object \textit{detection} models. The approach to input sampling remains unchanged, although the approach to weighting samples has been adjusted to accommodate differences in model output format, such as multiple predictions of varying confidence for each recognized class.
\end{itemize}

Though conceptually simple, and easily applicable to a wide range of models \cite{lime, rise}, these methods tend to be computationally intensive and slow compared to gradient-based approaches. Gathering a suitable sample may require evaluation of hundreds or thousands of perturbed images, depending on the algorithm and its configuration.
In case of RISE, 300 samples are usually enough to obtain a well-converged map from a coarse grid, but when a fine-grained grid is used, thousands or even tens of thousands of evaluations may be necessary to get a reliable and noiseless result.
However, in case of algorithms which do not optimize perturbations to a particular class - such as Occlusions, RISE or non-recursive GLAS - this disadvantage can be offset in some scenarios by their capability to simultaneously generate maps for all recognizable classes.

\section{Critique}
\label{sec:critique}
Another desirable property of black-box methods was made evident by the 2018 paper titled "Sanity checks for saliency maps" by Adebayo et al. \cite{sanityChecks}, which demonstrated that some gradient-based methods produce explanations more faithful to the input than the model itself.

The proposed sanity checks tested invariance of explanations under model and data randomization. Saliency maps are supposed to highlight evidence used by the classifier to provide a prediction. Therefore, if either the model or training labels are tampered with, the maps should change as well.
Model randomization involved replacement of progressively deeper layers of the network with random values and destroyed the predictive abilities of the model.
Data randomization involved retraining the examined model on a copy of the original training dataset with randomly shuffled labels.
If saliency maps, generated by a particular method, were not affected by either of these two modifications, it would mean that explanations provided by the method in question are simply reflecting the input data, rather than what the model is basing its prediction on.

The paper examined popular gradient-based methods and had shown that some of the refinements proposed in the literature - considered to be state-of-the-art at the time - had failed this sanity check. Methods like Guided BackProp \cite{guided_backprop} and Guided GradCAM \cite{gradcam}, were found to "\textit{approximate an elementwise product of input and gradient}" \cite{sanityChecks}, (page 8).

It's worth noting that not all of gradient-based methods are flawed. The "vanilla" gradient \cite{gradientMethod} approach and SmoothGrad \cite{smoothgrad} were shown to correctly respond to model randomization.

Black-box methods are by principle compliant with "Sanity Checks", since they rely solely on confidence scores returned by the interpreted model. If the model is destroyed by randomization, or trained on mislabeled data, the confidence scores will be incorrect and so will be the saliency map. Authors of RISE have experimentally confirmed this reasoning in their subsequent paper on application of their method to object detection models \cite{d_rise}.

\section{Metrics}
Another issue pointed out by "Sanity Checks", were the risks of assessing accuracy of saliency maps through visual inspection alone.
Due to a lack of established quantitative metrics at that time, publications often relied on qualitative, visual assessment \cite{guided_backprop, smoothgrad, gradcam}.
As maps generated for randomized networks still showed an outline of the salient object, they were no less visually convincing than maps generated for unaltered networks, even though the former had absolutely no basis for being an accurate representation of the model's state.
In consequence, solely qualitative assessment carried the risk of giving more credit to inaccurate but visually convincing explanations than accurate, but less impressive ones.

As for currently available quantitative alternatives, some rely on man-made annotations for object localization task (Pointing Game~\cite{zhang2016topdown}, overlap with bounding box~\cite{gradcam}), while others seek a human-independent approach (pixel-flipping \cite{lrp} and its extensions: AOPC~\cite{aopc} and CausalMetric~\cite{rise}).

The pixel-flipping metrics measure map quality by using saliency information contained in the map to guide the input perturbation process, starting with the most salient regions. Increasingly perturbed copies of the original image are passed to the model and resulting confidence scores for explained class are recorded. The quicker the scores decrease, the better the map.
The primary difference between AOPC and CausalMetric is how each method approaches perturbation. AOPC applies spatially contiguous perturbations ($k \times k$ non-overlapping pixel neighborhoods), while CausalMetric removes pixels in order of saliency and so the set of pixels removed in a single step might not be spatially contiguous. Furthermore, CausalMetric proposes an additional, "constructive" variant of this approach, which reintroduces the original image onto its blurred (or otherwise obfuscated) copy instead of removing it.

These human-independent metrics offer an arguably better approach, since saliency methods should be evaluated in terms of faithfulness to the explained model, not its compliance with human cognition. However, this group of metrics has recently come under scrutiny~\cite{sanity_metrics} for inconsistency and high variance of results.

Hooker et al.~\cite{koar_roar} proposed ROAR - a dedicated approach to evaluation of saliency methods, based on measuring prediction accuracy loss after retraining on dataset partially occluded by its own saliency maps. Here, saliency maps are generated for the entire dataset, then the most salient X\% of data is removed from each image according to its respective saliency map and the classifier is retrained on such modified dataset. This procedure is repeated for progressively larger values of X.
In consequence, this technique measures the loss of capacity \textit{to learn} to reliably recognize these object classes, rather than the loss of capacity \textit{to recognize} them.
A fundamental disadvantage of this technique is the enormous cost or repeated retraining, as well as unsuitability for evaluation of individual saliency maps.

However, the field has still not reached a consensus on what metrics should be used. Although solely qualitative assessment has been abandoned, an indisputably reliable quantitative metric is yet to be found. We elaborate on our choice of metrics in sec.~\ref{sec:metrics}.

\section{Voronoi mesh in saliency attribution}
We also searched for articles describing approaches similar to ours, to determine whether our idea is in fact novel.
The only instance of application of Voronoi meshes to detection of salient regions, we managed to find, is an article by Anh \& Quynh~\cite{voronoi2}.

Their approach differs significantly from ours.
Whereas our method identifies regions of an image \textit{relevant to a prediction made by a machine learning model}, their approach aims to perform an automated detection and extraction of a salient object from an image by using information metrics like local contrast and divergence. These metrics are used to guide placement of Voronoi seeds, resulting in a sort of super-pixel segmentation mesh, which is then thresholded using Otsu's algorithm to isolate the expected salient object.

Therefore, our method is distinct from that described in \cite{voronoi2}, as we do not use information contained in the image to guide the occlusion generation process and because \cite{voronoi2} is not a method of explaining predictions of machine learning models.

%% file: tex/solution/_intro.tex
Following a preliminary analysis of the accuracy of RISE saliency maps generated under various parameter configurations, we propose the following two modifications to the original method.

%% file: tex/solution/vrise_description.tex
\section{Voronoi occlusion meshes}
\label{ch:vrise_description}

Our proposed approach is simply a generalization of RISE~\cite{rise}, which relaxes the constraint placed on occlusion shapes.
Where RISE uses only square occlusions laid out on a grid, our solution, \textit{VRISE} (Voronoi-RISE), uses convex polygons - individual cells of a Voronoi mesh.

This modification is "backward-compatible" with RISE, as VRISE can be configured to use only square occlusions by being provided with mesh seed coordinates forming a square grid.
Other positive traits of RISE are also maintained:
\begin{itemize}
  \item The capacity to simultaneously generate saliency maps for all recognizable classes.
  \item Suitability to massive parallelization, due to use of randomized input sampling.
  \item Faithfulness to model, rather than input. (see sec.~\ref{sec:critique})
\end{itemize}

An important aspect of both RISE and VRISE, which significantly influences the quality of results, are the parameters used to configure the process. The set of parameters used by VRISE, which is a superset of parameters of RISE, contains the following configurable properties:
\begin{itemize}
  \item Parameters shared with RISE:
  \begin{itemize}
    \item $N$ - number of random occlusion patters used to build a single saliency map.
    \item $N_p$ - number of cells in Voronoi mesh, also referred to as "$\texttt{polygons}$", "$\texttt{cells}$" or \textit{"mesh density"}. Its counterpart in RISE is $s$, where $N_p = s^2$.
    \item $p_1$
    \begin{itemize}
      \item If a Threshold Grid Gen is used to select occlusions, $p_1$ expresses independent probability of remaining visible after masking for each cell in the mesh.
      \item If a Guaranteed GridGen is used, it expresses the percentage of cells to be left visible after masking.
    \end{itemize}
  \end{itemize}
  \item VRISE-specific parameters:
  \begin{itemize}
    \item $N_M$ - number of meshes used to generate a single saliency map. The number of masks sampled from each mesh is equal to $round(\frac{N}{N_M})$, also referred to as "$\texttt{meshcount}$".
    \item blur $\sigma$ - intensity of Gaussian blur applied to masks (more in section \ref{ch:blur_sigma})
  \end{itemize}
\end{itemize}

\subsection{Algorithm outline}
\begin{enumerate}
  \item Generate $M$ random Voronoi meshes consisting of $N_p$ seeds each. For each Voronoi mesh:
  \begin{enumerate}
    \item Generate $\frac{N}{M}$ masks by performing the following steps:
    \begin{enumerate}
      \item Select $p_1 \cdot N_p$ cells from ($i~\%~M)$-th Voronoi mesh and render them on a binary 2D array of size $H \times W$, matching the shape of the input layer of the explained model.
      \item Apply Gaussian Blur, of strength equal to $\sigma$, to the render
    \end{enumerate}
    \item Evaluate each occlusion mask:
    \begin{enumerate}
      \item Resize the evaluated image to size $H \times W$, matching the shape of the input layer
      \item Combine the blurred occlusion mask with the resized image via pointwise multiplication and feed it to the classifier
      \item Weigh the mask by the \textit{weighting factor} $\frac{\texttt{class\_confidence\_score}}{FR}$, where $FR$ is the mask's Fill Rate (mean value of the occlusion mask). If a mask occludes all pixels, this weighting factor is set to 0.
    \end{enumerate}
  \end{enumerate}
  \item Produce a saliency map as a mean of all weighted masks
\end{enumerate}

\subsection{Implementation details}

\begin{figure}[H]
  \centering
  \includegraphics[width=0.75\textwidth]{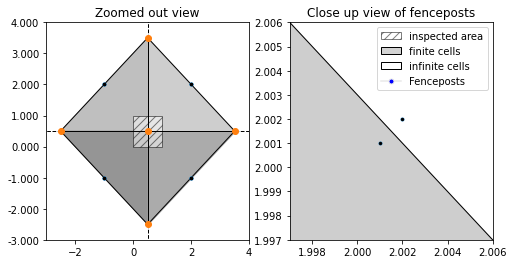}
  \caption[Demonstration of Distant Fenceposts]{Visualization of a Voronoi mesh containing only fencepost points.}
  \label{fig:vrise_desc.fenceposts.demo_viz}
\end{figure}

\subsubsection{Occlusion selection}
In RISE, occlusion selection step was implicit and occurred when the 2D grid of random values was converted to binary domain by application of $p_1$ as threshold.
In VRISE, the cells of a Voronoi mesh are enumerated and then selected according to a binary vector of shape $(1, \texttt{polygons})$, generated by a RISE grid generator (referred to as "\textit{occlusion selector}" in this context). The generator does not have to provide an informativeness guarantee, although in the Evaluation chapter we always rely on grid generators with guarantee (see sec.~\ref{sec:guaranteed_grid_gens}).

\subsubsection{\textit{Fenceposts}}
\begin{figure}[H]
  \centering
  \includegraphics[width=\textwidth]{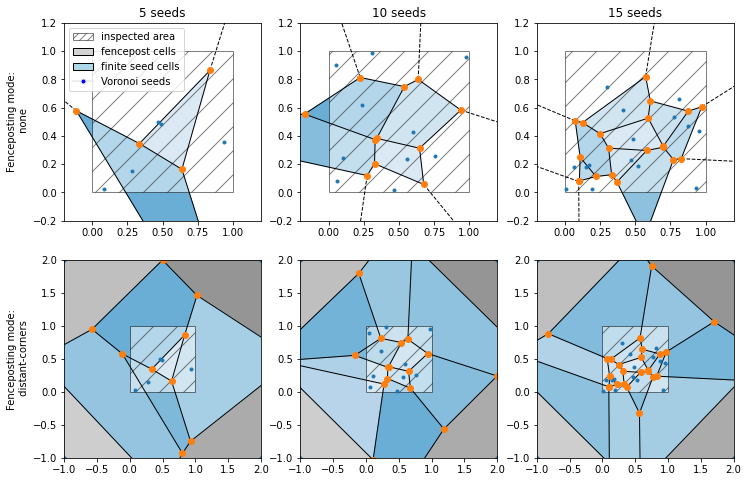}
  \caption[Sparse Voronoi meshes before and after addition of fenceposts]{Example of fenceposts' influence on the shape of a Voronoi mesh. Notice how the inspected area is never fully covered by finite cells when fenceposts are not applied.}
  \label{fig:vrise_desc.fenceposts.before_and_after}
\end{figure}

A Voronoi mesh contains two kinds of cells - finite, which are entirely enclosed by the edges of Voronoi diagram, and infinite, which are only partially enclosed. Presence of infinite cells in a Voronoi mesh is inevitable (\figref{fig:vrise_desc.fenceposts.before_and_after}), although whether they appear within the inspected area depends solely on the quantity and relative location of Voronoi seeds.
Our Voronoi mesh back-end, QHull \cite{qhull}, returns valid polygons only for finite cells, making infinite cells impossible to render without additional post-processing.

\begin{figure}[H]
  \centering
  \includegraphics[width=\textwidth]{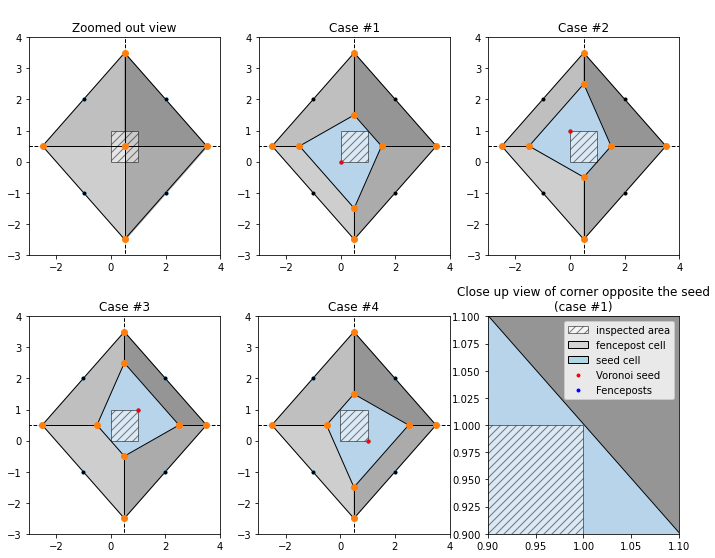}
  \caption[Visualization of Voronoi meshes with fenceposts, containing a single proper seed]{Visualization of Voronoi meshes with fenceposts, containing a single proper seed. Regardless of the placement of the seed, the entirety of inspected area is covered by its cell}
  \label{fig:vrise_desc.fenceposts.single_cell_example}
\end{figure}

To solve this problem, we have leveraged a basic property of the Voronoi diagram - each pair of neighbouring points generates a Voronoi edge between them, perpendicular to a line connecting these points.
Therefore, if four additional pairs of points (\textit{fenceposts}) were added around the rectangular inspected area, they would generate diagonal edges enclosing the inspected area and therefore cover it entirely with finite cells (\textit{fencepost cells}), as shown in \figref{fig:vrise_desc.fenceposts.demo_viz}.
Hence, any additional point added within the inspected area is guaranteed to neighbor only finite cells and therefore form a finite cell around itself, as demonstrated in \figref{fig:vrise_desc.fenceposts.single_cell_example}
To prevent the fencepost cells from intruding onto the inspected area (\textit{IA}), we place pairs of fenceposts along extensions of diagonals of the inspected area, at $d_1=3r$ and $d_2=3r+\epsilon$ away from the center point of IA (where $r$ is radius of a circle circumscribing the IA).
This way, any point within the IA will be closer to any of the seeds placed within the IA than to any of the fenceposts, which ensures that none of the fencepost cells will overlap with the IA if there's at least one seed within its bounds.

\begin{figure}[H]
  \centering
  \includegraphics[width=\textwidth]{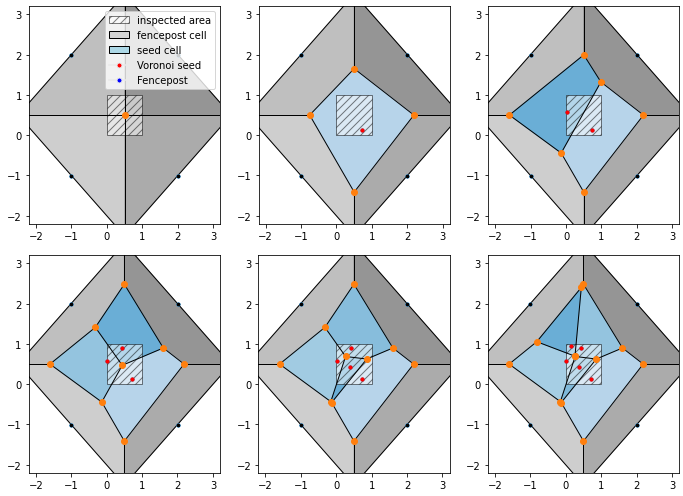}
  \caption[Demonstration of multi-seed Voronoi meshes with fenceposts]{Demonstration of multi-seed Voronoi meshes with fenceposts.}
  \label{fig:vrise_desc.fenceposts.multi_cell_example}
\end{figure}

\subsubsection{Blur}
The second, and final, step of RISE's occlusion generation algorithm uses bilinear interpolation to resize the occlusion grid to match the input layer of the network, and, in consequence, blurs the resulting mask.
Blurring smooths the edges between the occluded and visible parts of the image and reduces the impact of the phenomenon of introduced evidence mentioned in the original article. It also helps smooth out saliency maps produced by the algorithm.
In VRISE, bilinear interpolation cannot be used to apply blur, because masks are rendered in native resolution. In this situation, the target shape is equal to input shape and so bilinear interpolation would simply have no effect.
To preserve the desirable traits of blurred occlusions, we have selected Gaussian blur as a replacement for bilinear interpolation. This change introduces additional parameters which configure the blur kernel. The implementation we used, provided by PyTorch \cite{pytorch}, gives control over shape of the Gaussian kernel and, in consequence, over the intensity of blur. Bilinear interpolation does not offer this option, making blur strength strictly dependent on the ratio of grid size ($s$) to the classifier's input shape.

Influence of this change on the quality of results is examined closer in \chref{ch:grid_search}.
Chapters isolating the impact of occlusion shape on results, \ref{sec:targeted_s} and \ref{sec:imgnet}, use experimentally selected blur parameters which maximize similarity between Gaussian and Bilinear blur. The methodology is described in more detail in \chref{ch:blur_sigma}.

\subsubsection{Multi-mesh maps}
\begin{figure}[H]
  \centering
  \includegraphics[width=\textwidth]{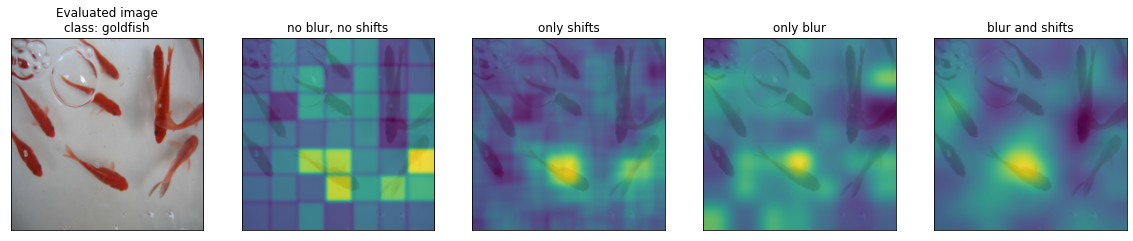}
  \caption[Demonstration of the influence of blur and shift on the state of RISE saliency maps]{Examples of RISE maps generated with and without blur and shift steps.}
  \label{fig:vrise_description.rise_blur_vs_shifts}
\end{figure}

\begin{minipage}[l]{0.4\textwidth}
Another modified occlusion pre-processing stage is the \textit{random shift}.
RISE shifts each mask by a random offset along X and Y axes, no farther than the edge length of a single occlusion along the corresponding axis.
This treatment moderates the severity of feature slicing and prevents the grid from imprinting itself onto the final saliency map, as shown in the leftmost example in \figref{fig:vrise_description.rise_blur_vs_shifts}.
Even when blur is applied, the original grid lines can still be located in the final saliency map (3rd example from the left) if this spatial jitter is not used.
\end{minipage}
\hspace{0.05\textwidth}
\begin{minipage}[r]{0.5\textwidth}
\begin{figure}[H]
  \centering
  \includegraphics[width=\textwidth]{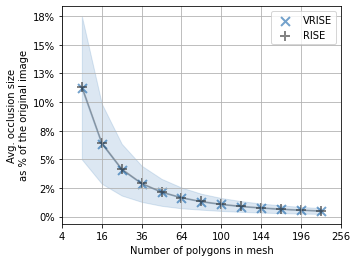}
  \caption[Average value and variance of occlusion size.]{Average value and variance of occlusion size for VRISE and RISE. Pale bands indicate standard deviation. The average occlusion size is nearly identical for both methods, but the minuscule variance of RISE occlusion size is not even visible in this scale. Occlusions for each X were sampled from 100 meshes.}
  \label{fig:vrise_description.occl_size_variance}
\end{figure}
\end{minipage}

\begin{figure}[H]
  \centering
  \includegraphics[width=\textwidth]{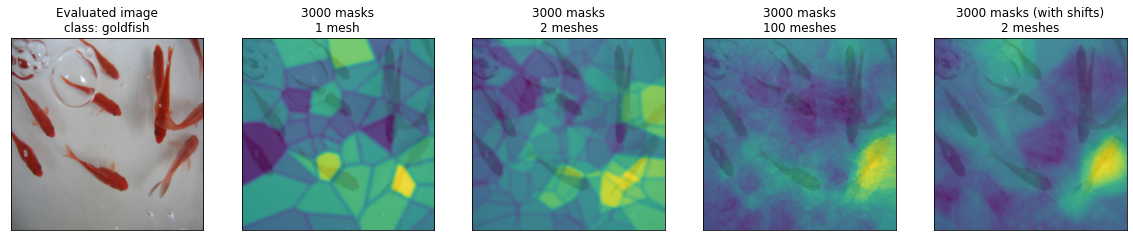}
  \caption[Demonstration of VRISE maps built with various quantities of Voronoi meshes]{Examples of VRISE maps generated using various meshcounts. High meshcount has an effect similar to random shifts.}
  \label{fig:vrise_description.meshcount_vs_blur}
\end{figure}

VRISE forgoes random shifts in favour of using multiple Voronoi meshes in a round-robin fashion, which has an effect similar to random shifts (\figref{fig:vrise_description.meshcount_vs_blur}), but should also provide an avenue of increasing the level of detail of saliency maps generated from low-density meshes.
Since Voronoi diagrams are generated randomly (as opposed to a constant, square grid of RISE), sampling from a greater number of meshes reduces the risk of basing an entire map on a mesh with an unfortunate occlusion layout.
This concern is particularly relevant to sparse meshes with few cells. As cell count increases, the absolute values of mean and standard deviation of cell size decrease (as shown in fig.~\ref{fig:vrise_description.occl_size_variance}), and individual occlusions become more regular from the perspective of fixed-size salient features.
A single mask will never mix occlusions from two different meshes, in order to avoid occlusion overlaps, which would interfere with mask weighting during saliency map composition.

\subsubsection{Disadvantages}

The core feature of VRISE - a relaxed constraint on occlusion shapes - limits its utility in domains other than computer vision and cases where data does not have 2D spatial relationships.
When applied to 1-dimensional input in domains like NLP, VRISE would be indistinguishable from RISE.
In three and more dimensions, VRISE could still be applied, although it would require an implementation of convex hull computation algorithm with support for n-dimensional spaces, as QHull does not offer that.
Moreover, RISE would be easier to adapt to n-dimensional data. For RISE, occlusion generation in N dimensions does not require more than setting its grid generator to operate on N-element coordinate vectors. However, the implementation of the upsampling algorithm would need to be adjusted to support inputs of dimensionality higher than 2.

Another drawback of switching to Voronoi meshes is the negative impact on performance. The mask generation step has become significantly more time consuming, as the original grid-based approach is as computationally efficient as it is simple.
RISE masks are generated as $(s, s)$-shaped arrays of random binary values and then bilinearly upsampled to size expected by the input layer of the examined model. In this process, upsampling is the only noticeably expensive operation.
In VRISE, the mask generation process consists of 4 steps: Voronoi mesh generation, occlusion selection, occlusion rendering and mask blurring. Occlusion selection is as expensive as occlusion grid generation in RISE, but each of the remaining three steps is comparable to bilinear upsampling in terms of computational cost. Moreover, the final cost is also increased further if high values of parameters \textit{blur} $\sigma$, $p_1$ and $\texttt{polygons}$ are used.

%% file: tex/improvements/guaranteed_grid_gens.tex
\section{Mask informativeness guarantee}
\label{sec:guaranteed_grid_gens}

RISE uses a simple and effective method of generation of masking grids (or "filters", as they are referred to in the original paper), which creates an $s \times s$ array of floats and applies $p_1$ as a threshold. If the value of given cell is less than $p_1$, it's replaced by 1, otherwise by 0.
Cells equal to 0 indicate those areas of the image which will be occluded during evaluation. This algorithm will be referred to as a "threshold-based" grid generator (TGG).
The term "fill rate" (FR) refers to the percentage of cells set to 1. It may appear counterintuitive, because masks are usually thought of as solid objects. A "full" mask is expected to be entirely opaque, while an "empty" one is expected to be transparent. To align this inverted definition with common sense, fill rate could be viewed through the lens of filling the grid with fragments of the evaluated image, rather than filling a mask with occlusions.
While reliable and easy to parallelize, this threshold-based algorithm operates on a "best effort" basis and has two properties which may become problematic in certain use-cases:

\begin{minipage}[l]{0.45\textwidth}
  \begin{enumerate}
      \item Probabilistic fill rate: the threshold-based generator cannot guarantee that the effective fill rate of each individual grid will be as close to $p_1$ as grid dimensions allow. \figref{fig:thr_gen_fill_rate} demonstrates this on a sample of 2 million grids.
      \item Uninformative grids: because the expected number of unoccluded cells is equal to $s^2 * p_1$, for extreme values of $p_1$ and low values of $s$, grids containing only 0s or 1s are likely to appear. For instance: at $s=3$ and $p_1=0.1$, each grid has a 38\% chance of consisting exclusively of zeros. These grids do not contribute to the final result in a meaningful way, because they do not carry any information about the location of salient features. The likelihood of occurrence of uninformative grids ($P_U$) is defined as $P_U = p_1^{s^2} + (1-p_1)^{s^2}$. \figref{fig:uninformative_prob} visualizes these probabilities for a relevant slice of the parameter space.
  \end{enumerate}
\end{minipage}
\hspace{0.1\textwidth}% spsacing between columns
\begin{minipage}[r]{0.45\textwidth}
  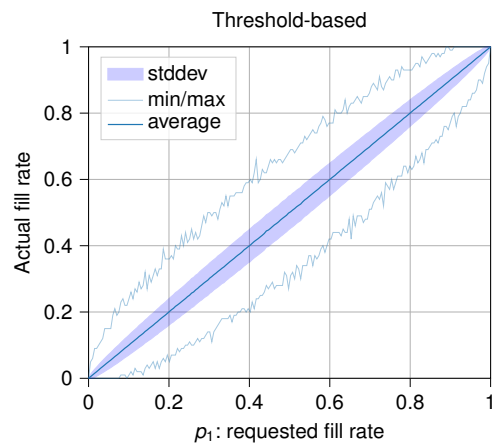
\begin{figure}[H]
      \centering
      \adjustbox{width=\textwidth, keepaspectratio}{
        \input{img/gridgen/fill_rate/thr.pgf}
      }
    \caption[Effective fill rate for Threshold-based Grid Gen]{Measured fill rates of grids at $s=10$ for the grid generation algorithm used by RISE}
    \label{fig:thr_gen_fill_rate}
  \end{figure}

  \begin{figure}[H]
    \centering
    \includegraphics[width=\textwidth]{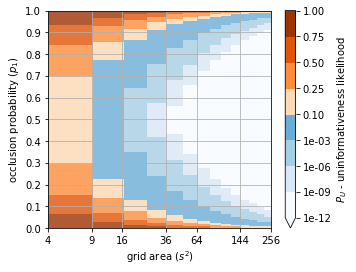}
    \caption[Uninformative grid likelihood for Threshold-based grid gen]{Likelihood of grid uninformativeness for various combinations of grid generator parameters}
    \label{fig:uninformative_prob}
  \end{figure}
\end{minipage}

The effect uninformative grids have on the saliency map depends on whether the grid is filled with zeros or ones:
\begin{itemize}
  \item an "empty" grid containing only zeros will occlude the entirety of the evaluated image. Furthermore, any saliency score it might receive will be lost in the mask combination process, after multiplication with a zero-valued grid. As a result, it uniformly deflates the map's saliency score.
  \item "full" grids, containing only ones, leave the entirety of the image visible. Those will yield a saliency score equal to class confidence score on evaluated image and evenly inflate it across the entire area of the saliency map, adding no information about the location of salient features.
\end{itemize}
In either case, the main detrimental effect is a decrease in total amount of saliency information obtainable from a constant number of evaluations. Uniform inflation or deflation of the map's saliency can be amended by simple min-max normalization.

\input{tex/improvements/grid_gen_impls_subsec_split.tex}

\subsection{Purpose}
\begin{figure}[H]
  \centering
  \includegraphics[width=\textwidth]{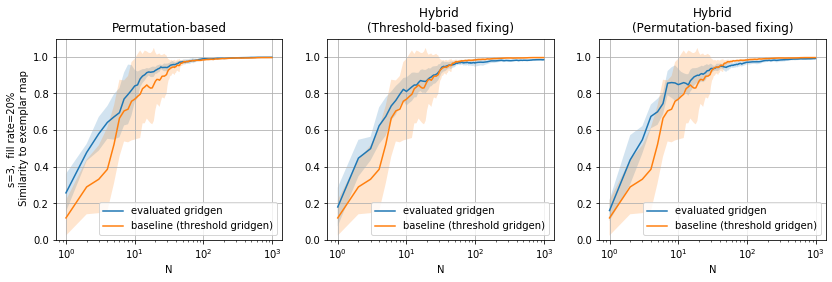}
  \caption[Comparison of convergence rates between generators with and without informativeness guarantee]{A comparison of convergence rates (similarity to a well-converged map) between generators with mask informativeness guarantee (blue) and without (orange). The informativeness guarantee noticeably improves the rate of convergence at the beginning of the process}
  \label{fig:gridgen_eval_convergence}
\end{figure}

Because uninformative masks do not introduce any new discriminative information, they stall the saliency map generation process. This effect can be demonstrated by measuring similarity between a well-converged saliency map and two unfinished maps generated by generators with- and without informativeness guarantee. At low values of $N$, a map free of uninformative masks should be more similar to a well-converged map than its baseline counterpart. This is demonstrated in \figref{fig:gridgen_eval_convergence}. Maps created by a generator with mask informativeness guarantee (blue data series) converge faster than those created by the standard generator (orange series). The higher the likelihood of uninformative masks, the slower the convergence.

This convergence rate evaluation measures similarity between each of the two sets of $M=7$ maps and 7 reference maps. Both map sets were generated for the same values of parameters $s$ and $p_1$. All maps were generated independently.
Each of non-reference maps was composed of $N=n{10^i} \cap 0\leq{N}\leq10000$ masks, while references were made up of $N_r=50000$ strictly informative masks. The criterion for choice of $N_r$ was intra-group consistency of reference maps, which, as tbl.~\ref{tbl:gridgen.ref_conv} shows, is sufficient for values of $p_1$ and $s$ used here. Similarity to reference was measured with SSIM.

\input{tbl/grid_gen_ref_conv.tex}

\begin{figure}[H]
  \centering
  \includegraphics[width=\textwidth]{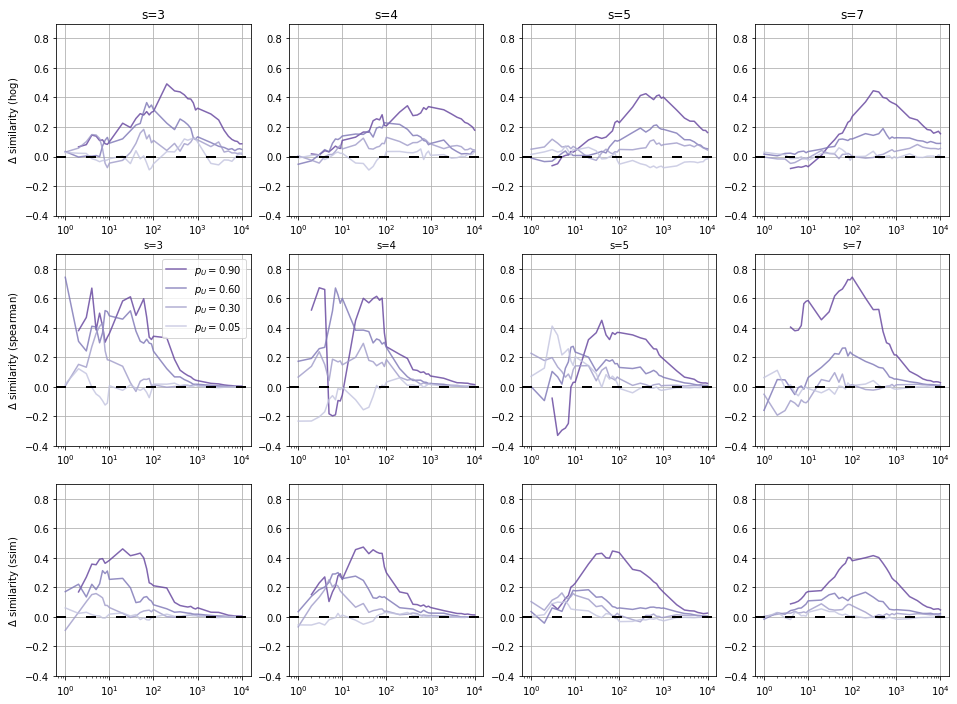}
  \caption[Difference in map consistency between map generators with and without mask informativeness guarantee]{Maps generated by a grid generator with informativeness guarantee are more consistent than their counterparts without this guarantee. The difference is more pronounced when $P_U$ is high.}
  \label{fig:gridgen_eval_consistency}
\end{figure}

\figref{fig:gridgen_eval_consistency} shows a positive impact of informativeness guarantee on consistency of results, especially early on in the map generation process. The data series in this figure show the absolute difference in average intra-group similarity scores between the guaranteed and baseline groups. Positive values of data series in this figure indicate that the 7 maps in the "guaranteed" group are on average more similar to one another than the 7 maps in the baseline group. As $N$ grows the impact of uninformative masks diminishes and both map groups become more consistent and mutually similar, the differences in similarity approach zero.
Nevertheless, these plots demonstrate that the size of consistency gap is correlated with the value of $P_U$ and the least consistent group is the one with the highest likelihood of occurrence of uninformative masks.

\pagebreak

\subsection{Concerns}
Grid generators with informativeness guarantee establish a lower limit of grid fill rate, equal to $\frac{1}{s^2}$, which silently boosts the average fill rate of maps for parameter sets where $p_1 < \frac{1}{s^2}$, which may lead to false conclusions about the quality of results.
After all, given any discrete grid, FR can only be equal to multiples of $\frac{1}{s^2}$, as occlusion cells cannot be partially filled.
In consequence, if grids consist of very few cells, the resulting mask will either be uninformative or occlude noticeably more or noticeably less area than requested.
This effect should be kept in mind when evaluating quality of saliency maps generated from $p_1 < \frac{1}{s^2}$ or $s < \sqrt{\frac{1}{p_1}}$.

\subsection{Performance measurements}
\input{tbl/grid_gen_perf.tex}

%% file: img/gridgen/fill_rate/thr.pgf
% This file was created by tikzplotlib v0.9.4.
\begin{tikzpicture}

\definecolor{color0}{rgb}{0.12156862745098,0.466666666666667,0.705882352941177}

\begin{axis}[
legend cell align={left},
legend style={fill opacity=0.8, draw opacity=1, text opacity=1, at={(0.03,0.97)}, anchor=north west, draw=white!80!black},
tick align=outside,
tick pos=left,
title={Threshold-based},
x grid style={white!69.0196078431373!black},
xlabel={\(\displaystyle p_1\): requested fill rate},
xmajorgrids,
xmin=0, xmax=1,
xtick style={color=black},
y grid style={white!69.0196078431373!black},
ylabel={Actual fill rate},
ymajorgrids,
ymin=0, ymax=1,
ytick style={color=black}
]
\path [draw=blue, fill=blue, opacity=0.2]
(axis cs:0,0)
--(axis cs:0,0)
--(axis cs:0.0050251256281407,-0.00212871190160513)
--(axis cs:0.0100502512562814,0.000102304853498936)
--(axis cs:0.0150753768844221,0.00291052740067244)
--(axis cs:0.0201005025125628,0.0060252221301198)
--(axis cs:0.0251256281407035,0.00964062102138996)
--(axis cs:0.0301507537688442,0.0135329328477383)
--(axis cs:0.0351758793969849,0.0169484037905931)
--(axis cs:0.0402010050251256,0.0206644125282764)
--(axis cs:0.0452261306532663,0.02479312941432)
--(axis cs:0.050251256281407,0.0283554252237082)
--(axis cs:0.0552763819095477,0.0321830175817013)
--(axis cs:0.0603015075376884,0.0364469047635794)
--(axis cs:0.0653266331658292,0.0406994167715311)
--(axis cs:0.0703517587939698,0.0443953908979893)
--(axis cs:0.0753768844221106,0.0488773975521326)
--(axis cs:0.0804020100502513,0.0532148685306311)
--(axis cs:0.085427135678392,0.0576874669641256)
--(axis cs:0.0904522613065327,0.0622212197631598)
--(axis cs:0.0954773869346734,0.0661053229123354)
--(axis cs:0.100502512562814,0.0703075230121613)
--(axis cs:0.105527638190955,0.0749084465205669)
--(axis cs:0.110552763819095,0.0790960527956486)
--(axis cs:0.115577889447236,0.0836581885814667)
--(axis cs:0.120603015075377,0.0880530178546906)
--(axis cs:0.125628140703518,0.0925223529338837)
--(axis cs:0.130653266331658,0.0965220518410206)
--(axis cs:0.135678391959799,0.101161446422338)
--(axis cs:0.14070351758794,0.106567863374949)
--(axis cs:0.14572864321608,0.110680960118771)
--(axis cs:0.150753768844221,0.115047872066498)
--(axis cs:0.155778894472362,0.118803422898054)
--(axis cs:0.160804020100503,0.123672153800726)
--(axis cs:0.165829145728643,0.128873199224472)
--(axis cs:0.170854271356784,0.13245140388608)
--(axis cs:0.175879396984925,0.138445895165205)
--(axis cs:0.180904522613065,0.142476476728916)
--(axis cs:0.185929648241206,0.147416260093451)
--(axis cs:0.190954773869347,0.151484791189432)
--(axis cs:0.195979899497487,0.155915640294552)
--(axis cs:0.201005025125628,0.161187015473843)
--(axis cs:0.206030150753769,0.166274450719357)
--(axis cs:0.21105527638191,0.169896990060806)
--(axis cs:0.21608040201005,0.175438441336155)
--(axis cs:0.221105527638191,0.179954919964075)
--(axis cs:0.226130653266332,0.183884859085083)
--(axis cs:0.231155778894472,0.18941954523325)
--(axis cs:0.236180904522613,0.19379311427474)
--(axis cs:0.241206030150754,0.198179695755243)
--(axis cs:0.246231155778894,0.203492175787687)
--(axis cs:0.251256281407035,0.207172218710184)
--(axis cs:0.256281407035176,0.212998241186142)
--(axis cs:0.261306532663317,0.216182731091976)
--(axis cs:0.266331658291457,0.222197074443102)
--(axis cs:0.271356783919598,0.227304514497519)
--(axis cs:0.276381909547739,0.232429251074791)
--(axis cs:0.281407035175879,0.236336521804333)
--(axis cs:0.28643216080402,0.241790696978569)
--(axis cs:0.291457286432161,0.246193688362837)
--(axis cs:0.296482412060302,0.249816387891769)
--(axis cs:0.301507537688442,0.256361160427332)
--(axis cs:0.306532663316583,0.260640725493431)
--(axis cs:0.311557788944724,0.266454931348562)
--(axis cs:0.316582914572864,0.27069865167141)
--(axis cs:0.321608040201005,0.275078419595957)
--(axis cs:0.326633165829146,0.280059218406677)
--(axis cs:0.331658291457286,0.285249661654234)
--(axis cs:0.336683417085427,0.288305435329676)
--(axis cs:0.341708542713568,0.29439439624548)
--(axis cs:0.346733668341709,0.299133315682411)
--(axis cs:0.351758793969849,0.304527785629034)
--(axis cs:0.35678391959799,0.308645784854889)
--(axis cs:0.361809045226131,0.31344748288393)
--(axis cs:0.366834170854271,0.318863894790411)
--(axis cs:0.371859296482412,0.323943793773651)
--(axis cs:0.376884422110553,0.327419597655535)
--(axis cs:0.381909547738693,0.333163790404797)
--(axis cs:0.386934673366834,0.338610164821148)
--(axis cs:0.391959798994975,0.343147233128548)
--(axis cs:0.396984924623116,0.348315868526697)
--(axis cs:0.402010050251256,0.351551987230778)
--(axis cs:0.407035175879397,0.35793250054121)
--(axis cs:0.412060301507538,0.363122895359993)
--(axis cs:0.417085427135678,0.367847867310047)
--(axis cs:0.422110552763819,0.372245687991381)
--(axis cs:0.42713567839196,0.377115793526173)
--(axis cs:0.4321608040201,0.381833147257566)
--(axis cs:0.437185929648241,0.387048788368702)
--(axis cs:0.442211055276382,0.395709339529276)
--(axis cs:0.447236180904523,0.397371016442776)
--(axis cs:0.452261306532663,0.402720414102077)
--(axis cs:0.457286432160804,0.408310513943434)
--(axis cs:0.462311557788945,0.413114063441753)
--(axis cs:0.467336683417085,0.41703487187624)
--(axis cs:0.472361809045226,0.422066818922758)
--(axis cs:0.477386934673367,0.427596613764763)
--(axis cs:0.482412060301508,0.432025760412216)
--(axis cs:0.487437185929648,0.437392212450504)
--(axis cs:0.492462311557789,0.441902313381433)
--(axis cs:0.49748743718593,0.447899047285318)
--(axis cs:0.50251256281407,0.452613323926926)
--(axis cs:0.507537688442211,0.456923495978117)
--(axis cs:0.512562814070352,0.461926713585854)
--(axis cs:0.517587939698492,0.467475797981024)
--(axis cs:0.522613065326633,0.471750408411026)
--(axis cs:0.527638190954774,0.475573413074017)
--(axis cs:0.532663316582915,0.482474308460951)
--(axis cs:0.537688442211055,0.488380718976259)
--(axis cs:0.542713567839196,0.492423977702856)
--(axis cs:0.547738693467337,0.497500408440828)
--(axis cs:0.552763819095477,0.503542758524418)
--(axis cs:0.557788944723618,0.507741384208202)
--(axis cs:0.562814070351759,0.512823916971684)
--(axis cs:0.5678391959799,0.518213454633951)
--(axis cs:0.57286432160804,0.523300718516111)
--(axis cs:0.577889447236181,0.528907004743814)
--(axis cs:0.582914572864322,0.534212045371532)
--(axis cs:0.587939698492462,0.538051404058933)
--(axis cs:0.592964824120603,0.54270938783884)
--(axis cs:0.597989949748744,0.54906789958477)
--(axis cs:0.603015075376884,0.553810991346836)
--(axis cs:0.608040201005025,0.559165980666876)
--(axis cs:0.613065326633166,0.563849732279778)
--(axis cs:0.618090452261307,0.569351516664028)
--(axis cs:0.623115577889447,0.574791938066483)
--(axis cs:0.628140703517588,0.580487195402384)
--(axis cs:0.633165829145729,0.585399121046066)
--(axis cs:0.638190954773869,0.590512383729219)
--(axis cs:0.64321608040201,0.595927260816097)
--(axis cs:0.648241206030151,0.600774638354778)
--(axis cs:0.653266331658292,0.606009427458048)
--(axis cs:0.658291457286432,0.610392358154058)
--(axis cs:0.663316582914573,0.615403719246387)
--(axis cs:0.668341708542714,0.621954210102558)
--(axis cs:0.673366834170854,0.62652787193656)
--(axis cs:0.678391959798995,0.630962319672108)
--(axis cs:0.683417085427136,0.637090407311916)
--(axis cs:0.688442211055276,0.642371978610754)
--(axis cs:0.693467336683417,0.646623156964779)
--(axis cs:0.698492462311558,0.651868473738432)
--(axis cs:0.703517587939699,0.658530067652464)
--(axis cs:0.708542713567839,0.663474105298519)
--(axis cs:0.71356783919598,0.668581284582615)
--(axis cs:0.718592964824121,0.674058303236961)
--(axis cs:0.723618090452261,0.678852714598179)
--(axis cs:0.728643216080402,0.684149373322725)
--(axis cs:0.733668341708543,0.689314864575863)
--(axis cs:0.738693467336683,0.695304896682501)
--(axis cs:0.743718592964824,0.70040038973093)
--(axis cs:0.748743718592965,0.705329824239016)
--(axis cs:0.753768844221106,0.710300605744123)
--(axis cs:0.758793969849246,0.715794172137976)
--(axis cs:0.763819095477387,0.721349485218525)
--(axis cs:0.768844221105528,0.7268903888762)
--(axis cs:0.773869346733668,0.732001081109047)
--(axis cs:0.778894472361809,0.73715141043067)
--(axis cs:0.78391959798995,0.74349869042635)
--(axis cs:0.78894472361809,0.747956398874521)
--(axis cs:0.793969849246231,0.753922004252672)
--(axis cs:0.798994974874372,0.759127236902714)
--(axis cs:0.804020100502513,0.764908466488123)
--(axis cs:0.809045226130653,0.770531285554171)
--(axis cs:0.814070351758794,0.775157980620861)
--(axis cs:0.819095477386935,0.780642285943031)
--(axis cs:0.824120603015075,0.785783912986517)
--(axis cs:0.829145728643216,0.79158578068018)
--(axis cs:0.834170854271357,0.797625981271267)
--(axis cs:0.839195979899497,0.802644453942776)
--(axis cs:0.844221105527638,0.808347187936306)
--(axis cs:0.849246231155779,0.813199628144503)
--(axis cs:0.85427135678392,0.818888913840055)
--(axis cs:0.85929648241206,0.824920207262039)
--(axis cs:0.864321608040201,0.830003879964352)
--(axis cs:0.869346733668342,0.835815288126469)
--(axis cs:0.874371859296482,0.841246336698532)
--(axis cs:0.879396984924623,0.846468262374401)
--(axis cs:0.884422110552764,0.852279979735613)
--(axis cs:0.889447236180904,0.858221132308245)
--(axis cs:0.894472361809045,0.863830721005797)
--(axis cs:0.899497487437186,0.869897481054068)
--(axis cs:0.904522613065327,0.875803960487247)
--(axis cs:0.909547738693467,0.881061607971787)
--(axis cs:0.914572864321608,0.886115619912744)
--(axis cs:0.919597989949749,0.891955574974418)
--(axis cs:0.924623115577889,0.898190669715405)
--(axis cs:0.92964824120603,0.904420712962747)
--(axis cs:0.934673366834171,0.909851999953389)
--(axis cs:0.939698492462312,0.915622597560287)
--(axis cs:0.944723618090452,0.922184083610773)
--(axis cs:0.949748743718593,0.927637116983533)
--(axis cs:0.954773869346734,0.933929048478603)
--(axis cs:0.959798994974874,0.940359190106392)
--(axis cs:0.964824120603015,0.946811240166426)
--(axis cs:0.969849246231156,0.952691925689578)
--(axis cs:0.974874371859296,0.959082068875432)
--(axis cs:0.979899497487437,0.965999176725745)
--(axis cs:0.984924623115578,0.972677479498088)
--(axis cs:0.989949748743719,0.980203092098236)
--(axis cs:0.994974874371859,0.98791432660073)
--(axis cs:1,1)
--(axis cs:1,1)
--(axis cs:1,1)
--(axis cs:0.994974874371859,1.00203370768577)
--(axis cs:0.989949748743719,0.999944865703583)
--(axis cs:0.984924623115578,0.997122515924275)
--(axis cs:0.979899497487437,0.994180867448449)
--(axis cs:0.974874371859296,0.990057883784175)
--(axis cs:0.969849246231156,0.987114058807492)
--(axis cs:0.964824120603015,0.983496744185686)
--(axis cs:0.959798994974874,0.979332849383354)
--(axis cs:0.954773869346734,0.975464977324009)
--(axis cs:0.949748743718593,0.971070868894458)
--(axis cs:0.944723618090452,0.967629935592413)
--(axis cs:0.939698492462312,0.963053459301591)
--(axis cs:0.934673366834171,0.959469942376018)
--(axis cs:0.92964824120603,0.955463310703635)
--(axis cs:0.924623115577889,0.951073355972767)
--(axis cs:0.919597989949749,0.946402469649911)
--(axis cs:0.914572864321608,0.942432334646583)
--(axis cs:0.909547738693467,0.938312357291579)
--(axis cs:0.904522613065327,0.933986054733396)
--(axis cs:0.899497487437186,0.929316524416208)
--(axis cs:0.894472361809045,0.924717271700501)
--(axis cs:0.889447236180904,0.920744817703962)
--(axis cs:0.884422110552764,0.91664807125926)
--(axis cs:0.879396984924623,0.911973781883717)
--(axis cs:0.874371859296482,0.90743163228035)
--(axis cs:0.869346733668342,0.90388872474432)
--(axis cs:0.864321608040201,0.898386098444462)
--(axis cs:0.85929648241206,0.894451826810837)
--(axis cs:0.85427135678392,0.889355052262545)
--(axis cs:0.849246231155779,0.885260354727507)
--(axis cs:0.844221105527638,0.880758799612522)
--(axis cs:0.839195979899497,0.875597514212132)
--(axis cs:0.834170854271357,0.870850004255772)
--(axis cs:0.829145728643216,0.86712422221899)
--(axis cs:0.824120603015075,0.861706111580133)
--(axis cs:0.819095477386935,0.858131751418114)
--(axis cs:0.814070351758794,0.853300042450428)
--(axis cs:0.809045226130653,0.849238764494658)
--(axis cs:0.804020100502513,0.844183530658484)
--(axis cs:0.798994974874372,0.839604757726192)
--(axis cs:0.793969849246231,0.834225993603468)
--(axis cs:0.78894472361809,0.829707618802786)
--(axis cs:0.78391959798995,0.825729362666607)
--(axis cs:0.778894472361809,0.820560548454523)
--(axis cs:0.773869346733668,0.81523297727108)
--(axis cs:0.768844221105528,0.811245616525412)
--(axis cs:0.763819095477387,0.806324474513531)
--(axis cs:0.758793969849246,0.801541838794947)
--(axis cs:0.753768844221106,0.796867448836565)
--(axis cs:0.748743718592965,0.792976211756468)
--(axis cs:0.743718592964824,0.788063608109951)
--(axis cs:0.738693467336683,0.782029125839472)
--(axis cs:0.733668341708543,0.777641154825687)
--(axis cs:0.728643216080402,0.773290645331144)
--(axis cs:0.723618090452261,0.769281230866909)
--(axis cs:0.718592964824121,0.763849750161171)
--(axis cs:0.71356783919598,0.758622728288174)
--(axis cs:0.708542713567839,0.754171945154667)
--(axis cs:0.703517587939699,0.7485759742558)
--(axis cs:0.698492462311558,0.743543494492769)
--(axis cs:0.693467336683417,0.739704824984074)
--(axis cs:0.688442211055276,0.735296044498682)
--(axis cs:0.683417085427136,0.730049647390842)
--(axis cs:0.678391959798995,0.725397638976574)
--(axis cs:0.673366834170854,0.720024142414331)
--(axis cs:0.668341708542714,0.715453736484051)
--(axis cs:0.663316582914573,0.710850290954113)
--(axis cs:0.658291457286432,0.705627653747797)
--(axis cs:0.653266331658292,0.70085459575057)
--(axis cs:0.648241206030151,0.696031339466572)
--(axis cs:0.64321608040201,0.691286779940128)
--(axis cs:0.638190954773869,0.686745654791594)
--(axis cs:0.633165829145729,0.681440860033035)
--(axis cs:0.628140703517588,0.677272852510214)
--(axis cs:0.623115577889447,0.671834081411362)
--(axis cs:0.618090452261307,0.666980542242527)
--(axis cs:0.613065326633166,0.660928323864937)
--(axis cs:0.608040201005025,0.65642997995019)
--(axis cs:0.603015075376884,0.65218099206686)
--(axis cs:0.597989949748744,0.646906092762947)
--(axis cs:0.592964824120603,0.642216645181179)
--(axis cs:0.587939698492462,0.63765462487936)
--(axis cs:0.582914572864322,0.632371969521046)
--(axis cs:0.577889447236181,0.627824958413839)
--(axis cs:0.57286432160804,0.621539283543825)
--(axis cs:0.5678391959799,0.616854604333639)
--(axis cs:0.562814070351759,0.61277212947607)
--(axis cs:0.557788944723618,0.607932634651661)
--(axis cs:0.552763819095477,0.601975224912167)
--(axis cs:0.547738693467337,0.597875606268644)
--(axis cs:0.542713567839196,0.592600021511316)
--(axis cs:0.537688442211055,0.588071297854185)
--(axis cs:0.532663316582915,0.582485694438219)
--(axis cs:0.527638190954774,0.575846560299397)
--(axis cs:0.522613065326633,0.571775645017624)
--(axis cs:0.517587939698492,0.567598197609186)
--(axis cs:0.512562814070352,0.562197312712669)
--(axis cs:0.507537688442211,0.558092463761568)
--(axis cs:0.50251256281407,0.552024632692337)
--(axis cs:0.49748743718593,0.549176927655935)
--(axis cs:0.492462311557789,0.541801657527685)
--(axis cs:0.487437185929648,0.537125788629055)
--(axis cs:0.482412060301508,0.533238261938095)
--(axis cs:0.477386934673367,0.527619376778603)
--(axis cs:0.472361809045226,0.52126919105649)
--(axis cs:0.467336683417085,0.516659133136272)
--(axis cs:0.462311557788945,0.512613959610462)
--(axis cs:0.457286432160804,0.506903488188982)
--(axis cs:0.452261306532663,0.50147757679224)
--(axis cs:0.447236180904523,0.497024990618229)
--(axis cs:0.442211055276382,0.493386682122946)
--(axis cs:0.437185929648241,0.48701923340559)
--(axis cs:0.4321608040201,0.481252837926149)
--(axis cs:0.42713567839196,0.476234190165997)
--(axis cs:0.422110552763819,0.470946293324232)
--(axis cs:0.417085427135678,0.46663212031126)
--(axis cs:0.412060301507538,0.462359115481377)
--(axis cs:0.407035175879397,0.456529505550861)
--(axis cs:0.402010050251256,0.451374016702175)
--(axis cs:0.396984924623116,0.445580150932074)
--(axis cs:0.391959798994975,0.440968796610832)
--(axis cs:0.386934673366834,0.436649806797504)
--(axis cs:0.381909547738693,0.429444201290607)
--(axis cs:0.376884422110553,0.424836378544569)
--(axis cs:0.371859296482412,0.419838190078735)
--(axis cs:0.366834170854271,0.41509411111474)
--(axis cs:0.361809045226131,0.409696511924267)
--(axis cs:0.35678391959799,0.404850244522095)
--(axis cs:0.351758793969849,0.399536225944757)
--(axis cs:0.346733668341709,0.394452676177025)
--(axis cs:0.341708542713568,0.388669587671757)
--(axis cs:0.336683417085427,0.383766558021307)
--(axis cs:0.331658291457286,0.378292310982943)
--(axis cs:0.326633165829146,0.373396754264832)
--(axis cs:0.321608040201005,0.369155582040548)
--(axis cs:0.316582914572864,0.362823322415352)
--(axis cs:0.311557788944724,0.357975069433451)
--(axis cs:0.306532663316583,0.353165283799171)
--(axis cs:0.301507537688442,0.348236825317144)
--(axis cs:0.296482412060302,0.342019587755203)
--(axis cs:0.291457286432161,0.336938340216875)
--(axis cs:0.28643216080402,0.332097306847572)
--(axis cs:0.281407035175879,0.326987452805042)
--(axis cs:0.276381909547739,0.321464732289314)
--(axis cs:0.271356783919598,0.314859513193369)
--(axis cs:0.266331658291457,0.311078947037458)
--(axis cs:0.261306532663317,0.304693259298801)
--(axis cs:0.256281407035176,0.300985783338547)
--(axis cs:0.251256281407035,0.294175799936056)
--(axis cs:0.246231155778894,0.289339829236269)
--(axis cs:0.241206030150754,0.283292289823294)
--(axis cs:0.236180904522613,0.27793288603425)
--(axis cs:0.231155778894472,0.272586457431316)
--(axis cs:0.226130653266332,0.267955154180527)
--(axis cs:0.221105527638191,0.26259708032012)
--(axis cs:0.21608040201005,0.258649550378323)
--(axis cs:0.21105527638191,0.252007007598877)
--(axis cs:0.206030150753769,0.246971555054188)
--(axis cs:0.201005025125628,0.241446979343891)
--(axis cs:0.195979899497487,0.235790349543095)
--(axis cs:0.190954773869347,0.229753222316504)
--(axis cs:0.185929648241206,0.224507752805948)
--(axis cs:0.180904522613065,0.219941519200802)
--(axis cs:0.175879396984925,0.213788110762835)
--(axis cs:0.170854271356784,0.208518605679274)
--(axis cs:0.165829145728643,0.202652812004089)
--(axis cs:0.160804020100503,0.197063837200403)
--(axis cs:0.155778894472362,0.192346591502428)
--(axis cs:0.150753768844221,0.186892122030258)
--(axis cs:0.14572864321608,0.180715031921864)
--(axis cs:0.14070351758794,0.175896134227514)
--(axis cs:0.135678391959799,0.169404540210962)
--(axis cs:0.130653266331658,0.164639960974455)
--(axis cs:0.125628140703518,0.159545660018921)
--(axis cs:0.120603015075377,0.152840986847878)
--(axis cs:0.115577889447236,0.147387817502022)
--(axis cs:0.110552763819095,0.141961943358183)
--(axis cs:0.105527638190955,0.136601556092501)
--(axis cs:0.100502512562814,0.130296483635902)
--(axis cs:0.0954773869346734,0.12476067058742)
--(axis cs:0.0904522613065327,0.119664782658219)
--(axis cs:0.085427135678392,0.113532537594438)
--(axis cs:0.0804020100502513,0.107753137126565)
--(axis cs:0.0753768844221106,0.101272603496909)
--(axis cs:0.0703517587939698,0.0958326123654842)
--(axis cs:0.0653266331658292,0.0903905835002661)
--(axis cs:0.0603015075376884,0.0839430931955576)
--(axis cs:0.0552763819095477,0.0777469836175442)
--(axis cs:0.050251256281407,0.0723085720092058)
--(axis cs:0.0452261306532663,0.0664168708026409)
--(axis cs:0.0402010050251256,0.060267586261034)
--(axis cs:0.0351758793969849,0.053943594917655)
--(axis cs:0.0301507537688442,0.0476350672543049)
--(axis cs:0.0251256281407035,0.0406693797558546)
--(axis cs:0.0201005025125628,0.0337907774373889)
--(axis cs:0.0150753768844221,0.0274114729836583)
--(axis cs:0.0100502512562814,0.0200416957959533)
--(axis cs:0.0050251256281407,0.012198711745441)
--(axis cs:0,0)
--cycle;
\addlegendimage{area legend, draw=blue, fill=blue, opacity=0.2}
\addlegendentry{stddev}

\addplot [color0, opacity=0.4]
table {%
0 0
0.0050251256281407 0
0.0100502512562814 0
0.0150753768844221 0
0.0201005025125628 0
0.0251256281407035 0
0.0301507537688442 0
0.0351758793969849 0
0.0402010050251256 0
0.0452261306532663 0
0.050251256281407 0
0.0552763819095477 0
0.0603015075376884 0
0.0653266331658292 0
0.0703517587939698 0
0.0753768844221106 0
0.0804020100502513 0.00999999977648258
0.085427135678392 0.00999999977648258
0.0904522613065327 0.00999999977648258
0.0954773869346734 0
0.100502512562814 0.00999999977648258
0.105527638190955 0.0199999995529652
0.110552763819095 0.00999999977648258
0.115577889447236 0.0299999993294477
0.120603015075377 0.0199999995529652
0.125628140703518 0.0199999995529652
0.130653266331658 0.0199999995529652
0.135678391959799 0.00999999977648258
0.14070351758794 0.0299999993294477
0.14572864321608 0.0399999991059303
0.150753768844221 0.0299999993294477
0.155778894472362 0.0399999991059303
0.160804020100503 0.0500000007450581
0.165829145728643 0.0299999993294477
0.170854271356784 0.0500000007450581
0.175879396984925 0.0299999993294477
0.180904522613065 0.0500000007450581
0.185929648241206 0.0599999986588955
0.190954773869347 0.0500000007450581
0.195979899497487 0.0700000002980232
0.201005025125628 0.0500000007450581
0.206030150753769 0.0700000002980232
0.21105527638191 0.0700000002980232
0.21608040201005 0.0700000002980232
0.221105527638191 0.0900000035762787
0.226130653266332 0.0799999982118607
0.231155778894472 0.0900000035762787
0.236180904522613 0.0900000035762787
0.241206030150754 0.100000001490116
0.246231155778894 0.100000001490116
0.251256281407035 0.109999999403954
0.256281407035176 0.100000001490116
0.261306532663317 0.100000001490116
0.266331658291457 0.109999999403954
0.271356783919598 0.119999997317791
0.276381909547739 0.109999999403954
0.281407035175879 0.119999997317791
0.28643216080402 0.119999997317791
0.291457286432161 0.129999995231628
0.296482412060302 0.140000000596046
0.301507537688442 0.140000000596046
0.306532663316583 0.109999999403954
0.311557788944724 0.140000000596046
0.316582914572864 0.150000005960464
0.321608040201005 0.159999996423721
0.326633165829146 0.170000001788139
0.331658291457286 0.180000007152557
0.336683417085427 0.180000007152557
0.341708542713568 0.180000007152557
0.346733668341709 0.180000007152557
0.351758793969849 0.189999997615814
0.35678391959799 0.170000001788139
0.361809045226131 0.189999997615814
0.366834170854271 0.189999997615814
0.371859296482412 0.200000002980232
0.376884422110553 0.200000002980232
0.381909547738693 0.209999993443489
0.386934673366834 0.209999993443489
0.391959798994975 0.200000002980232
0.396984924623116 0.209999993443489
0.402010050251256 0.200000002980232
0.407035175879397 0.230000004172325
0.412060301507538 0.209999993443489
0.417085427135678 0.239999994635582
0.422110552763819 0.239999994635582
0.42713567839196 0.25
0.4321608040201 0.25
0.437185929648241 0.259999990463257
0.442211055276382 0.280000001192093
0.447236180904523 0.270000010728836
0.452261306532663 0.280000001192093
0.457286432160804 0.28999999165535
0.462311557788945 0.270000010728836
0.467336683417085 0.280000001192093
0.472361809045226 0.280000001192093
0.477386934673367 0.280000001192093
0.482412060301508 0.280000001192093
0.487437185929648 0.319999992847443
0.492462311557789 0.310000002384186
0.49748743718593 0.270000010728836
0.50251256281407 0.310000002384186
0.507537688442211 0.319999992847443
0.512562814070352 0.300000011920929
0.517587939698492 0.340000003576279
0.522613065326633 0.330000013113022
0.527638190954774 0.300000011920929
0.532663316582915 0.340000003576279
0.537688442211055 0.340000003576279
0.542713567839196 0.340000003576279
0.547738693467337 0.370000004768372
0.552763819095477 0.370000004768372
0.557788944723618 0.340000003576279
0.562814070351759 0.370000004768372
0.5678391959799 0.360000014305115
0.57286432160804 0.370000004768372
0.577889447236181 0.389999985694885
0.582914572864322 0.400000005960464
0.587939698492462 0.379999995231628
0.592964824120603 0.400000005960464
0.597989949748744 0.419999986886978
0.603015075376884 0.419999986886978
0.608040201005025 0.439999997615814
0.613065326633166 0.439999997615814
0.618090452261307 0.439999997615814
0.623115577889447 0.439999997615814
0.628140703517588 0.430000007152557
0.633165829145729 0.449999988079071
0.638190954773869 0.439999997615814
0.64321608040201 0.439999997615814
0.648241206030151 0.479999989271164
0.653266331658292 0.419999986886978
0.658291457286432 0.490000009536743
0.663316582914573 0.46000000834465
0.668341708542714 0.490000009536743
0.673366834170854 0.490000009536743
0.678391959798995 0.490000009536743
0.683417085427136 0.469999998807907
0.688442211055276 0.479999989271164
0.693467336683417 0.479999989271164
0.698492462311558 0.509999990463257
0.703517587939699 0.509999990463257
0.708542713567839 0.529999971389771
0.71356783919598 0.540000021457672
0.718592964824121 0.550000011920929
0.723618090452261 0.540000021457672
0.728643216080402 0.560000002384186
0.733668341708543 0.519999980926514
0.738693467336683 0.589999973773956
0.743718592964824 0.579999983310699
0.748743718592965 0.569999992847443
0.753768844221106 0.600000023841858
0.758793969849246 0.569999992847443
0.763819095477387 0.589999973773956
0.768844221105528 0.610000014305115
0.773869346733668 0.610000014305115
0.778894472361809 0.610000014305115
0.78391959798995 0.639999985694885
0.78894472361809 0.639999985694885
0.793969849246231 0.620000004768372
0.798994974874372 0.639999985694885
0.804020100502513 0.629999995231628
0.809045226130653 0.639999985694885
0.814070351758794 0.670000016689301
0.819095477386935 0.649999976158142
0.824120603015075 0.660000026226044
0.829145728643216 0.680000007152557
0.834170854271357 0.649999976158142
0.839195979899497 0.689999997615814
0.844221105527638 0.689999997615814
0.849246231155779 0.699999988079071
0.85427135678392 0.709999978542328
0.85929648241206 0.709999978542328
0.864321608040201 0.720000028610229
0.869346733668342 0.709999978542328
0.874371859296482 0.730000019073486
0.879396984924623 0.730000019073486
0.884422110552764 0.759999990463257
0.889447236180904 0.759999990463257
0.894472361809045 0.75
0.899497487437186 0.769999980926514
0.904522613065327 0.790000021457672
0.909547738693467 0.800000011920929
0.914572864321608 0.759999990463257
0.919597989949749 0.790000021457672
0.924623115577889 0.800000011920929
0.92964824120603 0.819999992847443
0.934673366834171 0.819999992847443
0.939698492462312 0.819999992847443
0.944723618090452 0.839999973773956
0.949748743718593 0.839999973773956
0.954773869346734 0.860000014305115
0.959798994974874 0.860000014305115
0.964824120603015 0.870000004768372
0.969849246231156 0.899999976158142
0.974874371859296 0.899999976158142
0.979899497487437 0.889999985694885
0.984924623115578 0.920000016689301
0.989949748743719 0.939999997615814
0.994974874371859 0.949999988079071
1 1
};
\addlegendentry{min/max}
\addplot [color0, opacity=0.4, forget plot]
table {%
0 0
0.0050251256281407 0.0500000007450581
0.0100502512562814 0.0599999986588955
0.0150753768844221 0.0900000035762787
0.0201005025125628 0.0900000035762787
0.0251256281407035 0.100000001490116
0.0301507537688442 0.109999999403954
0.0351758793969849 0.119999997317791
0.0402010050251256 0.150000005960464
0.0452261306532663 0.150000005960464
0.050251256281407 0.150000005960464
0.0552763819095477 0.150000005960464
0.0603015075376884 0.189999997615814
0.0653266331658292 0.209999993443489
0.0703517587939698 0.189999997615814
0.0753768844221106 0.200000002980232
0.0804020100502513 0.230000004172325
0.085427135678392 0.239999994635582
0.0904522613065327 0.219999998807907
0.0954773869346734 0.219999998807907
0.100502512562814 0.239999994635582
0.105527638190955 0.259999990463257
0.110552763819095 0.239999994635582
0.115577889447236 0.270000010728836
0.120603015075377 0.259999990463257
0.125628140703518 0.28999999165535
0.130653266331658 0.270000010728836
0.135678391959799 0.280000001192093
0.14070351758794 0.319999992847443
0.14572864321608 0.270000010728836
0.150753768844221 0.310000002384186
0.155778894472362 0.28999999165535
0.160804020100503 0.310000002384186
0.165829145728643 0.330000013113022
0.170854271356784 0.310000002384186
0.175879396984925 0.330000013113022
0.180904522613065 0.330000013113022
0.185929648241206 0.389999985694885
0.190954773869347 0.360000014305115
0.195979899497487 0.360000014305115
0.201005025125628 0.360000014305115
0.206030150753769 0.370000004768372
0.21105527638191 0.360000014305115
0.21608040201005 0.400000005960464
0.221105527638191 0.389999985694885
0.226130653266332 0.389999985694885
0.231155778894472 0.430000007152557
0.236180904522613 0.400000005960464
0.241206030150754 0.439999997615814
0.246231155778894 0.409999996423721
0.251256281407035 0.419999986886978
0.256281407035176 0.449999988079071
0.261306532663317 0.419999986886978
0.266331658291457 0.439999997615814
0.271356783919598 0.419999986886978
0.276381909547739 0.430000007152557
0.281407035175879 0.46000000834465
0.28643216080402 0.490000009536743
0.291457286432161 0.469999998807907
0.296482412060302 0.5
0.301507537688442 0.5
0.306532663316583 0.490000009536743
0.311557788944724 0.5
0.316582914572864 0.5
0.321608040201005 0.479999989271164
0.326633165829146 0.509999990463257
0.331658291457286 0.529999971389771
0.336683417085427 0.519999980926514
0.341708542713568 0.550000011920929
0.346733668341709 0.519999980926514
0.351758793969849 0.550000011920929
0.35678391959799 0.550000011920929
0.361809045226131 0.550000011920929
0.366834170854271 0.560000002384186
0.371859296482412 0.560000002384186
0.376884422110553 0.560000002384186
0.381909547738693 0.560000002384186
0.386934673366834 0.579999983310699
0.391959798994975 0.589999973773956
0.396984924623116 0.589999973773956
0.402010050251256 0.600000023841858
0.407035175879397 0.589999973773956
0.412060301507538 0.620000004768372
0.417085427135678 0.660000026226044
0.422110552763819 0.589999973773956
0.42713567839196 0.600000023841858
0.4321608040201 0.629999995231628
0.437185929648241 0.620000004768372
0.442211055276382 0.620000004768372
0.447236180904523 0.639999985694885
0.452261306532663 0.660000026226044
0.457286432160804 0.649999976158142
0.462311557788945 0.660000026226044
0.467336683417085 0.649999976158142
0.472361809045226 0.660000026226044
0.477386934673367 0.660000026226044
0.482412060301508 0.649999976158142
0.487437185929648 0.660000026226044
0.492462311557789 0.670000016689301
0.49748743718593 0.680000007152557
0.50251256281407 0.689999997615814
0.507537688442211 0.689999997615814
0.512562814070352 0.680000007152557
0.517587939698492 0.689999997615814
0.522613065326633 0.699999988079071
0.527638190954774 0.709999978542328
0.532663316582915 0.709999978542328
0.537688442211055 0.730000019073486
0.542713567839196 0.720000028610229
0.547738693467337 0.730000019073486
0.552763819095477 0.740000009536743
0.557788944723618 0.75
0.562814070351759 0.740000009536743
0.5678391959799 0.740000009536743
0.57286432160804 0.759999990463257
0.577889447236181 0.769999980926514
0.582914572864322 0.790000021457672
0.587939698492462 0.779999971389771
0.592964824120603 0.759999990463257
0.597989949748744 0.769999980926514
0.603015075376884 0.769999980926514
0.608040201005025 0.769999980926514
0.613065326633166 0.810000002384186
0.618090452261307 0.800000011920929
0.623115577889447 0.800000011920929
0.628140703517588 0.800000011920929
0.633165829145729 0.810000002384186
0.638190954773869 0.839999973773956
0.64321608040201 0.800000011920929
0.648241206030151 0.819999992847443
0.653266331658292 0.829999983310699
0.658291457286432 0.829999983310699
0.663316582914573 0.839999973773956
0.668341708542714 0.850000023841858
0.673366834170854 0.839999973773956
0.678391959798995 0.829999983310699
0.683417085427136 0.850000023841858
0.688442211055276 0.850000023841858
0.693467336683417 0.850000023841858
0.698492462311558 0.860000014305115
0.703517587939699 0.879999995231628
0.708542713567839 0.889999985694885
0.71356783919598 0.879999995231628
0.718592964824121 0.870000004768372
0.723618090452261 0.910000026226044
0.728643216080402 0.879999995231628
0.733668341708543 0.870000004768372
0.738693467336683 0.889999985694885
0.743718592964824 0.899999976158142
0.748743718592965 0.910000026226044
0.753768844221106 0.899999976158142
0.758793969849246 0.899999976158142
0.763819095477387 0.930000007152557
0.768844221105528 0.930000007152557
0.773869346733668 0.910000026226044
0.778894472361809 0.939999997615814
0.78391959798995 0.930000007152557
0.78894472361809 0.930000007152557
0.793969849246231 0.930000007152557
0.798994974874372 0.930000007152557
0.804020100502513 0.930000007152557
0.809045226130653 0.939999997615814
0.814070351758794 0.949999988079071
0.819095477386935 0.949999988079071
0.824120603015075 0.939999997615814
0.829145728643216 0.970000028610229
0.834170854271357 0.959999978542328
0.839195979899497 0.949999988079071
0.844221105527638 0.959999978542328
0.849246231155779 0.970000028610229
0.85427135678392 0.980000019073486
0.85929648241206 0.980000019073486
0.864321608040201 0.970000028610229
0.869346733668342 0.980000019073486
0.874371859296482 0.970000028610229
0.879396984924623 0.980000019073486
0.884422110552764 0.980000019073486
0.889447236180904 1
0.894472361809045 1
0.899497487437186 0.990000009536743
0.904522613065327 0.990000009536743
0.909547738693467 1
0.914572864321608 1
0.919597989949749 1
0.924623115577889 1
0.92964824120603 1
0.934673366834171 1
0.939698492462312 1
0.944723618090452 1
0.949748743718593 1
0.954773869346734 1
0.959798994974874 1
0.964824120603015 1
0.969849246231156 1
0.974874371859296 1
0.979899497487437 1
0.984924623115578 1
0.989949748743719 1
0.994974874371859 1
1 1
};
\addplot [semithick, color0]
table {%
0 0
0.0050251256281407 0.00503499992191792
0.0100502512562814 0.0100720003247261
0.0150753768844221 0.0151610001921654
0.0201005025125628 0.0199079997837543
0.0251256281407035 0.0251550003886223
0.0301507537688442 0.0305840000510216
0.0351758793969849 0.0354459993541241
0.0402010050251256 0.0404659993946552
0.0452261306532663 0.0456050001084805
0.050251256281407 0.050331998616457
0.0552763819095477 0.0549650005996227
0.0603015075376884 0.0601949989795685
0.0653266331658292 0.0655450001358986
0.0703517587939698 0.0701140016317368
0.0753768844221106 0.0750750005245209
0.0804020100502513 0.080484002828598
0.085427135678392 0.0856100022792816
0.0904522613065327 0.0909430012106895
0.0954773869346734 0.0954329967498779
0.100502512562814 0.100302003324032
0.105527638190955 0.105755001306534
0.110552763819095 0.110528998076916
0.115577889447236 0.115523003041744
0.120603015075377 0.120447002351284
0.125628140703518 0.126034006476402
0.130653266331658 0.130581006407738
0.135678391959799 0.13528299331665
0.14070351758794 0.141231998801231
0.14572864321608 0.145697996020317
0.150753768844221 0.150969997048378
0.155778894472362 0.155575007200241
0.160804020100503 0.160367995500565
0.165829145728643 0.165763005614281
0.170854271356784 0.170485004782677
0.175879396984925 0.17611700296402
0.180904522613065 0.181208997964859
0.185929648241206 0.185962006449699
0.190954773869347 0.190619006752968
0.195979899497487 0.195852994918823
0.201005025125628 0.201316997408867
0.206030150753769 0.206623002886772
0.21105527638191 0.210951998829842
0.21608040201005 0.217043995857239
0.221105527638191 0.221276000142097
0.226130653266332 0.225920006632805
0.231155778894472 0.231003001332283
0.236180904522613 0.235863000154495
0.241206030150754 0.240735992789268
0.246231155778894 0.246416002511978
0.251256281407035 0.25067400932312
0.256281407035176 0.256992012262344
0.261306532663317 0.260437995195389
0.266331658291457 0.26663801074028
0.271356783919598 0.271082013845444
0.276381909547739 0.276946991682053
0.281407035175879 0.281661987304688
0.28643216080402 0.286944001913071
0.291457286432161 0.291566014289856
0.296482412060302 0.295917987823486
0.301507537688442 0.302298992872238
0.306532663316583 0.306903004646301
0.311557788944724 0.312215000391006
0.316582914572864 0.316760987043381
0.321608040201005 0.322117000818253
0.326633165829146 0.326727986335754
0.331658291457286 0.331770986318588
0.336683417085427 0.336035996675491
0.341708542713568 0.341531991958618
0.346733668341709 0.346792995929718
0.351758793969849 0.352032005786896
0.35678391959799 0.356748014688492
0.361809045226131 0.361571997404099
0.366834170854271 0.366979002952576
0.371859296482412 0.371890991926193
0.376884422110553 0.376127988100052
0.381909547738693 0.381303995847702
0.386934673366834 0.387629985809326
0.391959798994975 0.39205801486969
0.396984924623116 0.396948009729385
0.402010050251256 0.401463001966476
0.407035175879397 0.407231003046036
0.412060301507538 0.412741005420685
0.417085427135678 0.417239993810654
0.422110552763819 0.421595990657806
0.42713567839196 0.426674991846085
0.4321608040201 0.431542992591858
0.437185929648241 0.437034010887146
0.442211055276382 0.444548010826111
0.447236180904523 0.447198003530502
0.452261306532663 0.452098995447159
0.457286432160804 0.457607001066208
0.462311557788945 0.462864011526108
0.467336683417085 0.466847002506256
0.472361809045226 0.471668004989624
0.477386934673367 0.477607995271683
0.482412060301508 0.482632011175156
0.487437185929648 0.48725900053978
0.492462311557789 0.491851985454559
0.49748743718593 0.498537987470627
0.50251256281407 0.502318978309631
0.507537688442211 0.507507979869843
0.512562814070352 0.512062013149261
0.517587939698492 0.517536997795105
0.522613065326633 0.521763026714325
0.527638190954774 0.525709986686707
0.532663316582915 0.532480001449585
0.537688442211055 0.538226008415222
0.542713567839196 0.542511999607086
0.547738693467337 0.547688007354736
0.552763819095477 0.552758991718292
0.557788944723618 0.557837009429932
0.562814070351759 0.562798023223877
0.5678391959799 0.567534029483795
0.57286432160804 0.572420001029968
0.577889447236181 0.578365981578827
0.582914572864322 0.583292007446289
0.587939698492462 0.587853014469147
0.592964824120603 0.59246301651001
0.597989949748744 0.597986996173859
0.603015075376884 0.602995991706848
0.608040201005025 0.607797980308533
0.613065326633166 0.612389028072357
0.618090452261307 0.618166029453278
0.623115577889447 0.623313009738922
0.628140703517588 0.628880023956299
0.633165829145729 0.633419990539551
0.638190954773869 0.638629019260406
0.64321608040201 0.643607020378113
0.648241206030151 0.648402988910675
0.653266331658292 0.653432011604309
0.658291457286432 0.658010005950928
0.663316582914573 0.66312700510025
0.668341708542714 0.668703973293304
0.673366834170854 0.673276007175446
0.678391959798995 0.678179979324341
0.683417085427136 0.683570027351379
0.688442211055276 0.688834011554718
0.693467336683417 0.693163990974426
0.698492462311558 0.697705984115601
0.703517587939699 0.703553020954132
0.708542713567839 0.708823025226593
0.71356783919598 0.713602006435394
0.718592964824121 0.718954026699066
0.723618090452261 0.724066972732544
0.728643216080402 0.728720009326935
0.733668341708543 0.733478009700775
0.738693467336683 0.738667011260986
0.743718592964824 0.744231998920441
0.748743718592965 0.749153017997742
0.753768844221106 0.753584027290344
0.758793969849246 0.758668005466461
0.763819095477387 0.763836979866028
0.768844221105528 0.769068002700806
0.773869346733668 0.773617029190063
0.778894472361809 0.778855979442596
0.78391959798995 0.784614026546478
0.78894472361809 0.788832008838654
0.793969849246231 0.79407399892807
0.798994974874372 0.799365997314453
0.804020100502513 0.804545998573303
0.809045226130653 0.809885025024414
0.814070351758794 0.814229011535645
0.819095477386935 0.819387018680573
0.824120603015075 0.823745012283325
0.829145728643216 0.829355001449585
0.834170854271357 0.834237992763519
0.839195979899497 0.839120984077454
0.844221105527638 0.844552993774414
0.849246231155779 0.849229991436005
0.85427135678392 0.8541219830513
0.85929648241206 0.859686017036438
0.864321608040201 0.864194989204407
0.869346733668342 0.869852006435394
0.874371859296482 0.874338984489441
0.879396984924623 0.879221022129059
0.884422110552764 0.884464025497437
0.889447236180904 0.889482975006104
0.894472361809045 0.894273996353149
0.899497487437186 0.899607002735138
0.904522613065327 0.904895007610321
0.909547738693467 0.909686982631683
0.914572864321608 0.914273977279663
0.919597989949749 0.919179022312164
0.924623115577889 0.924632012844086
0.92964824120603 0.929942011833191
0.934673366834171 0.934660971164703
0.939698492462312 0.939338028430939
0.944723618090452 0.944907009601593
0.949748743718593 0.949353992938995
0.954773869346734 0.954697012901306
0.959798994974874 0.959846019744873
0.964824120603015 0.965153992176056
0.969849246231156 0.969902992248535
0.974874371859296 0.974569976329803
0.979899497487437 0.980090022087097
0.984924623115578 0.984899997711182
0.989949748743719 0.990073978900909
0.994974874371859 0.99497401714325
1 1
};
\addlegendentry{average}
\end{axis}

\end{tikzpicture}

%% file: tex/improvements/grid_gen_impls_subsec_split.tex
\subsection{Solution}
The shortcomings mentioned above could be addressed by providing a guarantee on fill rate value. Either "weak", where grids are guaranteed only to be neither empty nor full unless the explicitly requested by user, or "strong" where grids are additionally guaranteed to have exactly $round(s^2 * p_1)$ visible cells.

\begin{figure}[H]
  \centering
  \adjustbox{width=\textwidth, keepaspectratio}{
    \includegraphics{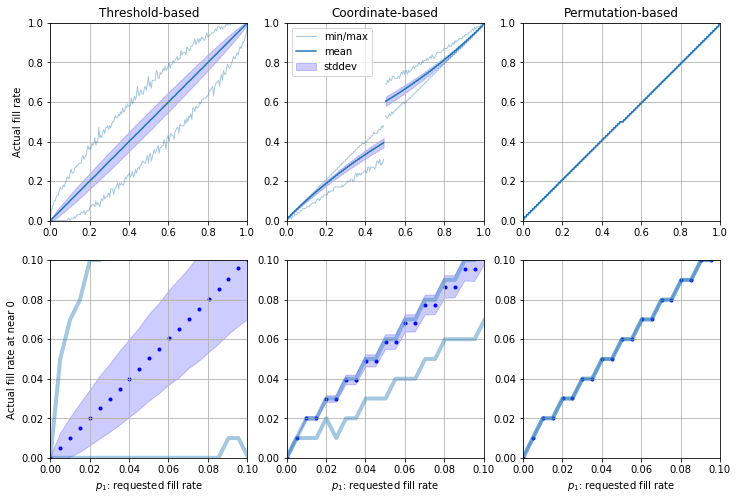}
  }
  \caption[Fill rate characteristics for grid generators with informativeness guarantee]{Fill rate visualization for the default grid generator of RISE (left) and proposed approaches (center and right), for grid containing 100 cells ($s=10$)}
  \label{fig:base_gens_fill_rate}
\end{figure}

\begin{figure}[H]
  \centering
  \adjustbox{width=\textwidth, keepaspectratio}{
    \includegraphics{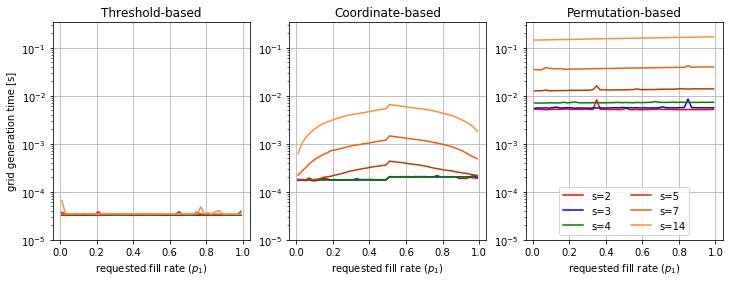}
  }
  \caption[Non-hybrid GridGen time performance]{Time performance of various grid generator implementations. Measurements were performed on GPU for batches of 1000 grids.}
  \label{fig:coord_perm_thr_cmp_perf}
\end{figure}

To evaluate the accuracy of these assumptions, two new grid generator implementations were created:
\begin{itemize}
  \item A \textbf{coordinate-based} solution which provides a "weak" guarantee by independently generating $ceil(s^2 * p_1)$ coordinate pairs and setting corresponding cells to visible. However, as $p_1$ increases, duplicates among coordinate pairs become increasingly frequent and fill rate cannot be maintained, as shown in \figref{fig:coord_and_smart_fill_rate}. To address this issue, for $p_1 > \frac{1}{2}$ the behaviour is inverted. The base grid is initially transparent, $floor(s^2 * (1-p_1))$ coordinates are generated and the corresponding cells are occluded.
  \begin{figure}[H]
    \centering
    \adjustbox{width=\textwidth, keepaspectratio}{
      \input{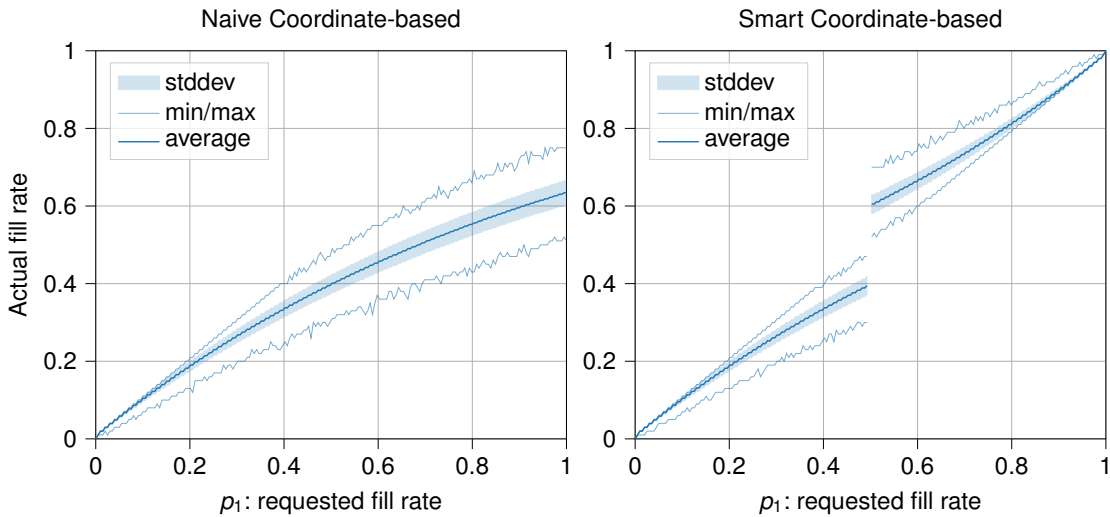}
    }
    \caption[Actual fill rate of coordinate-based grid generators]{Actual fill rates of the "naive" version of coordinate-based grid generator (left) and the "smart" version which inverts its behaviour for $p_1 > \frac{1}{2}$ (right)}
    \label{fig:coord_and_smart_fill_rate}
  \end{figure}
  \item A \textbf{permutation-based} solution which provides a "strong" guarantee by calculating a random permutation of the set of all grid cell coordinates and picking the first $ceil(s^2 * p_1)$ elements of the permuted set. This solution guarantees the exact same Fill Rate value for each grid (\figref{fig:base_gens_fill_rate}), but is significantly more computationally expensive than the coordinate-based approach, as shown by \figref{fig:coord_perm_thr_cmp_perf}.
\end{itemize}

\subsection{Further enhancements}
Because both of these guaranteed approaches are order-of-magnitude slower than the threshold-based algorithm, a hybrid solution was evaluated.
In the hybrid approach, one grid generator acts as a "base" and generates the initial set of grids, which is then checked for existence of uninformative grids. Next, a second generator is used to create replacements for all uninformative grids in the base set. This procedure is repeated until no further uninformative grids are found in the result.

\begin{figure}[H]
  \includegraphics[width=\textwidth]{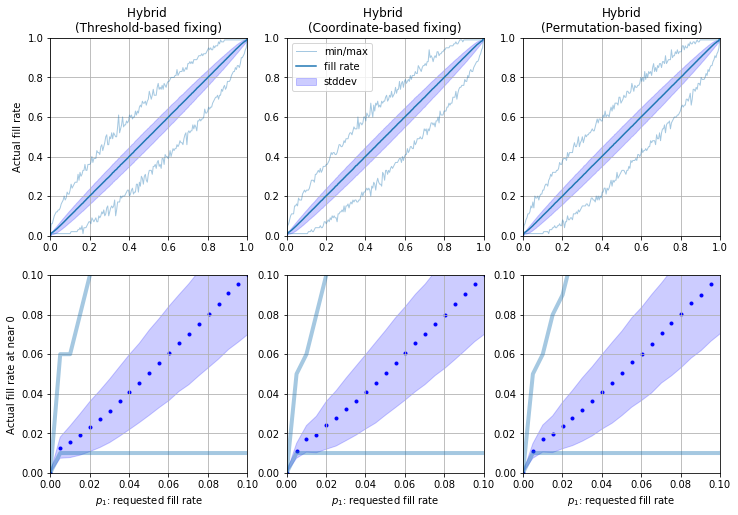}
  \caption[Fill rate comparison of Hybrid grid generators]{Fill rate comparison of Hybrid grid generators for grid of $s=10$.}
  \label{fig:hybrid_fill_rate}
\end{figure}

The resulting Fill Rate characteristic (\figref{fig:hybrid_fill_rate}) closely resembles that of the threshold-based approach (which was used as the base generator in all 3 examples). The most important difference is visible close to $p_1=0$ and $p_1=1$, where the lines marking extremes of Fill Rate value, respectively, never dip below $\frac{1}{s^2}$ or exceed $1 - \frac{1}{s^2}$, as opposed to unassisted TGG's fill rate characteristic presented in \figref{fig:base_gens_fill_rate})

\begin{figure}[H]
  \includegraphics[width=\textwidth]{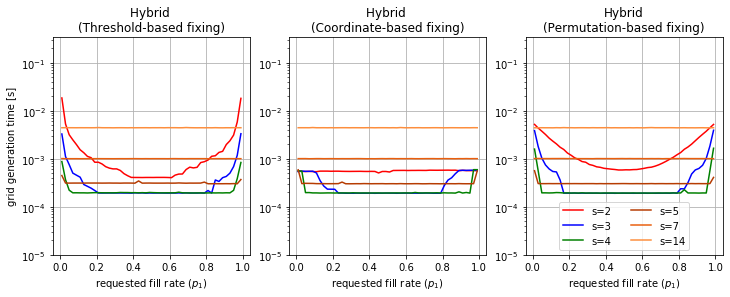}
  \caption[Performance comparison of Hybrid generators]{Performance comparison of various grid generator implementations in the role of a fixing generator in Hybrid approach.}
  \label{fig:hybrid_perf}
\end{figure}

\begin{figure}[H]
  \includegraphics[width=0.5\textwidth]{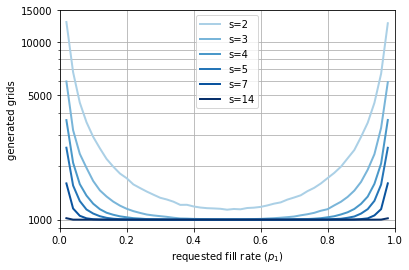}
  \caption{The expected number of random grids required to collect 1000 informative grids (if generated by TGG)}
  \label{fig:expected_n_grids}
\end{figure}

While the hybrid approach offers better performance than either of previously evaluated solutions, it has a flaw that becomes apparent upon closer inspection of performance measurements in \figref{fig:hybrid_perf} and \figref{fig:coord_perm_thr_cmp_perf}. Simply put, as the value of $s$ increases, the performance worsens. This pattern occurs regardless of choice of grid generators and therefore it must be caused by the logic of the hybrid generator itself. Specifically, the process of checking for uninformative grids. If we combine this observation with the fact that increasing $s$ exponentially decreases $P_U$ (\figref{fig:uninformative_prob}), it becomes clear that the informativeness checks are the costliest when they are needed the least.

\begin{figure}[H]
  \includegraphics[width=0.8\textwidth]{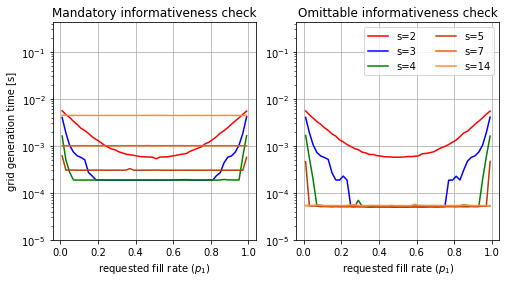}
  \caption[Performance comparison of Hybrid approach with mandatory and conditional informativeness check]{Comparison of performance of Hybrid approach with mandatory (left) and conditional informativeness checks (right).}
  \label{fig:hybrid_mandatory_perfcmp}
\end{figure}

This points at yet another enhancement opportunity: \textbf{conditional fixing}. An arbitrary, user-defined $P_U$ threshold will be used to determine whether the informativeness check, and subsequent grid replacement, should be carried out. Fixing is performed if~and~only~if $P_U$ exceeds this threshold. Otherwise, the hybrid generator immediately returns the output of the base generator. This mechanism lets the user set an upper limit on $P_U$ and control the trade-off between the cost of informativeness checks and risk of occurrence of uninformative masks. \figref{fig:hybrid_mandatory_perfcmp} demonstrates performance improvement brought on by this adjustment: large $s$ values no longer incur a costly overhead of the informativeness check, and the Hybrid Grid Generator performs on par with the base generator unless $P_U$ is high.

%% file: tbl/grid_gen_ref_conv.tex
\begin{table}[h]
    \begin{tabular}{ll|ccc}
    \toprule
      {} & {} &    \multicolumn{3}{c}{similarity metric}  \\
    s & $p_u$ &      {hog}  &   {spearman}        &      {ssim}  \\
    \midrule
    \multirow[c]{4}{*}{3} & 0.05 & 0.9387 &    0.9988 & 0.9999 \\
      & 0.30 & 0.9559 &    0.9994 & 0.9999 \\
      & 0.60 & 0.9749 &    0.9998 & 0.9998 \\
      & 0.90 & 0.9816 &    0.9998 & 0.9993 \\
    \midrule
    \multirow[c]{4}{*}{4} & 0.05 & 0.9120 &    0.9974 & 0.9999 \\
      & 0.30 & 0.9439 &    0.9986 & 0.9997 \\
      & 0.60 & 0.9548 &    0.9993 & 0.9996 \\
      & 0.90 & 0.9731 &    0.9996 & 0.9991 \\
      \midrule
    \multirow[c]{4}{*}{5} & 0.05 & 0.9248 &    0.9968 & 0.9998 \\
      & 0.30 & 0.9419 &    0.9974 & 0.9996 \\
      & 0.60 & 0.9495 &    0.9980 & 0.9991 \\
      & 0.90 & 0.9595 &    0.9983 & 0.9966 \\
      \midrule
    \multirow[c]{4}{*}{7} & 0.05 & 0.8791 &    0.9941 & 0.9992 \\
      & 0.30 & 0.8882 &    0.9952 & 0.9980 \\
      & 0.60 & 0.9169 &    0.9978 & 0.9945 \\
    \bottomrule
    \end{tabular}
  \caption[Intra-group similarity of reference maps]{Intra-group map similarities for a set of reference maps consisting of $N=50000$ masks.}
  \label{tbl:gridgen.ref_conv}
\end{table}

%% file: tbl/grid_gen_perf.tex
\begin{table}[h]
  \resizebox{\textwidth}{!}{
    \begin{tabular}{l|r|rr|rrr|rrr}
    \toprule
  	\multirow{2}{*}{s} & \multicolumn{3}{c|}{Base generators} & \multicolumn{3}{c|}{Hybrid (fixing) generators (threshold+...)} & \multicolumn{3}{c}{Hybrid with conditional fixing (threshold+...)} \\
    {}                 & threshold & coordinate & permutation &	threshold &  coordinate   &          permutation                & threshold      & coordinate       & permutation \\
    \midrule
      3                & 0.033 & \textbf{0.194} &   5.276     & 1.755     &         0.558 &            1.603                    &          1.798 &            0.557 &      1.608 \\
      7                & 0.032 & \textbf{0.197} &   5.701     & 0.414     &         0.299 &            0.508                    &          0.350 &            0.224 &      0.442 \\
      14               & 0.033 &         0.199  &   7.316     & 0.232     &         0.223 &            0.268                    & \textbf{0.109} &            0.099 &      0.142 \\
      28               & 0.033 &         0.299  &  13.524     & 0.308     &         0.310 &            0.308                    &          0.066 &   \textbf{0.071} &      0.071 \\
      56               & 0.034 &         0.904  &  37.636     & 1.006     &         1.006 &            1.010                    & \textbf{0.055} &            0.055 &      0.055 \\
      122              & 0.035 &         3.998  & 155.921     & 4.432     &         4.435 &            4.432                    &          0.056 &   \textbf{0.056} &      0.056 \\
    \bottomrule
    \end{tabular}
  }
  \caption[Grid generator performance comparison on GPU]{Comparison of grid generator performance on batch of 1000 masks, on GPU (values in milliseconds). Bold font indicates the best performing generator.}
\end{table}

\begin{table}[h]
  \resizebox{\textwidth}{!}{
    \begin{tabular}{l|r|rr|rrr|rrr}
    \toprule
  	 \multirow{2}{*}{s} & \multicolumn{3}{c|}{Base generators}   & \multicolumn{3}{c|}{Hybrid (fixing) generators (threshold+...)} & \multicolumn{3}{c}{Hybrid with conditional fixing (threshold+...)} \\
     {}                 & threshold & coordinate &  permutation  & threshold       &   coordinate      &        permutation        &	   threshold     &	       coordinate  &        permutation \\
    \midrule
      3                 &  0.050 & \textbf{0.147} &   5.217      &           0.876 &             0.317 &            1.345          &             0.868 &               0.317 &              1.364 \\
      7                 &  0.191 & \textbf{0.302} &   5.696      &           0.418 &             0.372 &            0.546          &             0.378 &               0.313 &              0.527 \\
      14                &  0.678 & \textbf{0.556} &   7.442      &           0.858 &             0.840 &            0.865          &             0.752 &               0.737 &              0.772 \\
      28                &  2.763 & \textbf{2.201} &  14.672      &           3.684 &             3.604 &            3.699          &             2.857 &               2.874 &              2.864 \\
      56                & 11.572 & \textbf{9.993} &  42.925      &          15.709 &            15.457 &           15.467          &            11.656 &              11.664 &             11.644 \\
      122               & 56.020 & \textbf{52.041} & 184.882     &          77.731 &            77.347 &           77.441          &            56.193 &              56.568 &             56.143 \\
    \bottomrule
    \end{tabular}
  }
  \caption[Grid generator performance comparison on CPU]{Comparison of grid generator performance on batch of 1000 masks, on CPU (values in milliseconds). Bold font indicates the best performing generator.}

\end{table}

%% file: tex/methodology/vrise_evaluation.tex
This section describes our goals and those parts of evaluation methodology common to all experiments. Experiments themselves are explained in detail in their respective sections in Chapter~\ref{ch:evaluation}.

One important piece of information to keep in mind while reviewing sections which discuss results is that the shape and quality of maps is dependent on the configuration of parameters. As sec.~\ref{ch:grid_search} will show, these changes are not arbitrary but gradual, following the changes in values of individual configuration parameters.

Our common convention in sections discussing results is that whenever performance \textit{improvement} for a certain configuration of parameters is mentioned, it is always expressed relative to a \textit{matching} RISE configuration, which shares values of $p_1$ and \texttt{polygons}, as these two parameters are common to both algorithms.
We also never calculate improvement between configurations whose values of $p_1$ and \texttt{polygons} don't match.

Both classifier models used in these evaluations - ResNet50~\cite{resnet50} and VGG16~\cite{vgg16} - were obtained from the PyTorch model repository and pre-trained on the ILSVRC2012 dataset. We have used these models as provided, without performing any additional fine-tuning.

Results are often grouped by the number of object instances per image. In these cases, we group results into three bins: "single" containing results for images with only a single object instance, "few" for images contains between 2 and 4 object instances (inclusive) and "many" when at least 5 objects are present. Because the ILSVRC2012 classification dataset does not mix object classes within a single image, there is no need to further subdivide them into single-class and multi-class groups.

\section{Quality metrics}
\label{sec:metrics}

Since one of our objectives is to compare the performance of our solution with results in the original paper, and because there is still no consensus on recommended metrics in the field \cite{sanity_metrics, sanityChecks}, we have decided to follow the choice made by authors of RISE~\cite{rise}.

We do not use ROAR~\cite{koar_roar} due to prohibitive cost of repeated model retraining required by this technique.

\subsection{Alteration Game}
Alteration Game is our alternative name for "Causal Metric" introduced in \cite{rise}.
The goal of Alteration Game is to measure quality of a saliency map by altering the image it has been generated for. Alterations gradually copy parts of the \textit{final}, fully altered image onto the \textit{initial} state of the explained image. Saliency values for individual pixels determine the order of alteration. The images are altered in steps, each one altering a batch of 224 most salient of remaining pixels.

\begin{itemize}
  \item The original Insertion Game copies pixels from the evaluated image onto its blurred copy. We have renamed this variant to \textit{Sharpen Game} to separate it from \textit{Blur Game}, which would gradually blur the original image.
  \item The original \textit{Deletion game} replaces pixels of the evaluated image with zeros. Its counterpart, \textit{Insertion game} (the name of which unfortunately collides with the original Insertion Game) copies pixels from the original image onto a blank canvas.
\end{itemize}

\begin{minipage}[l]{0.50\textwidth}

The four variants of Alteration Game can be grouped by two criteria:
\begin{itemize}
  \item \textit{Substrate}: the transformation used to produce the target state of the image.
  \item \textit{Entropy}: whether the original image is destroyed or reconstructed over the course of the game.
\end{itemize}

\end{minipage}
\hspace{0.05\textwidth}
\begin{minipage}[r]{0.40\textwidth}

  \resizebox{\textwidth}{!}{
    \begin{tabular}{c|cc}
      {} & \multicolumn{2}{c}{substrate} \\
      {entropy} & {zeros} & {blur} \\
      \midrule
      {constructive} & {Insert} & {\textbf{Sharpen}} \\
      {destructive} & {\textbf{Remove}} & {Blur} \\
    \end{tabular}
  }

\end{minipage}

\begin{figure}[H]
  \centering
  \includegraphics[width=\textwidth]{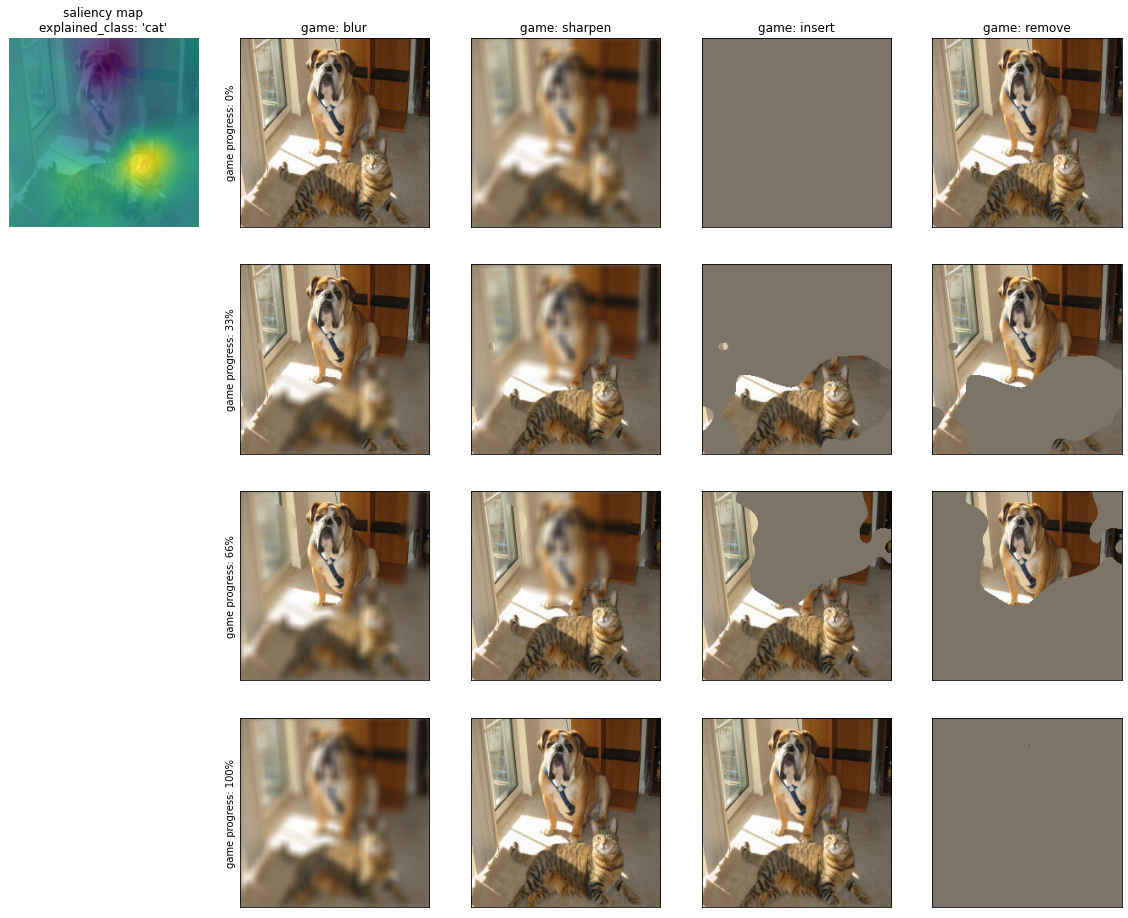}
  \caption[Demonstration of Alteration Game types]{Demonstration of Alteration Game types via 3-step examples}
  \label{fig:metrics.altgame_demo}
\end{figure}

The concept of Alteration Game is based on the assumption that the class confidence score is a direct consequence of presence (or lack) of salient features of that class in the input image. Therefore, if a saliency map accurately locates salient regions, removal of these areas should swiftly decrease the confidence score for observed class. Conversely, their reintroduction to the image should increase confidence. The more accurate the saliency map, the quicker this change should occur. The Area Under Curve (\textit{AUC}) of confidence score values gathered during the game quantifies the saliency map's quality.
In \textit{constructive} games, which restore the original image, the higher the score, the better. Conversely, in \textit{destructive} games, lower scores are better.

\subsection{Pointing Game}
Pointing Game, originally proposed by Zhang et al.~\cite{zhang2016topdown}, compares saliency maps against object instance bounding boxes. It returns a binary, "hit/miss" result.
A single, most salient pixel in a map is identified and its coordinates are compared against all bounding boxes annotating the instances of the observed class.
If this most salient point is located within boundaries of any of these bounding boxes, the map is considered accurate. Multiple hits, in case of bounding box overlaps, are ignored.

The core conceptual difference between Pointing and Alteration Game is that Pointing Game assesses whether the saliency map highlights those areas, which were deemed salient by the human annotator, whereas the Alteration Game checks only whether the map accurately identifies regions relevant to the classifier, whether they align with human intuition or not.

Using both Alteration Game and Pointing Game allows detection of cases in which the classifier had learned to detect certain classes in an effective, yet counterintuitive way.

\subsection{Consistency and convergence}
\label{sec:similarity_metrics}

\textit{Consistency} and \textit{convergence} use the Structural Similarity Index Metric (SSIM)~\cite{ssim} to gauge stability of results by measuring their visual similarity.

\textit{Consistency} measures variance of results at given $N_{masks}$ by calculating SSIM scores for all unique pairs (without repetition) of relevant maps. For $M$ maps in a set, the set of consistency results will contain $\frac{M!}{2(M-2)!}$ values.

\textit{Convergence} compares a selected saliency map to a reference map at a far more advanced stage of the evaluation process, providing information about the magnitude of visual differences between current and a "stable" state of the map.

Even though \textit{convergence} and \textit{consistency} might appear to be very similar, each is better suited to a different scenario.
\textit{Consistency} is more appropriate when there is no reliable reference map available, but the dataset contains multiple independently generated maps per parameter set. For instance, when experimental configurations perform multiple short runs, as is often the case in this work. \textit{Convergence}, on the other hand, is more suitable for monitoring the stability of a single result during ongoing map generation process when a "well-converged" map is available as a reference.

Similarity metrics were included for two reasons. First, some of the parameter configurations of RISE, in particular those using low values of $p_1$ and $s$ (\texttt{polygons}), tend to converge more slowly and exhibit greater inter-run variance.
Second, the use of randomly seeded Voronoi meshes, in place of regular square grids of RISE, is likely to further increase the variance of results.
Since we aim to improve the accuracy and precision of saliency maps while incurring as little negative impact on other qualities as possible, the addition of auxiliary metrics quantifying visual similarity of maps is necessary.

\subsection{Improvement metric}
\label{subsec:norm_diff}
The evaluation chapter often uses the relative difference (rel$\Delta$) between VRISE and RISE scores as a measure of performance improvement to allow aggregation of results across significant differences in magnitude. For instance, calculating average improvement in both constructive and destructive games would not be possible without a relative metric, because constructive game scores tend to fit in the range of 0.6-0.8, while destructive game often yields values between 0.05 and 0.15. An average of absolute differences would be skewed by improvement in constructive games.

In some cases, we rely on reference-adjusted relative difference, which, as opposed to relative difference, always yields values in range $[-1, 1]$, rather than $[-1, \frac{1-{ref}}{ref}]$. This property is useful when reference values are very close to 1.

\[
norm\Delta(X, R) =
  \begin{dcases}
    \frac{X - R}{max(1 - R, \epsilon)}, & {\text{if} X \geq R} \\
    \frac{X - R}{max(R, \epsilon)}, & {\text{otherwise}} \\
  \end{dcases}
\]
The value returned by this metric can be described as a fraction of maximum achievable improvement w.r.t. the reference value $R$, when $X \geq R$, or a fraction of maximum possible deterioration when $X < R$.

In cases where results for constructive and destructive games are aggregated together, the values of "minimizing" metrics (destructive variants of AltGame, specifically) where lower scores are better, are adjusted ($X' = 1 - X$) beforehand.

%% file: tex/methodology/blur_sigma_selection.tex
\section{Selecting blur strength}
\label{ch:blur_sigma}

\begin{minipage}[l]{0.50\textwidth}

In order to fairly compare our solution to RISE, we ought to eliminate all differences between the two algorithms, except the differences in shapes of the occlusions.
Blur is especially important, as it has a significant impact on result quality (as shown in sec.~\ref{ch:grid_search}). Therefore we need to make sure that VRISE's blur step is configured to be as similar to RISE's blur as possible for the purpose of comparison with results reported for the original algorithm.

\end{minipage}
\hspace{0.05\textwidth}
\begin{minipage}[l]{0.45\textwidth}
  \begin{figure}[H]
    \centering
    \includegraphics[width=0.8\textwidth]{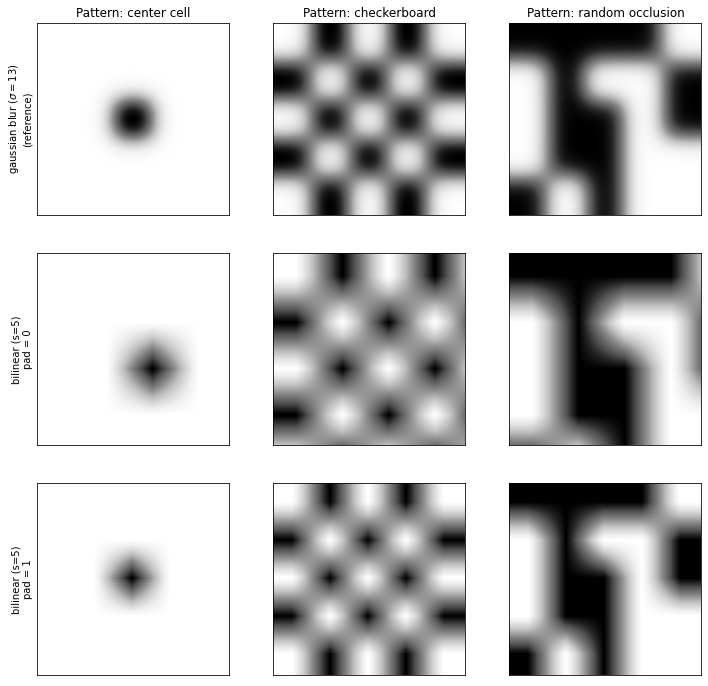}
    \caption[Comparison of normal and padded results of bilinear interpolation]{From top to bottom: Gaussian Blur, \mbox{Bilinear}, Padding+Bilinear}
    \label{fig:bilinear_alignment_example}
  \end{figure}

\end{minipage}

As mentioned in \chref{ch:vrise_description}, in the process of converting occlusion grids to masks, RISE uses a bilinear interpolation algorithm to enlarge $(s, s)$-shaped occlusion grids to the shape expected by the input of the inspected model. A welcome side-effect of this upsampling step is blurring of the original input, and conversion from binary values to floats in range $[0,1]$.
VRISE, however, renders occlusions in target resolution, making blurring via interpolation unsuitable. To preserve the properties introduced by blur, we chose to replace bilinear interpolation with Gaussian blur.

\sloppy The implementation we used, \texttt{torchvision.gaussian\_blur} gives the user control over the shape of the kernel and the value of standard deviation - $\sigma$. Therefore, we settled for reduction of blur-related parameters to one, emulating the convenient interface of \texttt{skimage.filters.gaussian} used in the early prototype.
\begin{itemize}
  \item Kernel aspect ratio has been set to 1:1, mimicking the aspect ratio of bilinear interpolation.
  \item \sloppy Kernel edge length has been bound to the value $\sigma$, using the following formula:
        \begin{equation}\text{kernel\_edge\_length} = 8\sigma + 1 - (\sigma \% 2)\label{eq:kernel_edge_length}\end{equation}
        which reflects the default sigma-to-kernel-size conversion performed by \texttt{skimage.filters.gaussian}.
  \item $\sigma$ was left as a configurable parameter as blur strength has a non-negligible effect on quality of results.
\end{itemize}

\subsection{Blur $\sigma$ matching experiment}

In the following experiment, we search for the most suitable value of $\sigma$, for given value of $s$, by measuring the similarity between an occlusion pattern upsampled via bilinear interpolation and its copies treated with Gaussian blur of varying strength.

\begin{figure}[H]
  \centering
  \includegraphics[width=\textwidth]{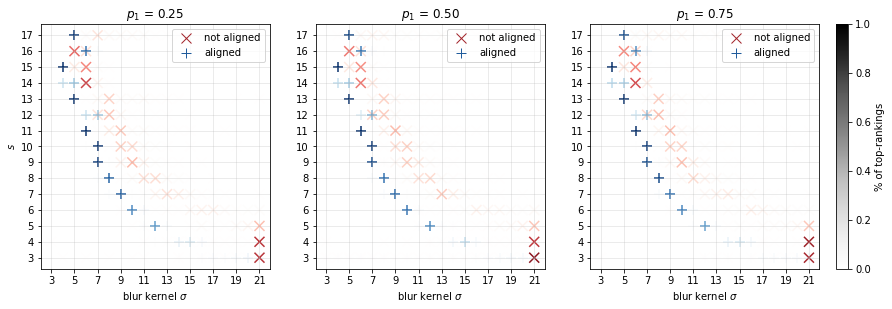}
  \caption[Comparison of top-ranking $\sigma$ values for aligned and unaligned masks]{Color intensity of datapoints corresponds to frequency of given $\sigma$ ranking as $1^{st}$ in terms of similarity to bilinear upsampling for given $s$}
  \label{fig:blur_sigma_rank_aligned_vs_unaligned}
\end{figure}

Prior to upsampling, each reference occlusion pattern has been asymmetrically padded (using the 'edge clone' method) along the bottom and right edge to align it with the mask rendered in native resolution.
\sloppy This extra step corrects the diagonal shift characteristic to \texttt{torch.nn.functional.interpolate} and reduces confusion of the~SSIM~metric~\cite{ssim} caused by misalignment of compared images. This method, while crude, visibly improves concentration of results, as shown in \figref{fig:blur_sigma_rank_aligned_vs_unaligned}. Without mask alignment, higher blur strength is favoured because it better compensates for the misalignment. The exact value varies from pattern to pattern, resulting in at least 2 comparably suitable $\sigma$ values for each $s$. However, when padding is applied, the results become more concentrated and also show that a much weaker blur setting results in greater similarity. \figref{fig:bilinear_alignment_example} demonstrates results of padding ("aligned" - padded, "not aligned" - without padding).

\begin{samepage}
  \FloatBarrier
  The experiment was carried out across a range of most commonly used $s$ values and equally spaced out $p_1$ values, to identify the best-fitting values of $\sigma$, using the following approach:
  \begin{enumerate}
    \item Generate an occlusion selector as a binary matrix of shape $(s, s)$, and set $p_1\frac{s^2}{2}$ cells to 1.
    \item Generate a reference mask:
    \begin{enumerate}
      \item Clone and pad the occlusion selector matrix along the bottom and right edge by cloning the outermost row and column, respectively.
      \item Resize the padded matrix of shape $(s+1, s+1)$ to $(224, 224)$ using bilinear interpolation.
    \end{enumerate}
    \item Generate a Gaussian mask:
    \begin{enumerate}
      \item Render the occlusion selector matrix as a 2D image of size $(224, 224)$
      \item Apply Gaussian blur with kernel configured according to equation \ref{eq:kernel_edge_length}
    \end{enumerate}
    \item Repeat above steps for each set of examined parameter values $(i, s_j, \sigma_k, p_{1_l})$
    \item Find the most suitable value of $\sigma$ for each $s$
    \begin{enumerate}
      \item calculate the value of SSIM metric between each Gaussian mask $(i, s_j, \mathbf{\sigma_x}, p_{1_l})$ and its corresponding reference mask $(i, s_j, p_{1_l})$
      \item identify the top-scoring $\sigma$ values for each set of $(s_j, p_{1_l})$, for all samples, using:
      \begin{itemize}
        \item highest mean value of SSIM across all samples
        \item highest frequency of top result
      \end{itemize}
    \end{enumerate}
  \end{enumerate}
\FloatBarrier
\end{samepage}

\input{tbl/methodology/blur_sigma_ssim_stats.tex}

\begin{samepage}
  \subsubsection{Results}
  Table \ref{tbl:blur_sigma_ssim_stats} shows the top three values of $\sigma$ for each value of $s$. Mean SSIM scores in the range of $(0.8, 0.9)$ confirm that the top-ranking $\sigma$ values produce results highly similar to the interpolated reference image, while low values of standard deviation confirm that the degree of similarity is consistent across samples. Consistency of rank across $p_1$ indicates that $\sigma$ values are robust and can be applied regardless of the desired fill rate.

  We have therefore identified a set of $(s, \sigma)$ pairs, which will be used in VRISE vs RISE evaluations.
\end{samepage}

%% file: tbl/methodology/blur_sigma_ssim_stats.tex
\begin{table}
  \begin{tabular}{l|l|rrr|rrr|rrr}
  \toprule
  {} & {metric} & \multicolumn{3}{c}{SSIM mean} & \multicolumn{3}{c}{SSIM stddev} & \multicolumn{3}{|c}{$\sigma$} \\
  {} & {rank} & \multicolumn{1}{c}{1} & \multicolumn{1}{c}{2} & \multicolumn{1}{c|}{3} & \multicolumn{1}{c}{1} & \multicolumn{1}{c}{2} & \multicolumn{1}{c}{3} & \multicolumn{1}{|c}{1} & \multicolumn{1}{c}{2} & \multicolumn{1}{c}{3} \\
  {$p_1$} & {$s$} & {} & {} & {} & {} & {} & {} & {} & {} & {} \\
  \midrule
  \multirow[c]{6}{*}{0.25} & 4 & 0.868 & 0.866 & 0.864 & 0.024 & 0.023 & 0.026 & 15 & 14 & 16 \\
   & 5 & 0.895 & 0.889 & 0.889 & 0.014 & 0.013 & 0.018 & 12 & 11 & 13 \\
   & 6 & 0.866 & 0.861 & 0.856 & 0.018 & 0.020 & 0.017 & 10 & 11 & 9 \\
   & 7 & 0.871 & 0.865 & 0.859 & 0.012 & 0.012 & 0.015 & 9 & 8 & 10 \\
   & 8 & 0.874 & 0.868 & 0.855 & 0.011 & 0.009 & 0.017 & 8 & 7 & 9 \\
   & 9 & 0.849 & 0.841 & 0.823 & 0.012 & 0.015 & 0.011 & 7 & 8 & 6 \\
  \midrule
  \multirow[c]{6}{*}{0.5} & 4 & 0.866 & 0.864 & 0.862 & 0.028 & 0.026 & 0.031 & 15 & 14 & 16 \\
   & 5 & 0.895 & 0.890 & 0.889 & 0.014 & 0.018 & 0.013 & 12 & 13 & 11 \\
   & 6 & 0.863 & 0.857 & 0.854 & 0.015 & 0.018 & 0.014 & 10 & 11 & 9 \\
   & 7 & 0.871 & 0.865 & 0.859 & 0.013 & 0.011 & 0.017 & 9 & 8 & 10 \\
   & 8 & 0.873 & 0.868 & 0.854 & 0.011 & 0.009 & 0.016 & 8 & 7 & 9 \\
   & 9 & 0.850 & 0.843 & 0.824 & 0.011 & 0.014 & 0.010 & 7 & 8 & 6 \\
  \midrule
  \multirow[c]{6}{*}{0.75} & 4 & 0.867 & 0.865 & 0.863 & 0.031 & 0.029 & 0.034 & 15 & 14 & 16 \\
   & 5 & 0.895 & 0.890 & 0.889 & 0.016 & 0.020 & 0.014 & 12 & 13 & 11 \\
   & 6 & 0.864 & 0.859 & 0.854 & 0.016 & 0.019 & 0.015 & 10 & 11 & 9 \\
   & 7 & 0.867 & 0.862 & 0.854 & 0.014 & 0.012 & 0.018 & 9 & 8 & 10 \\
   & 8 & 0.876 & 0.869 & 0.858 & 0.012 & 0.010 & 0.016 & 8 & 7 & 9 \\
   & 9 & 0.851 & 0.844 & 0.826 & 0.012 & 0.015 & 0.013 & 7 & 8 & 6 \\
   \bottomrule
  \end{tabular}
  \caption[Mean and standard deviation values of SSIM metric for top-ranking values of blur $\sigma$]{Mean and standard deviation values of SSIM metric for top-ranking values of $\sigma$ (sample size = 100)}
  \label{tbl:blur_sigma_ssim_stats}
\end{table}

%% file: tex/methodology/fp16.tex
\section{FP16 mode}
\label{ch:fp16_eval}

\begin{minipage}[l]{0.6\textwidth}

Building a saliency map via random input sampling requires many forward passes through the classifier, making evaluation of occlusion patterns the most computationally expensive step of both RISE and VRISE.
Due to sheer amount of computation required to carry out the experiments described in the evaluation chapter, we had to look for ways of accelerating the map generation process. One of the most effective approaches, and also by far the easiest to implement, was the reduction of numerical precision during mask evaluation. Although this step carries a risk of adversely impacting the quality of results, an opportunity to nearly halve the execution time, as well as the archive size, was a very good incentive to implement it and properly examine its influence on our use-cases.

\end{minipage}
\hspace{0.05\textwidth}
\begin{minipage}[r]{0.3\textwidth}
    \raggedright\footnotesize{\textbf{Runtime configuration}}
  \resizebox{\textwidth}{!}{
    \begin{tabular}{l|r}
      \toprule
      {parameter} & {value} \\
      \midrule
      $N_{masks}$ & 3000 \\
      \texttt{meshcount} & 1000 \\
      $p_1$ & 0.5 \\
      \texttt{polygons} & 49 \\
      blur $\sigma$ & 9.0 \\
      \midrule
      \midrule
      {model} & {ResNet50} \\
      {runs} & {5} \\
      {\#images} & {500} \\
      \bottomrule
    \end{tabular}
  }
\end{minipage}

Use of FP16 mode in model training and inference has already been examined by \cite{fp16training} and was shown to have negligible impact on prediction accuracy, while reducing inference time by up to 50\%.
Out primary concern related to the use of FP16 mode in VRISE evaluation was the cumulative effect of numerical errors compounded during saliency map generation (specifically, numerical discrepancies in confidence scores for observed classes) and map quality evaluation if Alteration Game were ran in FP16 mode.

\begin{samepage}
  We have evaluated the impact of using 16-bit floats on 3 stages of our evaluation pipeline:
  \begin{itemize}
    \item inference during saliency map generation
    \item result storage
    \item inference during Alteration Game
  \end{itemize}
\end{samepage}

By repeating the evaluation for each combination of precision modes in these three stages, we have obtained 8 result sets in total.
The configuration using FP32 mode across all steps was used as reference.

\subsection{Methodology}
A single set of 3000 masks based on 1000 Voronoi meshes (3 masks per mesh) was generated, applied to the input images and passed through the classifier to obtain two sets of confidence scores - one for the classifier in FP16 mode and the other for FP32 mode.
The evaluation dataset was composed of the first 500 images from ILSVRC2012 validation split.
By using the same set of masks for evaluations in both numerical precision modes, we ensure that the difference between them is a consequence of the selected numerical precision alone.
Map composition (weighting occlusion patterns with their respective confidence scores) was carried out in FP32 mode for two reasons. First, to prevent additional compounding of errors during mask aggregation. Second, mask aggregation is a relatively "cheap" operation and reducing its numerical precision would not yield any noticeable performance improvement.
This procedure was repeated 5 times to gather a larger sample.
Both saliency map sets were then stored in two copies - one using an FP32 storage archive and the other, FP16.
Finally, AltGame was ran twice on each of the four archives, one time for each of the two precision modes, yielding the final 8 sets of results.
All steps of the algorithm which were not included in the scope of this evaluation, were carried out in FP32 mode. For instance, if the step under evaluation returned FP16 values, then these would be cast back to FP32 afterwards.
Pointing game was ran on each of the 4 map sets.

The $r\Delta$ metric used in \tblref{tbl:fp16.total_stats} is calculated as relative difference between a single AuC value and its corresponding reference AuC from the all-FP32 set.

\subsection{Results and conclusions}

\FloatBarrier

\begin{samepage}
  Based on \tblref{tbl:fp16.total_stats}, we can make the following observations about impact of precision mode on the results:
  \begin{itemize}
    \item mean(r$\Delta$) barely exceeds $0.01\%$, confirming that FP16 mode is safe to use for large scale aggregations, where outliers have little impact on the final result.
    \item The quality of 98\% of results deviates by no more than $\pm5\%$ from the quality of their respective reference maps.
    \item The impact of running AltGame in FP16 mode is negligible, so FP16 can be considered safe to use in all cases.
    \item Using FP16 in storage has a noticeable impact on the distribution of $r\Delta$, compared to its effect in AltGame. Although "compression" of maps to FP16 appears to have most severe effect
      \begin{itemize}
        \item Use of FP16 during map generation increases the average $r\Delta$ from 0 to $\sim0.45\%$ and standard deviation to $\sim1\%$
        \item In comparison, FP16 storage alone increases the average $r\Delta$ to $\sim0.7\%$ and standard deviation to $\sim1.4\%$.
        \item However, FP16 storage error compounds with FP16 map error to a very limited degree. Hence, if FP16 storage has to be used, then there's no reason to not speed up the computation by using FP16 for other evaluation steps as well.
      \end{itemize}
    \item However, due to high values of $min(r\Delta)$ and $max(r\Delta)$, evaluations carried out on a handful of images should use FP32 mode for map generation and storage. If a performance boost is needed, a preliminary evaluation of r$\Delta$ for FP16 saliency maps of these specific images could be used to determine whether it's safe to carry out the entire experiment in reduced precision mode.
  \end{itemize}
\end{samepage}
\tblref{tbl:fp16.auc_variance_and_pg} shows that use of FP16 mode has a negligible impact on the inter-run consistency AuC scores, whereas pointing game results are only affected by reduction of storage precision. Similarity metrics in \tblref{tbl:fp16.similarity} show that visual similarity is not affected by reduced numerical precision.

\FloatBarrier

\input{tbl/methodology/fp16/tbl_total_stat.tex}

\begin{table}
  \caption[Comparison of raw metric values for FP16 and FP32 modes.]{Comparison of raw metric values for FP16 and FP32 modes.}
  \label{tbl:fp16.auc_variance_and_pg}
  \begin{minipage}[r]{0.5\textwidth}
    \vspace{2mm}
    \raggedright\footnotesize{Average inter-run standard deviation of AuC for various numerical precision modes. $N_{runs}=5$}
    \resizebox{\textwidth}{!}{
      \input{tbl/methodology/fp16/tbl_variance.tex}
    }
  \end{minipage}
  \hspace{0.05\textwidth}
  \begin{minipage}[r]{0.4\textwidth}
    \raggedright\footnotesize{Pointing game accuracy. $N_{maps}=7500$}
    \resizebox{\textwidth}{!}{
      \input{tbl/methodology/fp16/tbl_pointing_game.tex}
    }
  \end{minipage}
\end{table}

\input{tbl/methodology/fp16/tbl_similarity.tex}

%% file: tbl/methodology/fp16/tbl_total_stat.tex
\begin{table}[h]
\caption[Relative $\Delta(FP32, FP16)$ stats aggregated over all game types]{Relative $\Delta(FP32, FP16)$ stats aggregated over all game types, sorted by mean(|r$\Delta$|), N=80000. Results for the most frequently used precision settings are marked in bold}
\label{tbl:fp16.total_stats}
\resizebox{\textwidth}{!}{
  \begin{tabular}{ccc|rrr|rrrr}
  \toprule
  {} & {} & {metric} & {mean(r$\Delta$)} & {mean(|r$\Delta$|)} & {stddev(r$\Delta$)} & {min(r$\Delta$)} & {$P_{1\%}$(r$\Delta$)} & {$P_{99\%}$(r$\Delta$)} & {max(r$\Delta$)} \\
  {maps} & {storage} & {game} & {} & {} & {} & {} & {} & {} & {} \\
  \midrule
  \multirow[c]{2}{*}{FP32} & \multirow[c]{2}{*}{FP32} & FP32 & 0.000\% & 0.000\% & 0.000\% & 0.000\% & 0.000\% & 0.000\% & 0.000\% \\
   &  & \textit{FP16} & -0.001\% & 0.175\% & 0.296\% & -3.530\% & -0.871\% & 0.806\% & 3.239\% \\
  \midrule
  \multirow[c]{2}{*}{\textit{FP16}} & \multirow[c]{2}{*}{FP32} & FP32 & -0.004\% & 0.420\% & 0.898\% & -25.814\% & -2.536\% & 2.507\% & 11.405\% \\
   &  & \textit{FP16} & \textbf{-0.004\%} & \textbf{0.466\%} & \textbf{0.947\%} & \textbf{-25.433\%} & \textbf{-2.717\%} & \textbf{2.702\%} & \textbf{12.059\%} \\
  \midrule
  \multirow[c]{2}{*}{FP32} & \multirow[c]{2}{*}{\textit{FP16}} & FP32 & -0.002\% & 0.686\% & 1.409\% & -44.834\% & -3.630\% & 3.638\% & 35.308\% \\
   &  & \textit{FP16} & -0.003\% & 0.718\% & 1.437\% & -44.887\% & -3.737\% & 3.771\% & 35.184\% \\
  \midrule
  \multirow[c]{2}{*}{\textit{FP16}} & \multirow[c]{2}{*}{\textit{FP16}} & FP32 & 0.011\% & 0.725\% & 1.453\% & -25.234\% & -3.939\% & 3.983\% & 37.644\% \\
   &  & \textit{FP16} & \textbf{0.011\%} & \textbf{0.756\%} & \textbf{1.475\%} & \textbf{-25.516\%} & \textbf{-3.985\%} & \textbf{4.113\%} & \textbf{35.018\%} \\
  \bottomrule
  \end{tabular}
  }
\end{table}

%% file: tbl/methodology/fp16/tbl_variance.tex
\begin{tabular}{ccc|rr}
\toprule
{} & {} & {} & \multicolumn{2}{c}{Inter-run AuC} \\
{map} & {storage} & {game} & {stddev} & {$\Delta$stddev} \\
\midrule
\multirow[c]{4}{*}{FP32} & \multirow[c]{2}{*}{FP32} & FP32 & 0.027243 & 0.000000 \\
 &  & FP16 & 0.027240 & -0.000003 \\
 \cmidrule[.5pt]{2-5}
 & \multirow[c]{2}{*}{FP16} & FP32 & 0.027230 & -0.000013 \\
 &  & FP16 & 0.027227 & -0.000016 \\
 \midrule
\multirow[c]{4}{*}{FP16} & \multirow[c]{2}{*}{FP32} & FP32 & 0.027224 & -0.000020 \\
 &  & FP16 & 0.027220 & -0.000023 \\
 \cmidrule[.5pt]{2-5}
 & \multirow[c]{2}{*}{FP16} & FP32 & 0.027201 & -0.000042 \\
 &  & FP16 & 0.027199 & -0.000044 \\
\bottomrule
\end{tabular}

%% file: tbl/methodology/fp16/tbl_pointing_game.tex
\begin{tabular}{cc|c}
\toprule
{maps} & {storage} & {mean PG accuracy} \\
\midrule
\multirow[c]{2}{*}{FP16} & FP16 & {87.2\% $\pm$ 2.0\%} \\
 & FP32 & {86.9\% $\pm$ 2.2\%} \\
 \midrule
\multirow[c]{2}{*}{FP32} & FP16 & {87.2\% $\pm$ 1.9\%} \\
 & FP32 & {86.9\% $\pm$ 2.2\%} \\
\bottomrule
\end{tabular}

%% file: tbl/methodology/fp16/tbl_similarity.tex
\begin{table}
\caption[Mean similarity values between mixed precision maps and all-FP32 maps.]{Mean similarity between mixed precision maps and corresponding all-FP32 maps. N=2500}
\label{tbl:fp16.similarity}
\begin{tabular}{cc|cccc}
\toprule
{} & {} & \multicolumn{4}{c}{similarity metric (mean)} \\
{maps} & {storage} &  {pearson} & {pearson(HOG)} & {spearman} & {SSIM} \\
\midrule
\multirow[c]{2}{*}{FP16} & FP16 & 99.997\% & 99.581\% & 99.995\% & 99.999\% \\
 & FP32 & 99.999\% & 99.982\% & 99.999\% & 99.999\% \\
 \midrule
FP32 & FP16 & 100.000\% & 100.000\% & 100.000\% & 100.000\% \\
\bottomrule
\end{tabular}
\end{table}

%% file: tex/evaluation/_intro.tex
\begin{minipage}[l]{0.45\textwidth}
We assess the performance of our proposed solution in three different evaluations, each striking a different balance between the breadth of the parameter space and the size of image dataset. The "GridSearch" experiment evaluates a broad range of parameter sets against two particular images. "FixedSigma" targets 8 parameter sets and evaluates them against 200 randomly selected images from ImageNet. Finally, "ImageNet" evaluates a single configuration on all 50k images of the ILSVRC2012 validation split \cite{ilsvrc2012} using two different classifiers and fully replicating the setup used in the original RISE article \cite{rise}.

The majority of experiments' saliency map generation workload was performed on DGX Station's Tesla V100 GPUs.

\end{minipage}
\hspace{0.05\textwidth}
\begin{minipage}[r]{0.5\textwidth}
    \raggedright\footnotesize{\textbf{Quantitative summary of performed evaluations}}
    \resizebox{\textwidth}{!}{
      \begin{tabular}{c|cccc}
        \toprule
        \multirow[c]{2}{*}{evaluation} & \multirow[c]{2}{*}{GS} & \multirow[c]{2}{*}{F$\sigma$} & {ImageNet} & {ImageNet} \\
        {} & {} & {} & {ResNet50} & {VGG16} \\
        \midrule
        {paramsets} & {2352} & {8} & {1} & {1} \\
        {$N_{mask}$} & {5000} & {4000} & {8000} & {4000}  \\
        {VRISE runs} & {5} & {5} & {3} & {3} \\
        {RISE runs} & {17} & {7} & {3} & {3} \\
        {images} & {2} & {200} & {50000} & {50000} \\
        {SMaps} & {551K} & {192K} & {300K} & {300K} \\
        {\#fwdpass} & {1835M} & {786M} & {2400M} & {1200M} \\
        {filesize} & {72GB} & {38GB} & {67GB} & {67GB} \\
        \bottomrule
      \end{tabular}
    }
    \vspace{2mm}

    \begin{footnotesize}
      \begin{itemize}
        \item \textit{GS}: GridSearch (section~\ref{ch:grid_search})
        \item $F\sigma$: FixedSigma / Fixed $\sigma$ (section~\ref{sec:targeted_s})
        \item ImageNet (section~\ref{sec:imgnet})
      \end{itemize}
    \end{footnotesize}

\end{minipage}

%% file: tex/evaluation/grid_search.tex
\section{Exploration of parameter space ("GridSearch")}
\label{ch:grid_search}

\begin{minipage}[l]{0.45\textwidth}

The evaluation presented in this section aims to provide a better understanding of the relationships between VRISE parameters and improvement of map quality (relative to RISE), as well as identify cases where the performance of both algorithms differs the most. For the purposes of this chapter, a Cartesian product of parameter values across all four configurable dimensions (2352 distinct parameter sets in total) has been evaluated on two distinct images representing single- and multi-instance classes of inputs.

\end{minipage}
\hspace{0.05\textwidth}
\begin{minipage}[r]{0.45\textwidth}
  \raggedright\footnotesize{\textbf{GridSearch - parameter values}}
    \begin{tabular}{c|c}
      \toprule
      {parameter} & {included values} \\
      \midrule
      \midrule
      {blur $\sigma$} & {0.1, 3, 6, 9, 12, 15, 18} \\
      \midrule
      \multirow{2}{*}{\texttt{meshcount}} & {1, 2, 5, 10, 25, 50, 100,} \\
      {} & {250, 500, 1000, 2500, 5000} \\
      \midrule
      {\texttt{polygons}} & {16, 25, 49, 81} \\
      \midrule
      \multirow[c]{2}{*}{$p_1$} & {0.125, 0.25, 0.375, 0.5} \\
      {} & {0.625, 0.75, 0.875} \\
      \midrule
      \midrule
      {classifier} & {ResNet50} \\
      \midrule
      {SMap Gen mode} & {FP16} \\
      {Storage mode} & {FP32} \\
      {AltGame mode} & {FP16} \\
      \midrule
      {Occlusion Selector} & {HybridGridGen} \\
      \bottomrule
    \end{tabular}
\end{minipage}

\input{tbl/grid_search/grid_search_param_variance.tex}

\begin{figure}[H]
  \centering
  \includegraphics[width=0.65\textwidth]{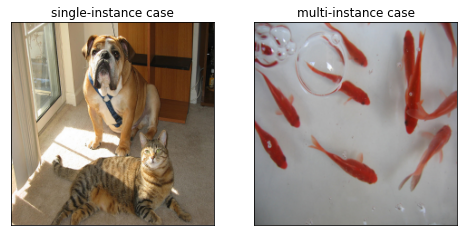}
  \caption[Input images used in the section "Exploration of parameter space"]{Input images used for evaluations in this section.}
  \label{fig:grid_search.input_images}
\end{figure}

\subsection{Methodology}
\begin{itemize}
  \item The data was gathered over 5 independent runs of VRISE, and 17 runs of RISE. Larger sample size for RISE improves the reliability of our reference point. Moreover, since the two most densely populated parameter dimensions (\texttt{meshcount} and blur $\sigma$) don't apply to RISE, significantly shrinking the number of reference parameter sets, and therefore, execution time.
  \item The state of each saliency map was captured at specific points during the map generation process in order to capture changes in map quality over time.
  \item Only two images were used for this evaluation (fig. \ref{fig:grid_search.input_images}) - one containing a single instance of each of observed classes, and the other containing multiple instances. This choice was motivated by significant differences in performance between these two types of input, observed in our initial assessments. The number of examples of each input type could not be increased due to time constraints.
  \item Both RISE and VRISE use an occlusion selector with informativeness guarantee (see section \ref{sec:guaranteed_grid_gens}) to increase fairness of comparison for paramsets where $p_1$ and $N_p$/$s$ are low.
  \item Relative difference of AuC scores ($r\Delta(AuC)$) is negated for destructive games (for which lower scores are better) so that positive values of $r\Delta(AuC)$ always express improvement in map quality. This approach allows us to apply aggregations over all game types.
  \item All plots show results for saliency maps consisting of $N_{max}$ occlusion masks, except for plots using $N_{masks}$ as the X-axis.
\end{itemize}

\subsection{Overview}
Figures in this section show both raw quality scores and \textit{improvement} (or lack thereof) expressed as relative difference between VRISE and RISE scores. Because each VRISE parameter must take some value, none of them could be measured independently from all others and therefore all plots (except for heatmaps) have to visualise aggregations over results for all parameter dimensions which were not granted an axis on that specific plot.

\begin{figure}[H]
  \centering
  \includegraphics[width=0.9\textwidth]{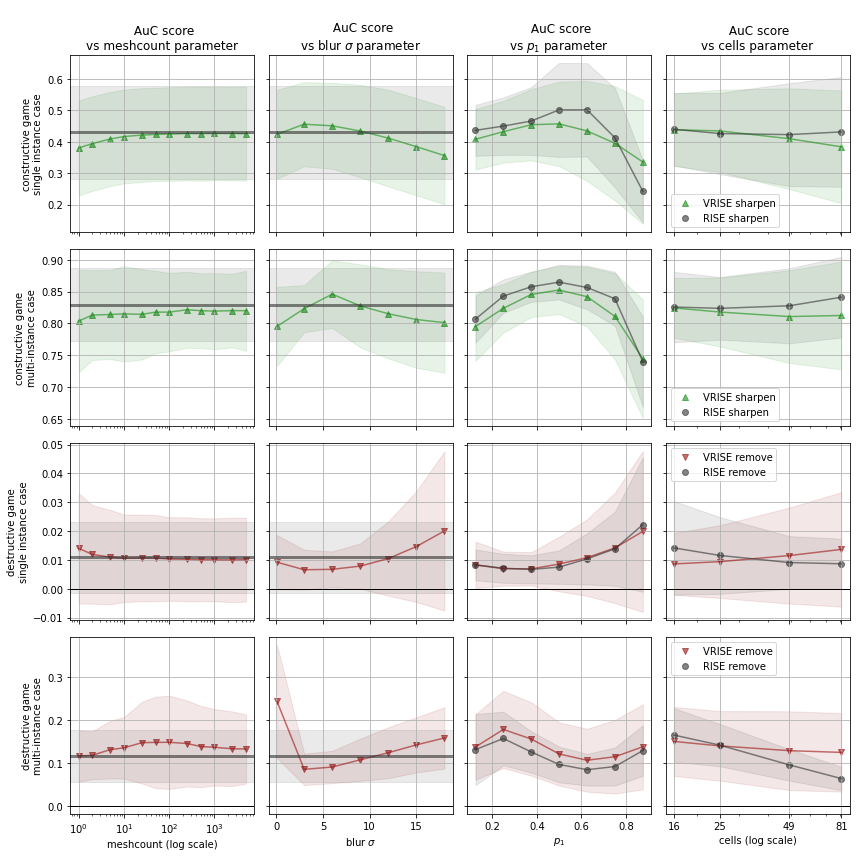}
  \caption[Relationships between parameter values and map quality as measured by various types of AltGame]{Plots of rel$\Delta$(AuC) for each parameter dimension and type of AltGame. Each data point is a mean rel$\Delta$ of all results involving it's x-axis parameter value. Odd rows contain results for single-instance case, while even rows, for multi-instance case.}
  \label{fig:grid_search.rawauc_origame}
\end{figure}

We use the following visualisation layouts:
\begin{itemize}
  \item 1D line plots (e.g. fig.~\ref{fig:grid_search.rawauc_origame}) prioritise comparison of trends between single- and multi-instance cases. Only one parameter dimension is shown per plot and each data point aggregates all results along all other VRISE parameter dimensions. Subplots for blur $\sigma$ and $\texttt{meshcount}$ show RISE score as a horizontal line because these two parameters are not used by RISE.
  \item Line plots arranged in a 2D grid layout (e.g. fig.~\ref{fig:grid_search.goldfish_pg_grid}), prioritise visibility of interactions between pairs of parameters. These plots are orthogonal 2D projections of 3D plots, using data series and colour saturation to visualise the $3^{rd}$ dimension. Results for each pair of parameter dimensions are presented from two perspectives. If we assume the plot at $(i, j)$ in the upper half of the grid is the "side view" of the data (looking down the Z-axis towards the origin point), then its counterpart at $(j, i)$ in the lower half displays a "frontal" projection, looking down the X-axis of plot $(i, j)$. The collapsed Z-axis of $(i, j)$ becomes the X-axis of $(j, i)$ and vice-versa. Y-axis is common for both plots. The results should be interpreted as "when (V)RISE is configured with X-param and Series-param, the average quality score is equal to Y"
  \item Heatmaps (e.g. fig.~\ref{fig:grid_search.goldfish_aucdiff_heatmap}) prioritise disaggregation of results, to serve as a reference for other types of plots. 2D heatmaps laid out on a 2D grid allow us to visualise results for each unique parameter set separately. Unless specified otherwise, the only aggregation performed is result averaging over runs.
\end{itemize}

In figures~\ref{fig:grid_search.catdog_aucdiff_heatmap_checkpoints}~and \ref{fig:grid_search.goldfish_aucdiff_heatmap_checkpoints}, results were aggregated over \texttt{meshcount} to make space for the $N_{masks}$ dimension.
Here, \texttt{meshcount} has been collapsed, because it has the smallest impact on the standard deviation of r$\Delta(AuC)$ out of all other parameters (as shown in \tblref{tbl:grid_search.param_stddev}).

\begin{figure}[H]
  \centering
  \includegraphics[width=0.90\textwidth]{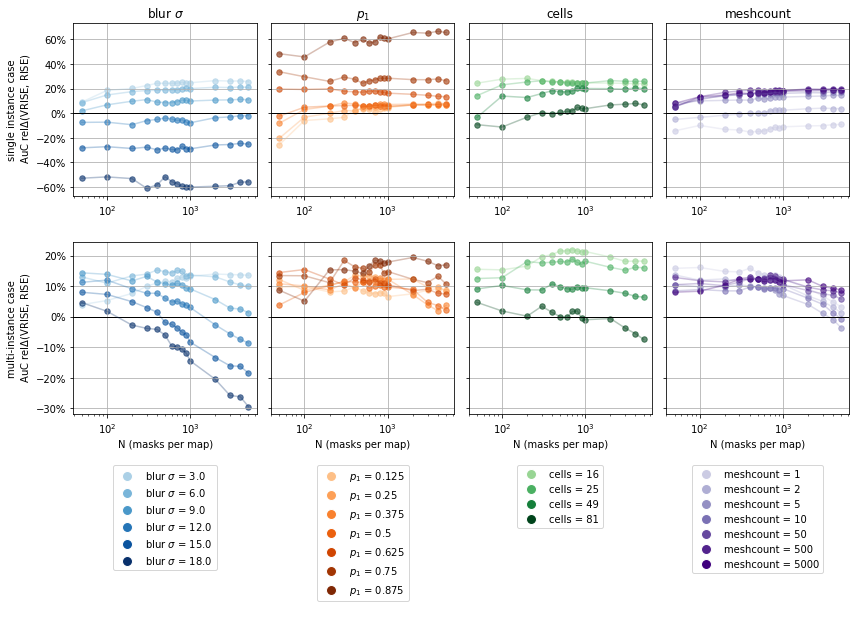}
  \caption[Relative map quality measured by rel$\Delta$ of AltGame scores vs number of masks per map]{Average map quality improvement measured by rel$\Delta$ of AltGame AuC. Values of blur $\sigma$ and \texttt{meshcount} were \textbf{narrowed down} to blur $\sigma \in [3, 6, 9]$ and $\texttt{meshcount} > 100$ in plots using \texttt{polygons} and $p_1$ as series.}
  \label{fig:grid_search.checkpoints_auc_reldiff_narrowed}
\end{figure}

\begin{figure}[H]
  \centering
  \includegraphics[width=0.90\textwidth]{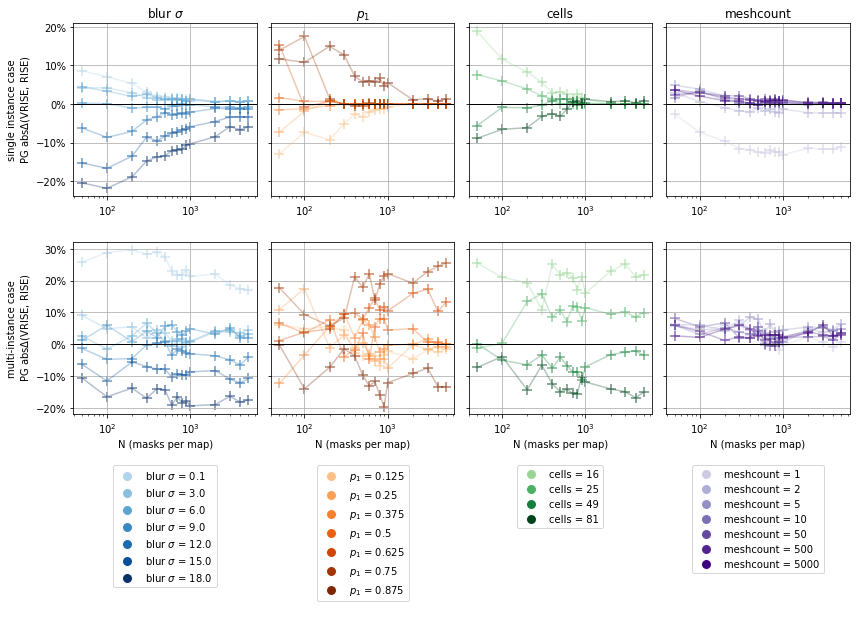}
  \caption[Relative map quality measured by abs$\Delta$ of Pointing Game vs number of masks per map]{Average map quality improvement measured by abs$\Delta$ of Pointing Game accuracy. Values of blur $\sigma$ and \texttt{meshcount} were \textbf{narrowed down} to blur $\sigma \in [3, 6, 9]$ and $\texttt{meshcount} > 100$ in plots using \texttt{polygons} and $p_1$ as series.}
  \label{fig:grid_search.checkpoints_pg_absdiff_narrowed}
\end{figure}

In figures~\ref{fig:grid_search.checkpoints_auc_reldiff_narrowed} and \ref{fig:grid_search.checkpoints_pg_absdiff_narrowed}, the slope of the data series indicates how quickly the quality gap between VRISE and RISE changes as the number of evaluations increases along the X-axis of each sub-plot.
Positive slope indicates that VRISE maps improve faster than RISE's and vice-versa. Zero slope indicates that quality of both maps improves at the same pace.

\begin{figure}[H]
  \centering
  \includegraphics[width=\textwidth]{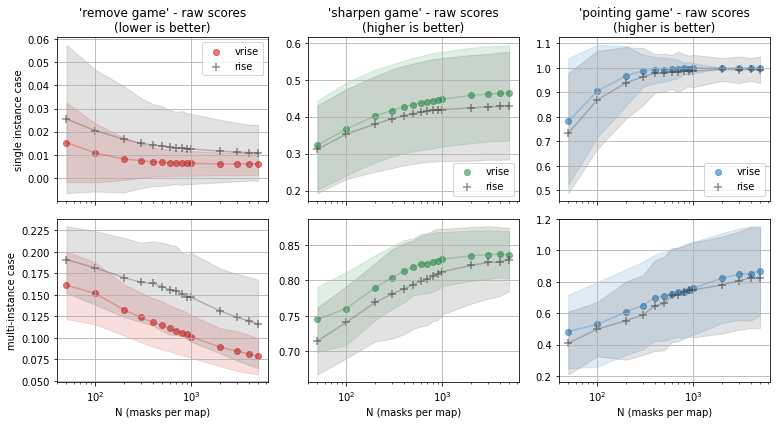}
  \caption[Raw scores of quality metrics, plotted against $N_{masks}$, averaged over all parameter sets. Low-blur subset.]{Raw scores of quality metrics, plotted against $N_{masks}$, averaged over all matching parameter sets. \textbf{Results for detrimental values of VRISE-specific params were excluded}. Notice that VRISE now scores better than RISE in all cases. High stddev is a consequence of aggregation over results for all RISE-specific ($p_1$, \texttt{polygons}) parameters pairs.}
  \label{fig:grid_search.rawscores_vs_N_better_subset}
\end{figure}

\begin{figure}[H]
  \centering
  \includegraphics[width=0.9\textwidth]{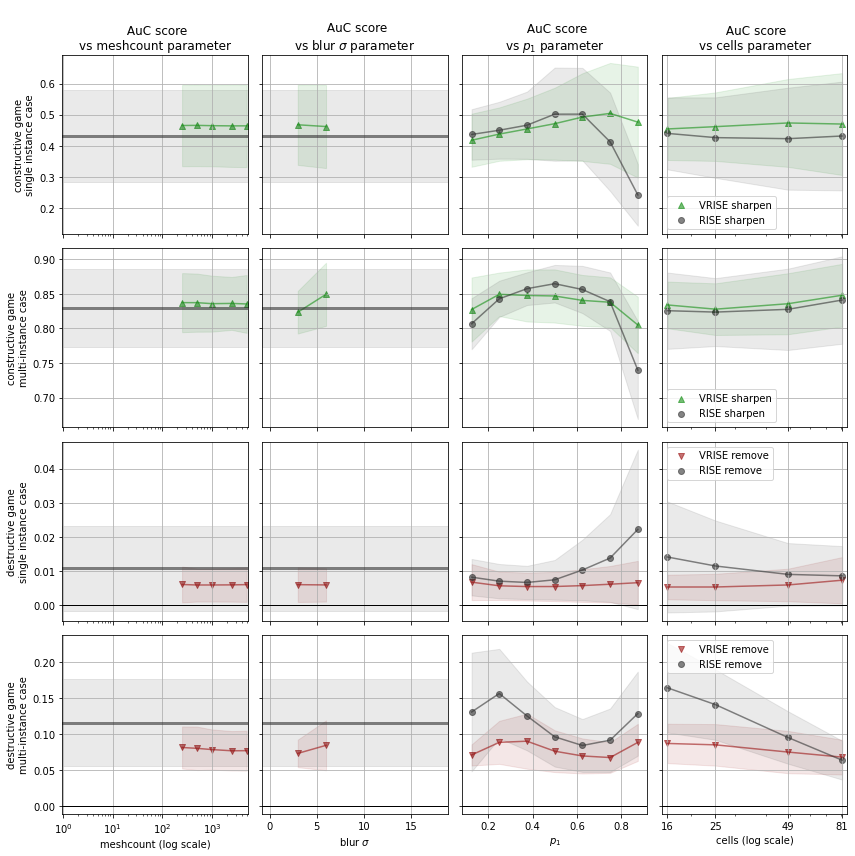}
  \caption[Relationships between parameter values and map quality as measured by various types of AltGame. ("better" subset)]{Plots of rel$\Delta$(AuC) for each parameter dimension and type of AltGame. Each data point is a mean rel$\Delta$ of all results involving it's x-axis parameter value. \textbf{Results for detrimental values of VRISE-specific params were excluded}, as evidenced by truncated series for \texttt{meshcount} and blur $\sigma$}
  \label{fig:grid_search.rawauc_origame_better_subset}
\end{figure}

Because certain values of blur $\sigma$ and \texttt{meshcount} significantly impair the quality of maps, we provide a complementary set of filtered aggregations, which exclude extreme values of blur $\sigma$ and \texttt{meshcount}. Dataset filtering rules are explained in plot captions. The validity of results and conclusions is not affected by this adjustment, as VRISE and RISE are still being compared on the full spectrum of RISE-specific parameters ($p_1$, \texttt{polygons}).
Figures~\ref{fig:grid_search.rawscores_vs_N_better_subset} - \ref{fig:grid_search.rawauc_origame} show aggregate results before and after removal of detrimental parameter sets.
Once the extremes are removed, VRISE almost universally improves over RISE, regardless of the configuration of $p_1$ and \texttt{polygons}. Figures~\ref{fig:grid_search.rawauc_origame} and \ref{fig:grid_search.rawauc_origame_better_subset} are the best visual summary of the implications of this change.

\begin{figure}[H]
  \centering
  \includegraphics[width=0.90\textwidth]{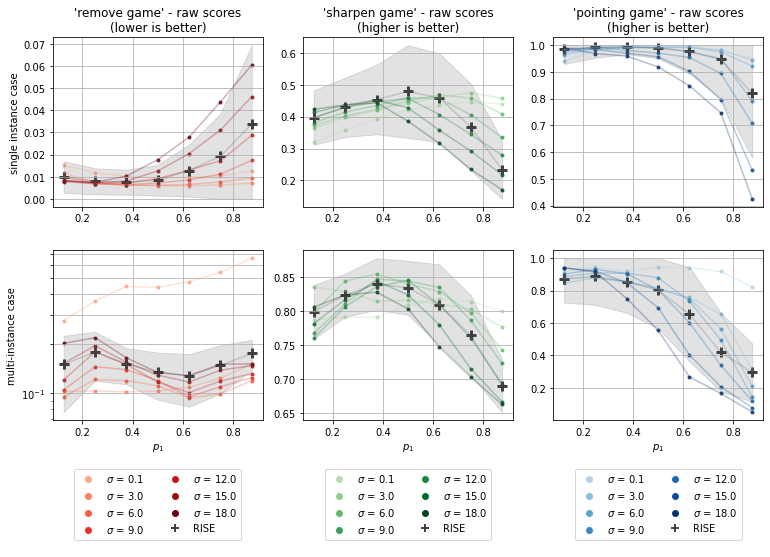}
  \caption[Influence of VRISE blur $\sigma$ on quality metrics, compared against RISE. Plotted against $p_1$. Narrowed param space.]{Comparison of quality scores for VRISE and RISE - blur $\sigma$ vs. $p_1$. Data for $\texttt{meshcount} < 100$ was excluded from aggregation to regularize results. Shaded area visualizes standard deviation of RISE results.}
  \label{fig:grid_search.rawscore_sigma_VS_p1_narrowed}
\end{figure}

\begin{figure}[H]
  \centering
  \includegraphics[width=0.90\textwidth]{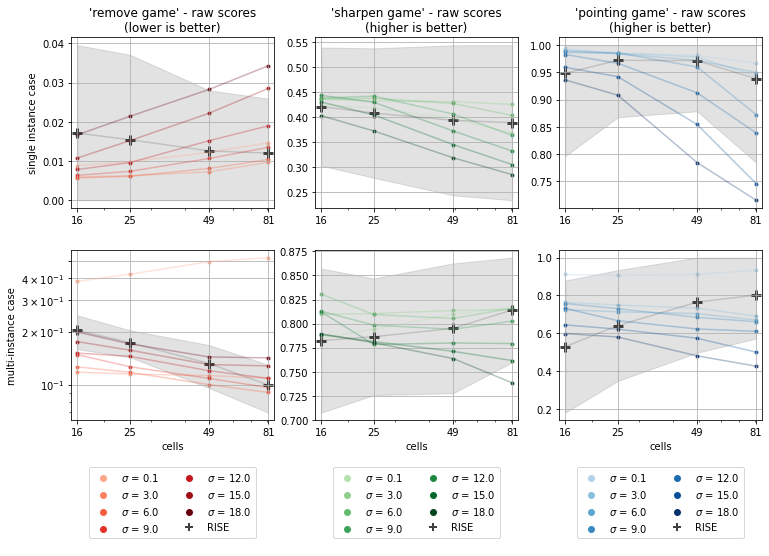}
  \caption[Influence of VRISE blur $\sigma$ on quality metrics, compared against RISE. Plotted against \texttt{polygons}]{Influence of VRISE blur $\sigma$ on quality metrics, compared against RISE. Plotted against \texttt{polygons}. Data for $\texttt{meshcount} < 100$ was excluded from aggregation to regularize results. Shaded area visualizes standard deviation of RISE results.}
  \label{fig:grid_search.rawscore_sigma_VS_polygons_narrowed}
\end{figure}

Figures~\ref{fig:grid_search.rawscore_sigma_VS_p1_narrowed} and \ref{fig:grid_search.rawscore_sigma_VS_polygons_narrowed} show a more detailed breakdown of relationships between raw values of quality metrics and combinations of values of ($\sigma$, $p_1$) and ($\sigma$, \texttt{polygons}), respectively.

\begin{figure}[H]
  \centering
  \includegraphics[width=\textwidth]{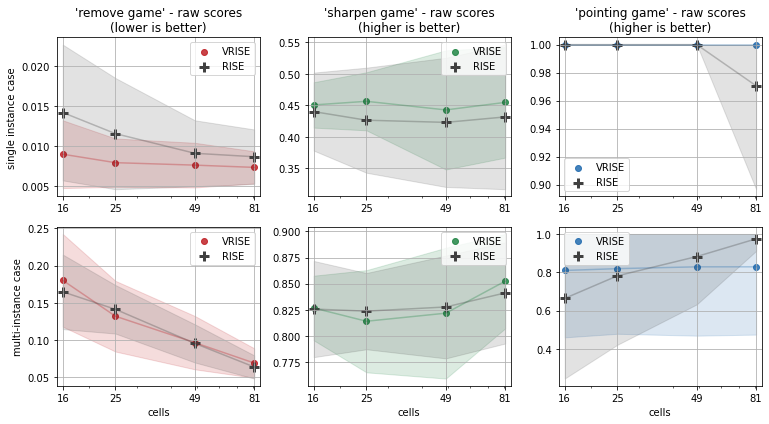}
  \caption[Influence of VRISE blur $\sigma$ on quality metrics, compared against RISE. Plotted against \texttt{polygons}, using only configurations directly matching those of RISE.]{Influence of VRISE blur $\sigma$ on quality metrics, compared against RISE. Plotted against \texttt{polygons}. \textbf{Values of $\sigma$ are determined by \texttt{polygons} and match RISE}. Data for $\texttt{meshcount} < 100$ was excluded from aggregation to regularize results. Shaded area visualizes standard deviation. High stddev is a consequence of aggregating over all values of $p_1$.}
  \label{fig:grid_search.rawscore_sigma_VS_polygons_narrowed_fixedblur}
\end{figure}

\begin{figure}[H]
  \centering
  \includegraphics[width=\textwidth]{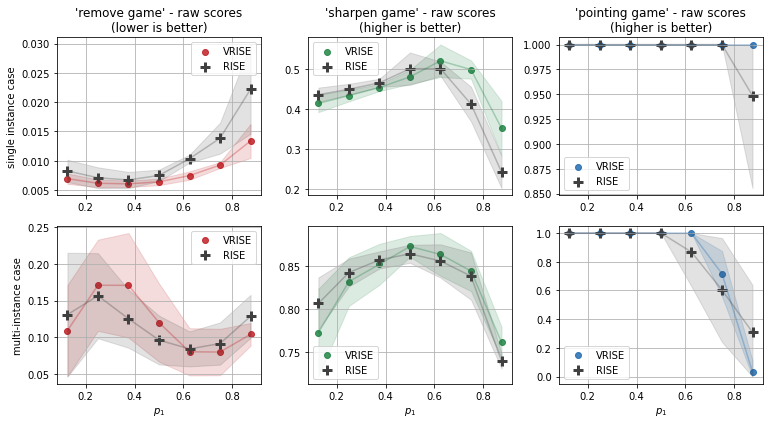}
  \caption[Influence of VRISE blur $\sigma$ on quality metrics, compared against RISE. Plotted against $p_1$,using only configurations directly matching those of RISE.]{Influence of VRISE blur $\sigma$ on quality metrics, compared against RISE. Plotted against $p_1$. \textbf{Values of $\sigma$ are determined by \texttt{polygons} and match RISE}. Data for $\texttt{meshcount} < 100$ was excluded from aggregation to regularize results. Shaded area visualizes standard deviation.}
  \label{fig:grid_search.rawscore_sigma_VS_p1_narrowed_fixedblur}
\end{figure}

Figures~\ref{fig:grid_search.rawscore_sigma_VS_polygons_narrowed_fixedblur} and \ref{fig:grid_search.rawscore_sigma_VS_p1_narrowed_fixedblur} show an even more strongly filtered subset of data from the previous two figures. This time, VRISE series use only those parameter sets which match RISE configurations exactly, providing the most precise comparison of these two algorithms.

\subsection{Analysis}

\input{tex/evaluation/gs_observations_reldiff_origame.tex}

\subsection{Conclusions}

It appears that VRISE has the greatest advantage over RISE in the multi-instance case. Irregular occlusion shapes combined with use of multiple meshes increase the likelihood of capturing salient features more accurately than a rectangular grid could, especially when individual occlusions are large.

\texttt{meshcount} has a limited capacity to improve map quality, but as few as 100 meshes are enough to exhaust it. Since each additional mesh increases the computational cost of mask generation, it's better that the potential for improvement is exhausted quickly.

The other parameter introduced by VRISE, blur $\sigma$, influences the results similarly to $p_1$. Intermediate values tend to improve quality, while extremes significantly impair it.
When blur strength is reduced, VRISE becomes less sensitive than RISE to very high values of $p_1$, resulting in relative improvement. Blur $\sigma$ also appears to have the greatest influence over map quality, out of all of our modifications. In practice, it can make or break the performance of any $p_1$ + \texttt{polygons} configuration. It's surprising that this coincidental implementation detail, which unties blur strength from mesh density, has turned out to have far more influence on the level of improvement than use of Voronoi meshes.

The results presented in this section, show that VRISE is capable of consistently outperforming RISE on a larger dataset, but only if blur $\sigma$ and \texttt{meshcount} are appropriately configured.

\begin{figure}[H]
  \centering
  \includegraphics[width=0.9\textwidth]{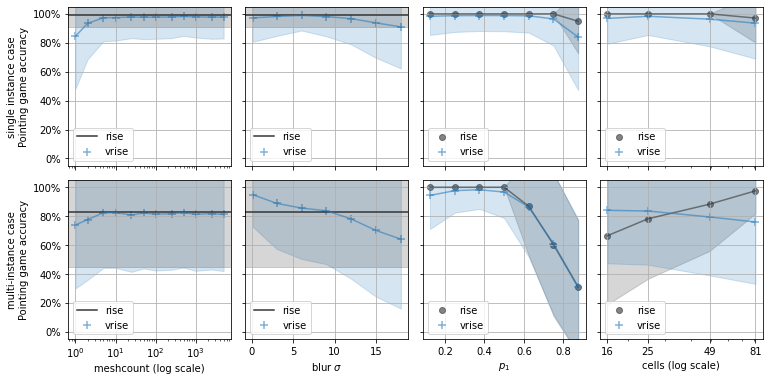}
  \caption[Pointing game accuracy scores for RISE and VRISE]{Pointing game accuracy for RISE and VRISE. Each data point is a mean rel$\Delta$ of all results involving its corresponding x-axis parameter value. Shaded bands visualise standard deviation for corresponding data points.}
  \label{fig:grid_search.pointing_game_rawscore}
\end{figure}

\begin{figure}[H]
  \centering
  \includegraphics[width=0.85\textwidth]{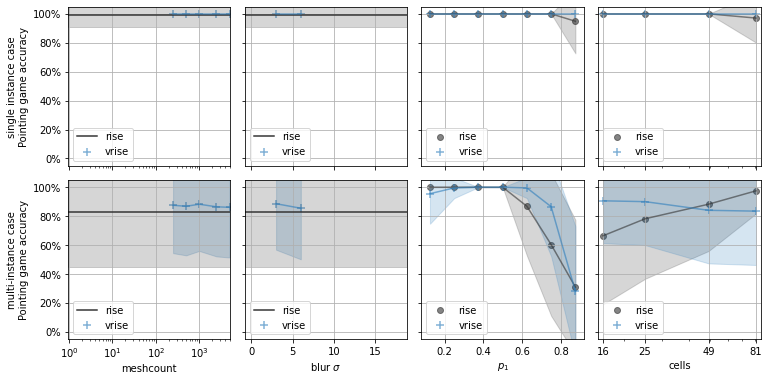}
  \caption[Pointing game accuracy scores for RISE and VRISE ("better" subset)]{Pointing game accuracy for RISE and VRISE. Each data point is a mean rel$\Delta$ of all results involving its corresponding x-axis parameter value. Shaded bands visualise standard deviation for corresponding data points. \textbf{Results for detrimental values of VRISE-specific params were excluded}, as evidenced by truncated series for \texttt{meshcount} and blur $\sigma$}
  \label{fig:grid_search.pointing_game_rawscore_better_subset}
\end{figure}

\begin{figure}[H]
  \centering
  \includegraphics[width=\textwidth]{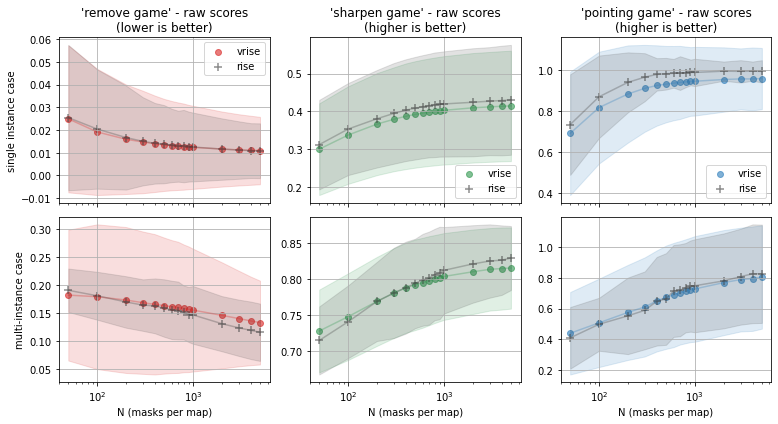}
  \caption[Raw scores of quality metrics, plotted against $N_{masks}$, averaged over all parameter sets.]{Raw scores of quality metrics, plotted against $N_{masks}$, averaged over all parameter sets.}
  \label{fig:grid_search.rawscores_vs_N}
\end{figure}

\begin{figure}[H]
  \centering
  \includegraphics[width=\textwidth]{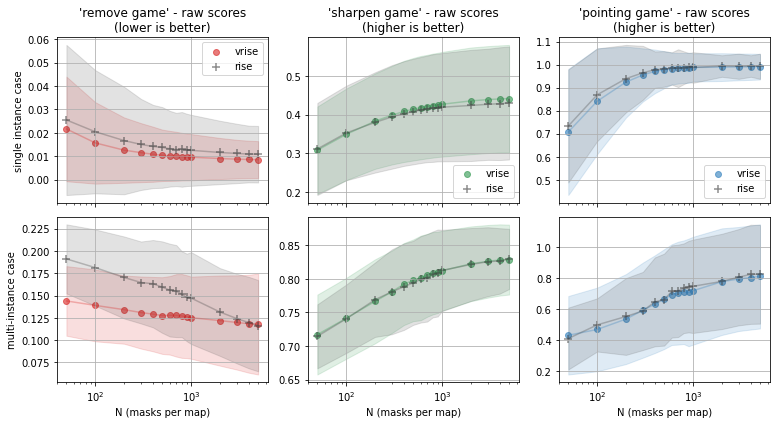}
  \caption[Raw scores of quality metrics, plotted against $N_{masks}$, averaged over all parameter sets. Subset strictly matching RISE.]{Raw scores of quality metrics, plotted against $N_{masks}$. Averaged over all matching parameter sets. \textbf{Contains only results for paramsets which fully match RISE configurations}. Notice that after exclusion of low-blur configurations the results have become worse than those in the previous figure. High stddev is a consequence of aggregation over all $p_1$ values}
  \label{fig:grid_search.rawscores_vs_N_matching}
\end{figure}

%% file: tbl/grid_search/grid_search_param_variance.tex
\begin{table}
\caption[Standard deviations of result quality across each parameter dimension.]{Standard deviation of result quality for each parameter dimension. $N_{masks} = 5000$}
\label{tbl:grid_search.param_stddev}
\begin{tabular}{c||cc|cc|cc|c}
\toprule
{metric >} & \multicolumn{2}{c|}{AltGame (all)} & \multicolumn{2}{c|}{Pointing Game} & \multicolumn{2}{c|}{Consistency} & \multirow{3}{*}{mean} \\
{class instances >} & \multirow{2}{*}{single} & \multirow{2}{*}{multiple} & \multirow{2}{*}{single} & \multirow{2}{*}{multiple} & \multirow{2}{*}{single} & \multirow{2}{*}{multiple} & {} \\
{param} & {} & {} & {} & {} & {} & {} & {} \\
\midrule
\texttt{polygons} & 0.0188 & 0.0223 & 0.0198 & 0.0380 & 0.0060 & 0.0033 & 0.0180 \\
meshcount & 0.0094 & 0.0230 & 0.0393 & 0.0252 & 0.0211 & 0.0144 & 0.0221 \\
blur $\sigma$ & 0.0179 & 0.0832 & 0.0287 & 0.1076 & 0.0073 & 0.0139 & 0.0431 \\
$p_1$ & 0.0251 & 0.0484 & 0.0548 & 0.2556 & 0.0066 & 0.0040 & 0.0657 \\
\bottomrule
\end{tabular}
\end{table}

%% file: tex/evaluation/gs_observations_reldiff_origame.tex
Observations based on results for fully converged maps (plots \ref{fig:grid_search.catdog_rauc_grid} - \ref{fig:grid_search.goldfish_pg_grid} (in appendix)):
\begin{itemize}
  \item $p_1$
  \begin{itemize}
    \item ($p_1$, single-case, both game types) - increased variance of outcomes. Low values of other params tend to benefit from larger sigma but that's mostly due to RISE score dropping faster than VRISE (\ref{fig:grid_search.rawauc_origame}). Trends are consistent for both PG and AuC.
    \item ($p_1$, multi-case, AltGame) - VRISE has an advantage when $p_1 < 0.5$, especially when paired with low blur $\sigma$, but there is no clear trend for $p_1$.
    \item ($p_1$, multi-case, both metrics) - The variance of Pointing Game outcomes increases as $p_1$ grows but the AuC remains relatively constant. Combinations with high values of other parameters cause a significant decrease of PG accuracy.
  \end{itemize}
  \item \texttt{polygons}
  \begin{itemize}
    \item (\texttt{polygons}, both cases, both game types) - the improvement is inversely correlated with density of the mesh and we see the greatest improvement when the number of polygons is low. It's visible even in highly aggregated results (fig.~\ref{fig:grid_search.rawauc_origame}), especially for the destructive variant of AltGame.
  \end{itemize}
  \item blur $\sigma$
  \begin{itemize}
    \item (blur $\sigma$, both cases, both game types) - similarly to \texttt{polygons}, this parameter is best kept low, although not too low. When blur is disabled completely, quality decreases and results become significantly less consistent. Using weaker blur than RISE would, improves both AltGame AuC and Pointing Game results.
  \end{itemize}
  \item \texttt{meshcount}
  \begin{itemize}
    \item (\texttt{meshcount}, both cases, both game types) - high \texttt{meshcount} is unable to improve results on its own. Because the "improvement" is asymptotic and ceases almost completely once the number of meshes used reaches 100, "low \texttt{meshcount} impairs map quality" would be a more accurate summary of its effect. In absence of shifts, VRISE must use at least 100 meshes to be able to compete against RISE and other methods.
  \end{itemize}
\end{itemize}

Observations of changes in map quality over time (plots \ref{fig:grid_search.checkpoints_auc_reldiff_narrowed}, \ref{fig:grid_search.checkpoints_pg_absdiff_narrowed}):
\begin{itemize}
  \item In the majority of cases, VRISE tends to maintain improvement on a steady level as the map generation process progresses. On one hand, it's regrettable that VRISE doesn't widen this gap, on the other, however, it's good to see that the full relative improvement potential is achieved early on. If few masks must be used, the resulting map will still be better on average.
  \item (all parameters, both cases, PG) - the majority of series trend towards zero or slightly above it. That's because establishing the location of a single salient region requires relatively few evaluations, compared to a multi-instance case with several salient regions scattered across the image. By the time $N = 5000$, virtually all maps will have correctly located that single object instance, and so there will be very little room for improvement since Pointing Game operates on a binary, \texttt{hit/miss} basis. (See \figref{fig:grid_search.rawscores_vs_N_better_subset})
  \item ($p_1$, both cases, PG) - there is no value of $p_1$ which would give VRISE a clear advantage.
  \item ($p_1$, both cases, AltGame) - Surprisingly enough, VRISE tends to improve over RISE the most when $p_1$ is high. \figref{fig:grid_search.rawauc_origame_better_subset} offers an explanation. VRISE doesn't necessarily improve results in this case, it only withstands the negative influence of high $p_1$ better than RISE does.
  \item (\texttt{polygons}, both cases, both metrics) - low polygon count consistently correlates with improvement in this aggregation, matching the results in fig.~\ref{fig:grid_search.rawauc_origame_better_subset}
  \item (blur $\sigma$) - lower blur sigma allows VRISE to maintain or even increase its lead over RISE. High blur $\sigma$ causes VRISE to fall behind, especially in the multi-instance case.
  \item (\texttt{meshcount}) - VRISE needs to use about 100 meshes per map to match the result quality of RISE. Very low values of \texttt{meshcount} ($\leq5$) significantly decrease map quality.
\end{itemize}

Observations of the influence of decreased blur strength on raw quality measures (fig.~\ref{fig:grid_search.rawscore_sigma_VS_p1_narrowed},  \ref{fig:grid_search.rawscore_sigma_VS_polygons_narrowed_fixedblur}):
\begin{itemize}
  \item Destructive AltGame - "remove" variant: When plotted against $p_1$, RISE and VRISE follow the same trend - the higher the $p_1$, the worse the results. The comparison against \texttt{polygons} parameter values (fig.~\ref{fig:grid_search.rawscore_sigma_VS_polygons_narrowed}), however, shows that RISE scores tend to improve as mesh density increases, while VRISE scores decline. This effect is likely caused by the fact that the strength of blur used by RISE is inversely proportional to grid density. According to \tblref{tbl:blur_sigma_ssim_stats}, the closest RISE blur strength values corresponding to the mesh density values used in this experiment, are, consecutively: 15, 12, 9 and 7. If we were to keep only those VRISE datapoints which match the parameter values used by RISE, these new VRISE data series would trend downwards as well. Not because of the increasing mesh density, but because of the blur $\sigma$ decreasing as mesh density grows. Figures~\ref{fig:grid_search.rawscore_sigma_VS_polygons_narrowed_fixedblur} and \ref{fig:grid_search.rawscore_sigma_VS_p1_narrowed_fixedblur} show this exact outcome.

  \item Constructive AltGame - "sharpen" variant: When plotted against \texttt{polygons} (fig.~\ref{fig:grid_search.rawscore_sigma_VS_polygons_narrowed}), RISE datapoints follow the trend described in the previous point. The tendency of VRISE scores to drop as $p_1$ increases is discussed in more detail in sec.~\ref{sec:eval.outro}

  \item Pointing Game: High blur $\sigma$ universally impairs accuracy. Both figures show that VRISE without blur is surprisingly resistant to changes in $p_1$ as well as \texttt{polygons}, in the multi-instance case. In case of RISE, dense meshes significantly improve accuracy in the multi-instance case, but impair it in single-instance case.

  \item VRISE vs RISE: In nearly all configurations (except for the sharpen game evaluation of low $p_1$ maps), VRISE performs comparably to or better than RISE.

\end{itemize}

Observations of results for configurations which match the blur strength used by RISE (fig.~\ref{fig:grid_search.rawscore_sigma_VS_p1_narrowed_fixedblur}, \ref{fig:grid_search.rawscore_sigma_VS_polygons_narrowed_fixedblur})
\begin{itemize}
    \item \texttt{polygons} - when blur strength is configured to match RISE, mesh density becomes the most relevant parameter for map quality (fig.~\ref{fig:grid_search.rawscore_sigma_VS_p1_narrowed_fixedblur})

    \item Destructive AltGame (Remove) - When blur strength is configured to match RISE, VRISE performs better on average in the single-instance case. In the multi-instance case, judging by the sinusoidal characteristic and significant standard deviation in  fig.~\ref{fig:grid_search.rawscore_sigma_VS_p1_narrowed_fixedblur}, the outcome appears to be highly sensitive to the combination of $p_1$ and \texttt{polygons}.

    \item Constructive AltGame (Sharpen) - the significance of $p_1$ and \texttt{polygons} is inverted. Sharpen game results are significantly influenced by $p_1$, as shown by parabolic trends in fig.~\ref{fig:grid_search.rawscore_sigma_VS_p1_narrowed_fixedblur}, but almost unaffected by mesh density fig.~\ref{fig:grid_search.rawscore_sigma_VS_p1_narrowed_fixedblur}, regardless of instance count. VRISE is more resistant to high $p_1$, scoring higher than RISE in these cases.

    \item Pointing Game - Both VRISE and RISE achieve 100\% accuracy in all configurations except ($p_1 = 0.875$; $\texttt{polygons} = 81$), where only VRISE is able to maintain 100\% accuracy across all samples. Multi-instance case is the polar opposite of single-instance. VRISE is very sensitive to the configuration of $p_1$, regardless of mesh density, as evidenced by constant width of standard deviation bands in fig.~\ref{fig:grid_search.rawscore_sigma_VS_polygons_narrowed_fixedblur}, and almost indifferent to \texttt{polygons} as shown by small stddev in fig.~\ref{fig:grid_search.rawscore_sigma_VS_p1_narrowed_fixedblur}. Whereas for RISE, increasing mesh density improves Pointing Game accuracy.
\end{itemize}

%% file: tex/evaluation/targeted_s.tex
\section{Evaluation controlling for differences in blur strength ("FixedSigma")}
\label{sec:targeted_s}

\begin{minipage}[l]{0.50\textwidth}

In this evaluation we aimed to strike a balance between several objectives:
\begin{itemize}
  \item Use equal blur strength for both algorithms, to isolate the influence of multi-mesh Voronoi occlusion patterns on saliency map quality. To accomplish this, values of \texttt{polygons} were paired with specific blur $\sigma$ values according to section \ref{ch:blur_sigma}.
  \item Evaluate our solution on a larger dataset, to make our conclusions less likely to be skewed by peculiarities of individual images.
  \item Keep the array of parameter sets broad enough to be able to verify observations made in the previous section on a larger dataset, while limiting the computational cost of a single run to that of a single run of GridSearch.
  \item Share parameter sets with GridSearch and ImgNet to maintain direct comparability of results between these experiments.
\end{itemize}

In consequence, although this study focuses primarily on measuring the impact of Voronoi meshes on result quality in isolation from influence of modified blur strength, it also acts as a bridge between GridSearch and ImgNet.

The images used in this evaluation were randomly selected from the validation split of ILSVRC2012.

Both algorithms use an occlusion selector with informativeness guarantee (see sec.~\ref{sec:guaranteed_grid_gens}) and never encounter uninformative masks during map generation.

\end{minipage}
\hspace{0.05\textwidth}
\begin{minipage}[r]{0.45\textwidth}
  \raggedright\footnotesize{\textbf{FixedSigma - configuration}}
  \resizebox{\textwidth}{!}{
    \begin{tabular}{c|cc}
      \toprule
      {parameter} & \multicolumn{2}{c}{included values} \\
      \midrule
      \midrule
      \multirow{2}{*}{(\texttt{polygons}; blur $\sigma$)} & {(16; 15.0)} & {(25; 12.0)} \\
      {} & {(49; 9.0)} & {(81; 7.0)} \\
      \midrule
      {$N_{masks}$} & \multicolumn{2}{c}{4000} \\
      \midrule
      {\texttt{meshcount}} & \multicolumn{2}{c}{1000} \\
      \midrule
      {$p_1$} & {0.25} & {0.5} \\

      \midrule
      \midrule

      {$\#images$} & \multicolumn{2}{c}{200} \\
      \midrule
      {classifier} & \multicolumn{2}{c}{ResNet50} \\
      \midrule
      {SMap Gen mode} & \multicolumn{2}{c}{FP16} \\
      {Storage mode} & \multicolumn{2}{c}{FP32} \\
      {AltGame mode} & \multicolumn{2}{c}{FP16} \\
      \midrule
      {Occlusion Selector} & \multicolumn{2}{c}{HybridGridGen} \\
      \bottomrule
    \end{tabular}
  }
  \vspace{0.3cm}

  \raggedright\footnotesize{\textbf{Dataset subgroups}}
  \resizebox{\textwidth}{!}{
    \begin{tabular}{cc|c|c}
      \toprule
      \multicolumn{2}{c|}{group name} & {\texttt{\#img}} & {criterion} \\
      \midrule
      \midrule
      \multicolumn{2}{c|}{"single"} & {142} & {$\texttt{\#obj} = 1$} \\
      \midrule
      \multirow{2}{*}{"multiple"} & {"few"} & {41} & {$1 < \texttt{\#obj} < 5$} \\
      \cmidrule[.5pt]{2-4}
      {} & {"many"} & {17} & {$\texttt{\#obj} \geq 5$} \\
      \bottomrule
    \end{tabular}
  }
\end{minipage}

Tables~\ref{tbl:targeted_s.rawscores.chk} and \ref{tbl:targeted_s.rawscores.params} provide raw values of quality metrics and context for the relative differences presented in tables~\ref{tbl:targeted_s.reldiff.chk} and \ref{tbl:targeted_s.reldiff.params_instances}.

Table~\ref{tbl:targeted_s.reldiff.params_instances} shows that VRISE performs better than RISE when meshes are coarse and loses its slight advantage as the number of polygons increases (visible more clearly in table~\ref{tbl:targeted_s.rawscores.params} through highlights of better results in each category).
The differences in intra-set map consistency between the two algorithms are negligible, regardless of configuration, with RISE scoring minimally higher in nearly all cases (fig.~\ref{fig:targeted_s.consistency}).
The results of pointing game also show the greatest positive difference between the two algorithms in low-polygon, multi-instance cases, where accuracy of VRISE maps is up to 15\% higher (fig.~\ref{fig:targeted_s.pg_grid}).

Both table~\ref{tbl:targeted_s.reldiff.params_instances} and fig.~\ref{fig:targeted_s.auc_remove} show a trend in destructive AltGame scores, which opposes its counterpart from the GridSearch evaluation. Here, the destructive game AuC is lower for the lower $p_1$ configuration.

Table~\ref{tbl:targeted_s.reldiff.chk} shows how quality improvement changes with time. The general trend is convergence towards the same level of quality, with VRISE lagging slightly behind in single-instance Pointing Game and exceeding in AltGame. The only exception is the multi-instance Pointing Game, where VRISE continues to slowly climb up throughout the entire evaluation.
Plots~\ref{fig:targeted_s.auc_remove} - \ref{fig:targeted_s.pg_grid} visualize the same set of results, using a more granular grouping by number of object instances, which highlights differences within the multi-instance result group, now additionally subdivided into "few" and "many" groups.

\input{tbl/targeted_s/rawscores_chk_instances.tex}
\input{tbl/targeted_s/rawscores_params.tex}

\input{tbl/targeted_s/reldiff_chk_instances.tex}
\input{tbl/targeted_s/reldiff_params_instances.tex}

\begin{minipage}[l]{0.48\textwidth}
  \begin{figure}[H]
    \centering
    \includegraphics[width=\textwidth]{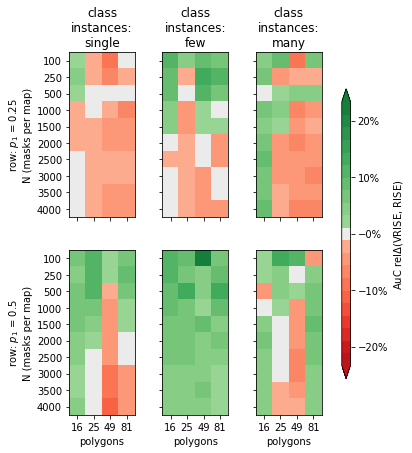}
    \caption[AuC rel$\Delta$ between VRISE and RISE maps (destructive AltGame) - controlling for blur strength]{\textbf{Destructive AltGame} AuC rel$\Delta$ between VRISE and RISE maps. Positive values indicate better results for VRISE maps}
    \label{fig:targeted_s.auc_remove}
  \end{figure}
\end{minipage}
\hspace{0.04\textwidth}
\begin{minipage}[r]{0.48\textwidth}
  \begin{figure}[H]
    \centering
    \includegraphics[width=\textwidth]{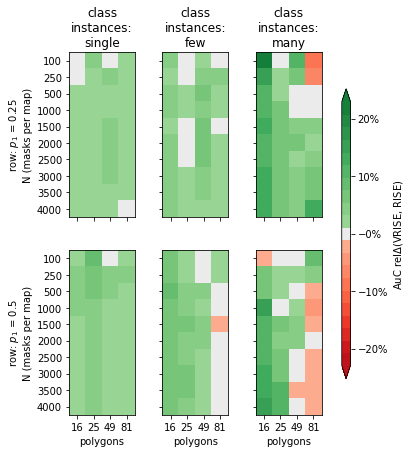}
    \caption[AuC rel$\Delta$ between VRISE and RISE maps (constructive AltGame) - controlling for blur strength]{\textbf{Constructive AltGame} AuC rel$\Delta$ between VRISE and RISE maps.}
    \label{fig:targeted_s.auc_sharpen}
  \end{figure}
\end{minipage}

\begin{minipage}[l]{0.48\textwidth}
  \begin{figure}[H]
    \centering
    \includegraphics[width=\textwidth]{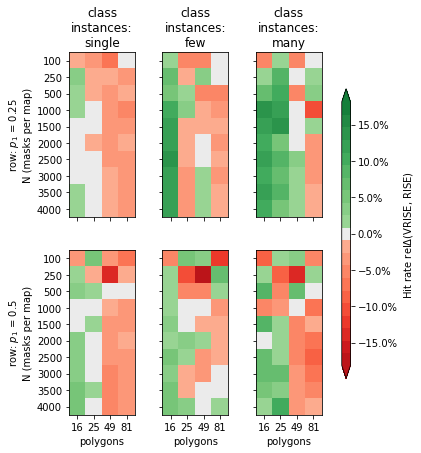}
    \caption[Pointing game accuracy rel$\Delta$ between VRISE and RISE maps - controlling for blur strength]{\textbf{Pointing game} accuracy rel$\Delta$ between VRISE and RISE maps.}
    \label{fig:targeted_s.pg_grid}
  \end{figure}
\end{minipage}
\hspace{0.04\textwidth}
\begin{minipage}[r]{0.48\textwidth}
  \begin{figure}[H]
    \centering
    \includegraphics[width=\textwidth]{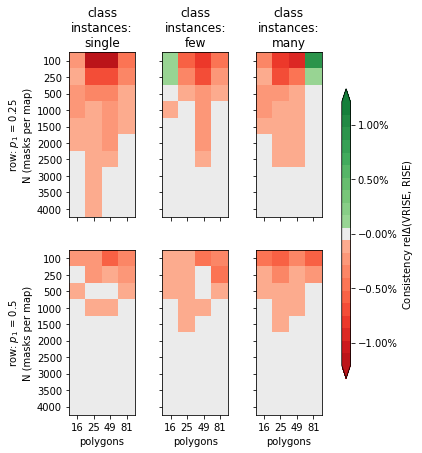}
    \caption[rel$\Delta$ consistency between VRISE and RISE maps - controlling for blur strength]{\textbf{Consistency} rel$\Delta$ between VRISE and RISE maps. Positive values indicate greater consistency of VRISE maps}
    \label{fig:targeted_s.consistency}
  \end{figure}
\end{minipage}

%% file: tbl/targeted_s/rawscores_chk_instances.tex
\begin{table}
\caption[Evolution of raw scores and standard deviations for RISE and VRISE over time.]{Evolution of raw scores and standard deviations for RISE and VRISE over time. Averaged over all parameter sets. Highlights indicate the better result in each pair of columns.}
\label{tbl:targeted_s.rawscores.chk}
\resizebox{\textwidth}{!}{
  \begin{tabular}{ccc||cc|cc|cc||cc|cc|cc}
  \toprule
  \multicolumn{3}{c||}{aggregation >} & \multicolumn{6}{c||}{avg($intraset$) \textit{(higher is better)}} & \multicolumn{6}{c}{stddev($intraset$) \textit{(lower is better)}} \\
  \multicolumn{3}{c||}{metric >} & \multicolumn{2}{c}{AltGame} & \multicolumn{2}{c}{PointingGame} & \multicolumn{2}{c||}{Consistency} & \multicolumn{2}{c}{AltGame} & \multicolumn{2}{c}{PointingGame} & \multicolumn{2}{c}{Consistency} \\
  \multicolumn{3}{c||}{algorithm >} & {rise} & {vrise} & {rise} & {vrise} & {rise} & {vrise} & {rise} & {vrise} & {rise} & {vrise} & {rise} & {vrise} \\
  {\#instances} & {game} & {$N_{mask}$} & {} & {} & {} & {} & {} & {} & {} & {} & {} & {} & {} & {} \\
  \midrule
  \multirow[c]{10}{*}{single} & \multirow[c]{5}{*}{remove} & 100 & 0.1250 & \bfseries 0.1113 & \bfseries 0.7591 & 0.7329 & \bfseries 0.9572 & 0.9519 & 0.0378 & \bfseries 0.0320 & \bfseries 0.2321 & 0.2421 & 0.0180 & \bfseries 0.0157 \\
   &  & 250 & 0.1108 & \bfseries 0.0990 & \bfseries 0.8309 & 0.8062 & \bfseries 0.9820 & 0.9791 & 0.0302 & \bfseries 0.0256 & \bfseries 0.1362 & 0.1637 & 0.0077 & \bfseries 0.0073 \\
   &  & 500 & 0.1051 & \bfseries 0.0933 & \bfseries 0.8387 & 0.8366 & \bfseries 0.9911 & 0.9896 & 0.0245 & \bfseries 0.0200 & \bfseries 0.1058 & 0.1065 & 0.0037 & \bfseries 0.0036 \\
   &  & 1000 & 0.0976 & \bfseries 0.0918 & \bfseries 0.8534 & 0.8382 & \bfseries 0.9957 & 0.9946 & 0.0193 & \bfseries 0.0175 & \bfseries 0.0790 & 0.0948 & \bfseries 0.0018 & 0.0019 \\
   &  & 4000 & 0.0910 & \bfseries 0.0876 & \bfseries 0.8657 & 0.8555 & \bfseries 0.9988 & 0.9985 & 0.0126 & \bfseries 0.0115 & \bfseries 0.0431 & 0.0564 & \bfseries 0.0005 & 0.0006 \\
     \cmidrule[.5pt]{2-15}
   & \multirow[c]{5}{*}{sharpen} & 100 & 0.6158 & \bfseries 0.6187 & \bfseries 0.7591 & 0.7329 & \bfseries 0.9572 & 0.9519 & 0.0521 & \bfseries 0.0495 & \bfseries 0.2321 & 0.2421 & 0.0180 & \bfseries 0.0157 \\
   &  & 250 & 0.6393 & \bfseries 0.6455 & \bfseries 0.8309 & 0.8062 & \bfseries 0.9820 & 0.9791 & 0.0413 & \bfseries 0.0397 & \bfseries 0.1362 & 0.1637 & 0.0077 & \bfseries 0.0073 \\
   &  & 500 & 0.6509 & \bfseries 0.6589 & \bfseries 0.8387 & 0.8366 & \bfseries 0.9911 & 0.9896 & 0.0336 & \bfseries 0.0322 & \bfseries 0.1058 & 0.1065 & 0.0037 & \bfseries 0.0036 \\
   &  & 1000 & 0.6611 & \bfseries 0.6684 & \bfseries 0.8534 & 0.8382 & \bfseries 0.9957 & 0.9946 & 0.0266 & \bfseries 0.0263 & \bfseries 0.0790 & 0.0948 & \bfseries 0.0018 & 0.0019 \\
   &  & 4000 & 0.6704 & \bfseries 0.6780 & \bfseries 0.8657 & 0.8555 & \bfseries 0.9988 & 0.9985 & 0.0171 & \bfseries 0.0164 & \bfseries 0.0431 & 0.0564 & \bfseries 0.0005 & 0.0006 \\
   \midrule
  \multirow[c]{10}{*}{multiple} & \multirow[c]{5}{*}{remove} & 100 & 0.1621 & \bfseries 0.1498 & \bfseries 0.6228 & 0.6052 & \bfseries 0.9622 & 0.9587 & 0.0402 & \bfseries 0.0370 & 0.3007 & \bfseries 0.2968 & 0.0145 & \bfseries 0.0129 \\
   &  & 250 & 0.1489 & \bfseries 0.1386 & \bfseries 0.7044 & 0.6927 & \bfseries 0.9840 & 0.9816 & 0.0344 & \bfseries 0.0303 & \bfseries 0.2151 & 0.2220 & 0.0063 & \bfseries 0.0062 \\
   &  & 500 & 0.1427 & \bfseries 0.1310 & 0.7131 & \bfseries 0.7168 & \bfseries 0.9919 & 0.9908 & 0.0291 & \bfseries 0.0266 & \bfseries 0.1711 & 0.1847 & 0.0032 & \bfseries 0.0031 \\
   &  & 1000 & 0.1365 & \bfseries 0.1286 & 0.7195 & \bfseries 0.7276 & \bfseries 0.9959 & 0.9952 & 0.0246 & \bfseries 0.0234 & \bfseries 0.1381 & 0.1394 & 0.0017 & \bfseries 0.0016 \\
   &  & 4000 & 0.1321 & \bfseries 0.1232 & 0.7315 & \bfseries 0.7543 & \bfseries 0.9989 & 0.9987 & 0.0174 & \bfseries 0.0174 & \bfseries 0.0770 & 0.0926 & 0.0005 & \bfseries 0.0005 \\
     \cmidrule[.5pt]{2-15}
   & \multirow[c]{5}{*}{sharpen} & 100 & 0.5165 & \bfseries 0.5209 & \bfseries 0.6228 & 0.6052 & \bfseries 0.9622 & 0.9587 & 0.0543 & \bfseries 0.0506 & 0.3007 & \bfseries 0.2968 & 0.0145 & \bfseries 0.0129 \\
   &  & 250 & 0.5439 & \bfseries 0.5487 & \bfseries 0.7044 & 0.6927 & \bfseries 0.9840 & 0.9816 & 0.0450 & \bfseries 0.0425 & \bfseries 0.2151 & 0.2220 & 0.0063 & \bfseries 0.0062 \\
   &  & 500 & 0.5600 & \bfseries 0.5642 & 0.7131 & \bfseries 0.7168 & \bfseries 0.9919 & 0.9908 & 0.0361 & \bfseries 0.0346 & \bfseries 0.1711 & 0.1847 & 0.0032 & \bfseries 0.0031 \\
   &  & 1000 & 0.5711 & \bfseries 0.5745 & 0.7195 & \bfseries 0.7276 & \bfseries 0.9959 & 0.9952 & 0.0286 & \bfseries 0.0283 & \bfseries 0.1381 & 0.1394 & 0.0017 & \bfseries 0.0016 \\
   &  & 4000 & 0.5800 & \bfseries 0.5858 & 0.7315 & \bfseries 0.7543 & \bfseries 0.9989 & 0.9987 & 0.0193 & \bfseries 0.0189 & \bfseries 0.0770 & 0.0926 & 0.0005 & \bfseries 0.0005 \\
  \bottomrule
  \end{tabular}
  }
\end{table}

%% file: tbl/targeted_s/rawscores_params.tex
\begin{table}
\caption[Raw scores and standard deviations for RISE and VRISE, for each parameter set separately.]{Raw scores and standard deviations for RISE and VRISE, for each paramset used in this section. $N_{masks} = 4000$. Bold font indicates the better of results in each (VRISE, RISE) pair.}
\label{tbl:targeted_s.rawscores.params}
\resizebox{\textwidth}{!}{
\begin{tabular}{ccc||cc|cc|cc||cc|cc|cc}
\toprule
{} & {} & {agg >} & \multicolumn{6}{c||}{avg($intraset$) \textit{(higher is better)}} & \multicolumn{6}{c}{stddev($intraset$) \textit{(lower is better)}} \\
{} & {} & {metric >} & \multicolumn{2}{c|}{AltGame} & \multicolumn{2}{c|}{PointingGame} & \multicolumn{2}{c||}{Consistency} & \multicolumn{2}{c|}{AltGame} & \multicolumn{2}{c|}{PointingGame} & \multicolumn{2}{c}{Consistency} \\
{} & {} & {algorithm >} & {rise} & {vrise} & {rise} & {vrise} & {rise} & {vrise} & {rise} & {vrise} & {rise} & {vrise} & {rise} & {vrise} \\
{polygons} & {$p_1$} & {game} & {} & {} & {} & {} & {} & {} & {} & {} & {} & {} & {} & {} \\
\midrule
\multirow[c]{4}{*}{16} & \multirow[c]{2}{*}{0.25} & remove & 0.1120 & \bfseries 0.1055 & 0.7729 & \bfseries 0.8120 & \bfseries 0.9987 & 0.9984 & 0.0084 & \bfseries 0.0080 & \bfseries 0.0398 & 0.0537 & \bfseries 0.0007 & 0.0007 \\
 &   & sharpen & 0.6295 & \bfseries 0.6381 & 0.7729 & \bfseries 0.8120 & \bfseries 0.9987 & 0.9984 & \bfseries 0.0099 & 0.0100 & \bfseries 0.0398 & 0.0537 & \bfseries 0.0007 & 0.0007 \\
 \cmidrule[.5pt]{2-15}
 & \multirow[c]{2}{*}{0.5} & remove & 0.1257 & \bfseries 0.1091 & 0.7129 & \bfseries 0.7590 & \bfseries 0.9996 & 0.9995 & 0.0174 & \bfseries 0.0140 & \bfseries 0.0756 & 0.0868 & 0.0002 & \bfseries 0.0002 \\
 &  & sharpen & 0.6284 & \bfseries 0.6398 & 0.7129 & \bfseries 0.7590 & \bfseries 0.9996 & 0.9995 & 0.0145 & \bfseries 0.0126 & \bfseries 0.0756 & 0.0868 & 0.0002 & \bfseries 0.0002 \\
 \cmidrule[.5pt]{1-15}
\multirow[c]{4}{*}{25} & \multirow[c]{2}{*}{0.25} & remove & 0.1006 & \bfseries 0.0994 & 0.8379 & \bfseries 0.8390 & \bfseries 0.9986 & 0.9980 & \bfseries 0.0101 & 0.0105 & \bfseries 0.0415 & 0.0622 & \bfseries 0.0007 & 0.0009 \\
 &  & sharpen & 0.6405 & \bfseries 0.6466 & 0.8379 & \bfseries 0.8390 & \bfseries 0.9986 & 0.9980 & \bfseries 0.0122 & 0.0125 & \bfseries 0.0415 & 0.0622 & \bfseries 0.0007 & 0.0009 \\
 \cmidrule[.5pt]{2-15}
 & \multirow[c]{2}{*}{0.5} & remove & 0.1101 & \bfseries 0.1009 & 0.7886 & \bfseries 0.8070 & \bfseries 0.9995 & 0.9993 & 0.0169 & \bfseries 0.0142 & \bfseries 0.0736 & 0.0769 & 0.0003 & \bfseries 0.0002 \\
 &  & sharpen & 0.6427 & \bfseries 0.6531 & 0.7886 & \bfseries 0.8070 & \bfseries 0.9995 & 0.9993 & 0.0159 & \bfseries 0.0157 & \bfseries 0.0736 & 0.0769 & 0.0003 & \bfseries 0.0002 \\
 \cmidrule[.5pt]{1-15}
\multirow[c]{4}{*}{49} & \multirow[c]{2}{*}{0.25} & remove & 0.0939 & \bfseries 0.0930 & \bfseries 0.8871 & 0.8770 & \bfseries 0.9981 & 0.9976 & \bfseries 0.0114 & 0.0125 & \bfseries 0.0334 & 0.0508 & \bfseries 0.0009 & 0.0010 \\
 &  & sharpen & 0.6476 & \bfseries 0.6519 & \bfseries 0.8871 & 0.8770 & \bfseries 0.9981 & 0.9976 & 0.0176 & \bfseries 0.0173 & \bfseries 0.0334 & 0.0508 & \bfseries 0.0009 & 0.0010 \\
 \cmidrule[.5pt]{2-15}
 & \multirow[c]{2}{*}{0.5} & remove & 0.0985 & \bfseries 0.0955 & \bfseries 0.8536 & 0.8070 & \bfseries 0.9993 & 0.9992 & 0.0175 & \bfseries 0.0170 & \bfseries 0.0478 & 0.0756 & \bfseries 0.0003 & 0.0003 \\
 &  & sharpen & 0.6575 & \bfseries 0.6624 & \bfseries 0.8536 & 0.8070 & \bfseries 0.9993 & 0.9992 & 0.0219 & \bfseries 0.0206 & \bfseries 0.0478 & 0.0756 & \bfseries 0.0003 & 0.0003 \\
 \cmidrule[.5pt]{1-15}
\multirow[c]{4}{*}{81} & \multirow[c]{2}{*}{0.25} & remove & 0.0898 & \bfseries 0.0895 & \bfseries 0.9029 & 0.8730 & \bfseries 0.9981 & 0.9977 & 0.0132 & \bfseries 0.0129 & \bfseries 0.0280 & 0.0356 & \bfseries 0.0007 & 0.0009 \\
 &  & sharpen & 0.6464 & \bfseries 0.6511 & \bfseries 0.9029 & 0.8730 & \bfseries 0.9981 & 0.9977 & 0.0223 & \bfseries 0.0215 & \bfseries 0.0280 & 0.0356 & \bfseries 0.0007 & 0.0009 \\
 \cmidrule[.5pt]{2-15}

 & \multirow[c]{2}{*}{0.5} & remove & 0.0928 & \bfseries 0.0904 & \bfseries 0.8586 & 0.8350 & \bfseries 0.9990 & 0.9989 & 0.0167 & \bfseries 0.0164 & \bfseries 0.0837 & 0.0940 & \bfseries 0.0003 & 0.0003 \\
 &  & sharpen & 0.6611 & \bfseries 0.6669 & \bfseries 0.8586 & 0.8350 & \bfseries 0.9990 & 0.9989 & 0.0275 & \bfseries 0.0268 & \bfseries 0.0837 & 0.0940 & \bfseries 0.0003 & 0.0003 \\
\bottomrule
\end{tabular}
}
\end{table}

%% file: tbl/targeted_s/reldiff_chk_instances.tex
\begin{table}
\caption[Evolution of rel$\Delta$(VRISE, RISE) over time.]{Evolution of rel$\Delta$(VRISE, RISE) over time. In the 'AltGame*' column rel$\Delta$ values of destructive games were negated prior to aggregation to adjust them to the "higher is better" convention.}
\label{tbl:targeted_s.reldiff.chk}
\resizebox{\textwidth}{!}{

  \begin{tabular}{cc|ccc|ccc}
  \toprule
  \multicolumn{2}{c|}{aggregation >} & \multicolumn{3}{c|}{avg(rel$\Delta$(VRISE, RISE))} & \multicolumn{3}{c}{avg(rel$\Delta$(stddev(VRISE), stddev(RISE))} \\
  \multicolumn{2}{c|}{measure >} & {AltGame*} & {PointingGame} & {Consistency} & {AltGame} & {PointingGame} & {Consistency} \\
  {\#instances} & {$N_{masks}$} & \multicolumn{3}{c|}{\textit{(higher is better)}} & \multicolumn{3}{c}{\textit{(lower is better)}} \\
  \midrule
  \multirow[c]{7}{*}{single} & 100 & \phantom{-}2.39\% & -2.61\% & -0.55\% & \phantom{-}0.42\% & -1.00\% & \phantom{-}0.23\% \\
   & 250 & \phantom{-}3.17\% & -2.47\% & -0.30\% & \phantom{-}0.31\% & -2.76\% & \phantom{-}0.04\% \\
   & 500 & \phantom{-}3.18\% & -0.20\% & -0.15\% & \phantom{-}0.29\% & -0.07\% & \phantom{-}0.01\% \\
   & 1000 & \phantom{-}1.30\% & -1.52\% & -0.11\% & \phantom{-}0.10\% & -1.58\% & -0.01\% \\
   & 2000 & \phantom{-}0.80\% & -1.06\% & -0.06\% & \phantom{-}0.14\% & -1.38\% & -0.01\% \\
   & 3000 & \phantom{-}0.16\% & -1.50\% & -0.03\% & \phantom{-}0.13\% & -1.46\% & \phantom{-}0.00\% \\
   & 4000 & -0.14\% & -1.02\% & -0.03\% & \phantom{-}0.09\% & -1.33\% & -0.01\% \\
   \midrule
  \multirow[c]{7}{*}{multiple} & 100 & \phantom{-}5.47\% & -1.77\% & -0.36\% & \phantom{-}0.35\% & \phantom{-}0.38\% & \phantom{-}0.16\% \\
   & 250 & \phantom{-}4.26\% & -1.18\% & -0.24\% & \phantom{-}0.33\% & -0.69\% & \phantom{-}0.02\% \\
   & 500 & \phantom{-}4.65\% & \phantom{-}0.38\% & -0.11\% & \phantom{-}0.20\% & -1.36\% & \phantom{-}0.01\% \\
   & 1000 & \phantom{-}3.12\% & \phantom{-}0.81\% & -0.07\% & \phantom{-}0.08\% & -0.12\% & \phantom{-}0.01\% \\
   & 2000 & \phantom{-}2.91\% & \phantom{-}1.12\% & -0.03\% & \phantom{-}0.12\% & -1.43\% & \phantom{-}0.01\% \\
   & 3000 & \phantom{-}2.47\% & \phantom{-}1.14\% & -0.02\% & \phantom{-}0.06\% & -1.17\% & \phantom{-}0.01\% \\
   & 4000 & \phantom{-}2.54\% & \phantom{-}2.28\% & -0.02\% & \phantom{-}0.02\% & -1.57\% & \phantom{-}0.00\% \\
  \bottomrule
  \end{tabular}
  }
\end{table}

%% file: tbl/targeted_s/reldiff_params_instances.tex
\begin{table}
\caption[Relative difference of VRISE and RISE scores, grouped by paramset and object instances.]{Relative difference of VRISE and RISE scores, grouped by paramset and object instances. $N_{masks} = 4000$. In the 'AltGame*' column rel$\Delta$ values of destructive games were negated prior to aggregation to adjust them to the "higher is better" convention. Bold font indicates the better result in each pair of columns}
\label{tbl:targeted_s.reldiff.params_instances}
\resizebox{\textwidth}{!}{

\begin{tabular}{ccc||cc|cc|cc||cc|cc|cc}
\toprule
\multicolumn{3}{c||}{aggregation >} & \multicolumn{6}{c||}{avg(rel$\Delta$(VRISE, RISE))} & \multicolumn{6}{c}{avg(rel$\Delta$(stddev(VRISE), stddev(RISE))} \\
\multicolumn{3}{c||}{measure >} & \multicolumn{2}{c|}{AltGame*} & \multicolumn{2}{c|}{PointingGame} & \multicolumn{2}{c||}{Consistency} & \multicolumn{2}{c|}{AltGame} & \multicolumn{2}{c|}{PointingGame} & \multicolumn{2}{c}{Consistency} \\
\multicolumn{3}{c||}{instances >} & {single} & {multiple} & {single} & {multiple} & {single} & {multiple} & {single} & {multiple} & {single} & {multiple} & {single} & {multiple} \\
{game} & {$p_1$} & {cells} & {} & {} & {} & {} & {} & {} & {} & {} & {} & {} & {} & {} \\
\midrule
\multirow[c]{8}{*}{remove} & \multirow[c]{4}{*}{0.25} & 16 & -0.48\% & \bfseries \phantom{-}2.98\% & \phantom{-}1.07\% & \bfseries 10.89\% & -0.04\% & \bfseries -0.02\% & \phantom{-}0.10\% & \bfseries -0.07\% & \bfseries -1.95\% & -0.03\% & \bfseries -0.01\% & \phantom{-}0.01\% \\
 &  & 25 & -2.07\% & \bfseries -2.03\% & \bfseries \phantom{-}0.38\% & -0.54\% & -0.07\% & \bfseries -0.03\% & -0.03\% & \bfseries -0.07\% & -1.51\% & \bfseries -3.42\% & \bfseries -0.03\% & \phantom{-}0.01\% \\
 &  & 49 & \bfseries -4.25\% & -4.74\% & -2.07\% & \bfseries \phantom{-}1.58\% & -0.04\% & \bfseries -0.04\% & \bfseries -0.16\% & \phantom{-}0.02\% & -1.65\% & \bfseries -1.96\% & -0.01\% & \bfseries -0.01\% \\
 &  & 81 & -5.13\% & \bfseries -4.96\% & -3.26\% & \bfseries -2.32\% & -0.05\% & \bfseries -0.01\% & \bfseries \phantom{-}0.03\% & \phantom{-}0.05\% & \bfseries -0.93\% & -0.33\% & \bfseries -0.02\% & \phantom{-}0.01\% \\
 \cmidrule[.5pt]{2-15}
 & \multirow[c]{4}{*}{0.5} & 16 & \phantom{-}3.54\% & \bfseries \phantom{-}4.18\% & \bfseries \phantom{-}4.93\% & \phantom{-}3.84\% & \bfseries -0.00\% & -0.01\% & \phantom{-}0.44\% & \bfseries \phantom{-}0.11\% & \phantom{-}0.57\% & \bfseries -5.26\% & \phantom{-}0.01\% & \bfseries \phantom{-}0.00\% \\
 &  & 25 & -0.33\% & \bfseries \phantom{-}2.43\% & \phantom{-}0.44\% & \bfseries \phantom{-}5.27\% & -0.01\% & \bfseries -0.01\% & \phantom{-}0.32\% & \bfseries \phantom{-}0.14\% & \phantom{-}0.62\% & \bfseries -2.64\% & \bfseries \phantom{-}0.00\% & \phantom{-}0.00\% \\
 &  & 49 & -10.77\% & \bfseries \phantom{-}1.50\% & -5.94\% & \bfseries -1.53\% & \bfseries -0.01\% & -0.02\% & \phantom{-}0.09\% & \bfseries -0.06\% & \bfseries -3.56\% & -0.86\% & -0.00\% & \bfseries -0.01\% \\
 &  & 81 & -4.38\% & \bfseries \phantom{-}3.68\% & -3.74\% & \bfseries \phantom{-}1.03\% & -0.02\% & \bfseries -0.01\% & \phantom{-}0.07\% & \bfseries -0.07\% & \bfseries -2.26\% & \phantom{-}1.98\% & \bfseries -0.00\% & \phantom{-}0.00\% \\
\midrule
\multirow[c]{8}{*}{sharpen} & \multirow[c]{4}{*}{0.25} & 16 & \phantom{-}2.57\% & \bfseries \phantom{-}6.78\% & \phantom{-}1.07\% & \bfseries 10.89\% & -0.04\% & \bfseries -0.02\% & \bfseries -0.02\% & -0.00\% & \bfseries -1.95\% & -0.03\% & \bfseries -0.01\% & \phantom{-}0.01\% \\
 &  & 25 & \phantom{-}2.34\% & \bfseries \phantom{-}3.04\% & \bfseries \phantom{-}0.38\% & -0.54\% & -0.07\% & \bfseries -0.03\% & \bfseries -0.07\% & \phantom{-}0.08\% & -1.51\% & \bfseries -3.42\% & \bfseries -0.03\% & \phantom{-}0.01\% \\
 &  & 49 & \phantom{-}3.22\% & \bfseries \phantom{-}3.83\% & -2.07\% & \bfseries \phantom{-}1.58\% & -0.04\% & \bfseries -0.04\% & \bfseries \phantom{-}0.02\% & \phantom{-}0.07\% & -1.65\% & \bfseries -1.96\% & -0.01\% & \bfseries -0.01\% \\
 &  & 81 & \phantom{-}1.02\% & \bfseries \phantom{-}5.77\% & -3.26\% & \bfseries -2.32\% & -0.05\% & \bfseries -0.01\% & \phantom{-}0.14\% & \bfseries -0.06\% & \bfseries -0.93\% & -0.33\% & \bfseries -0.02\% & \phantom{-}0.01\% \\
 \cmidrule[.5pt]{2-15}
 & \multirow[c]{4}{*}{0.5} & 16 & \phantom{-}3.12\% & \bfseries \phantom{-}9.96\% & \bfseries \phantom{-}4.93\% & \phantom{-}3.84\% & \bfseries -0.00\% & -0.01\% & \phantom{-}0.22\% & \bfseries \phantom{-}0.13\% & \phantom{-}0.57\% & \bfseries -5.26\% & \phantom{-}0.01\% & \bfseries \phantom{-}0.00\% \\
 &  & 25 & \phantom{-}5.33\% & \bfseries \phantom{-}6.94\% & \phantom{-}0.44\% & \bfseries \phantom{-}5.27\% & -0.01\% & \bfseries -0.01\% & \phantom{-}0.03\% & \bfseries \phantom{-}0.00\% & \phantom{-}0.62\% & \bfseries -2.64\% & \bfseries \phantom{-}0.00\% & \phantom{-}0.00\% \\
 &  & 49 & \bfseries \phantom{-}2.52\% & \phantom{-}1.90\% & -5.94\% & \bfseries -1.53\% & \bfseries -0.01\% & -0.02\% & \phantom{-}0.15\% & \bfseries \phantom{-}0.05\% & \bfseries -3.56\% & -0.86\% & -0.00\% & \bfseries -0.01\% \\
 &  & 81 & \bfseries \phantom{-}1.51\% & -0.69\% & -3.74\% & \bfseries \phantom{-}1.03\% & -0.02\% & \bfseries -0.01\% & \phantom{-}0.08\% & \bfseries \phantom{-}0.06\% & \bfseries -2.26\% & \phantom{-}1.98\% & \bfseries -0.00\% & \phantom{-}0.00\% \\
\bottomrule
\end{tabular}
}
\end{table}

%% file: tex/evaluation/imgnet.tex
\section{Evaluation on ILSVRC2012 validation split ("ImgNet")}
\label{sec:imgnet}

\begin{minipage}[l]{0.50\textwidth}

In this section, we present results obtained on the entire validation split of ILSVRC2012 \cite{ilsvrc2012}, containing 50 images per each of 1000 object classes. During evaluation, RISE used the exact same configuration as in the original article and VRISE was configured to match it as closely as possible. The exact parameter values are specified in the table on the right hand side. The only difference is the configuration of numerical precision (see sec.~\ref{ch:fp16_eval}). In the original article, FP32 mode was most likely used for all steps, whereas we use FP16 mode for inference, to accelerate the experiment. The value of blur $\sigma$ for VRISE configuration has been established experimentally (see sec.~\ref{ch:blur_sigma}).

\end{minipage}
\hspace{0.05\textwidth}
\begin{minipage}[r]{0.4\textwidth}
    \raggedright\footnotesize{\textbf{ImageNet evaluation configuration}}
    \resizebox{\textwidth}{!}{
      \begin{tabular}{c|cc}
        \toprule
        {parameter} & {ResNet50} & {VGG16} \\
        \midrule
        {$N_{masks}$} & {8000} & {4000} \\
        {$p_1$} & \multicolumn{2}{c}{0.5} \\
        {\texttt{polygons}} & \multicolumn{2}{c}{49} \\
        {\texttt{meshcount}} & \multicolumn{2}{c}{1000} \\
        {blur $\sigma$} & \multicolumn{2}{c}{9.0} \\
        \midrule
        \midrule
        {\#images} & \multicolumn{2}{c}{50000} \\
        {runs} & \multicolumn{2}{c}{3} \\
        \midrule
        {SMap Gen mode} & \multicolumn{2}{c}{FP16} \\
        {Storage mode} & \multicolumn{2}{c}{FP32} \\
        {AltGame mode} & \multicolumn{2}{c}{FP16} \\
        \midrule
        {Occlusion Selector} & \multicolumn{2}{c}{HybridGridGen} \\
        \bottomrule
      \end{tabular}
    }
    \vspace{3mm}

\end{minipage}

76\% of images in this dataset contain only a single instance of a single object class, while the remaining 24\% contain two or more instances of a single class. There are no images which would mix instances of multiple classes.

\begin{table}[H]
  \caption[Bounding box statistics for ILSVRC2012 validation split]{Bounding box statistics for ILSVRC2012 validation split. "Area of BBox overlaps" counts image pixels covered by 2 or more bounding boxes, so overlaps of several BBoxes do not inflate this statistic.}
  \label{tbl:imgnet.bbox_stats_full}
      \begin{tabular}{c||c|c|cc|cc|cc}
        \toprule
        {instances} & \multirow{2}{*}{\#imgs} & \multirow{2}{*}{\#bboxes} & \multicolumn{2}{c|}{Area of BBox union} & \multicolumn{2}{c|}{Area of BBox overlaps} & \multicolumn{2}{c}{Area of a single BBox} \\
        {per img} & {} & {} & {mean} & {stddev} & {mean} & {stddev} & {mean} & {stddev} \\
        \midrule
        {1} & {38285} & {38285} & {48.67\%} & {28.52\%} & {0.0\%} & {0.0\%} & {48.66\%} & {28.52\%} \\
        {$2 \leq x < 5$} & {9337} & {22989} & {43.13\%} & {26.62\%} & {7.42\%} & {14.13\%} & {22.03\%} & {16.49\%} \\
        {$\geq 5$} & {2378} & {19203} & {42.24\%} & {28.37\%} & {11.07\%} & {16.09\%} & {7.78\%} & {7.18\%} \\
        \midrule
        {all} & {50000} & {80477} & {47.32\%} & {28.27\%} & {1.91\%} & {7.88\%} & {41.74\%} & {28.99\%} \\
        \bottomrule
      \end{tabular}
\end{table}

Table~\ref{tbl:imgnet.by_instance.reldiff} shows improved rel$\Delta$AuC for multi-instance images ($\#instances \geq 2$), in both constructive and destructive variant. However, at the same time, the $rel\Delta$ of Pointing Game deteriorates. There is no clear trend between improvement and the number of instances. In Sharpen game, the more instances the greater the improvement, while in Remove game trends differ between classifiers (although in both cases the improvement is greater for multi-instance cases), while Pointing Game exhibits an A-shaped trend.

Table~\ref{tbl:imgnet.by_instance.raw} reveals that raw scores of all quality metrics drop as the number of instances increases. However, presence of multiple class instances affects each metric in a different way.
For Alteration Game, more alteration steps are required to remove or re-introduce all salient features. Since AltGame measures how quickly salient features are affected by alterations, multiple instances act like redundant infrastructure, keeping the class confidence scores at a steady level despite the tampering, therefore negatively affecting AuC scores of both constructive and destructive variants.
In the case of Pointing Game, a decrease in hit rate - in response to increased instance count - appears outright paradoxical.
Intuitively, with more object instances in the image, hitting one of them should be more likely, even if only statistically.
However,~the~bounding~box~statistics~in~Table~\ref{tbl:imgnet.bbox_stats_full},~show~that the total area covered by bounding boxes is actually slightly smaller for multi-instance images. That is because the object instances themselves are more likely to be smaller than in images featuring only a single object.

\input{tbl/imgnet/by_instance_count.tex}
\input{tbl/imgnet/by_instance_count_multimodel_pmformat.tex}

\input{tbl/imgnet/raw_all_methods.tex}

A side-by-side comparison with results reported in the original article (\tblref{tbl:imgnet.all_methods}), shows that our results for both algorithms have significantly higher standard deviation, and the "sharpen" game scores are also noticeably worse.
Although we have carried out an additional analysis of results gathered at various stages of development of our solution, and managed to rule out: our result analysis code; the evaluation logic; use of FP16 mode; as potential root causes, the exact origin of this discrepancy remains unknown.
It's worth noting that the relationships between score categories in~\tblref{tbl:imgnet.all_methods} remain unchanged despite the difference in absolute values. Both the reproduced and original results show the same relative difference between quality scores for both models and the same preference towards a specific variant of Alteration Game. Therefore, since this discrepancy appears to be systematic, an additional set of artificially adjusted results has been included in this table.

The results of this experiment indicate that Voronoi meshes alone are not enough to universally and significantly improve results. The observed improvement of AltGame scores is rather modest and Pointing Game scores have actually deteriorated a bit. The difference in performance between single- and multi-instance cases, on the other hand, is still noticeable.

%% file: tbl/imgnet/by_instance_count.tex
\begin{table}[H]
\caption[Breakdown of r$\Delta$(avg(VRISE), avg(RISE)) by number of class instances in image]{Breakdown of r$\Delta$(avg(VRISE), avg(RISE)) by number of class instances in image. 'AltGame*' column negates rel$\Delta$ values of destructive games to adjust them to the 'higher is better' convention of other metrics.}
\label{tbl:imgnet.by_instance.reldiff}
\begin{tabular}{cc|cc|cc}
\toprule
{} & {metric >} & \multicolumn{2}{c|}{AltGame*} & \multirow[c]{2}{*}{PointingGame} & \multirow[c]{2}{*}{Consistency} \\
{model} & {instances} & {remove} & {sharpen} & {} & {} \\
\midrule
\multirow[c]{4}{*}{ResNet50} & {1} & 6.88\% & 1.09\% & -1.47\% & 0.00\% \\
 & {$2 \leq x < 5$} & 7.78\% & 1.34\% & -1.36\% & 0.00\% \\
 & {$\geq 5$} & 8.28\% & 1.72\% & -2.62\% & 0.00\% \\
 \cmidrule[.5pt]{2-6}
 & {all} & 7.14\% & 1.16\% & -1.50\% & 0.00\% \\
 \midrule
\multirow[c]{4}{*}{VGG16} & 1 & 8.29\% & 1.15\% & -1.83\% & -0.01\% \\
 & {$2 \leq x < 5$} & 8.86\% & 1.32\% & -1.38\% & -0.01\% \\
 & {$\geq 5$} & 8.03\% & 1.74\% & -2.25\% & -0.01\% \\
 \cmidrule[.5pt]{2-6}
 & {all} & 8.38\% & 1.20\% & -1.77\% & 0.00\% \\
\bottomrule
\end{tabular}
\end{table}

%% file: tbl/imgnet/by_instance_count_multimodel_pmformat.tex
\begin{table}[H]
\caption[Raw metric scores grouped by number of instances per image. (imagenet val split)]{Raw metric scores grouped by number of instances per image. Highlights indicate the better result in each RISE+VRISE pair of results. Results in range [0, 1].}
\label{tbl:imgnet.by_instance.raw}
  \resizebox{\textwidth}{!}{
\begin{tabular}{ccc||cc|cc|cc}
\toprule
{} & {} & {metric >} & \multicolumn{2}{c|}{AltGame} & \multicolumn{2}{c|}{PointingGame} & \multicolumn{2}{c}{Consistency} \\
{} & {} & {algorithm >} & \multirow{2}{*}{VRISE} & \multirow{2}{*}{RISE} & \multirow{2}{*}{VRISE} & \multirow{2}{*}{RISE} & \multirow{2}{*}{VRISE} & \multirow{2}{*}{RISE} \\
{game} & {model} & {instances} & {}  & {}  & {}  & {}  & {}  & {}  \\
\midrule
\multirow[c]{8}{*}{remove} & \multirow[c]{4}{*}{ResNet50} & 1 & \bfseries 0.0889 $\pm$ 0.0121 & 0.0955 $\pm$ 0.0141 & 0.8739 $\pm$ 0.0502 & \bfseries 0.8870 $\pm$ \bfseries 0.0394 & \bfseries 0.9996 $\pm$ \bfseries 0.0001 & 0.9996 $\pm$ 0.0001 \\
 &  & {$2 \leq x < 5$} & \bfseries 0.0959 $\pm$ \bfseries 0.0134 & 0.1039 $\pm$ 0.0152 & 0.8347 $\pm$ 0.0586 & \bfseries 0.8462 $\pm$ \bfseries 0.0482 & \bfseries 0.9996 $\pm$ \bfseries 0.0001 & 0.9996 $\pm$ 0.0001 \\
 &  & {$\geq 5$} & \bfseries 0.1211 $\pm$ \bfseries 0.0172 & 0.1320 $\pm$ 0.0195 & 0.7389 $\pm$ 0.0978 & \bfseries 0.7588 $\pm$ \bfseries 0.0842 & \bfseries 0.9996 $\pm$ \bfseries 0.0001 & 0.9996 $\pm$ 0.0001 \\
 \cmidrule[.5pt]{3-9}
 & & all & \bfseries 0.0917 $\pm$ \bfseries 0.0126 & 0.0988 $\pm$ 0.0145 & 0.8602 $\pm$ 0.0541 & \bfseries 0.8732 $\pm$ \bfseries 0.0432 & \bfseries 0.9996 $\pm$ \bfseries 0.0001 & 0.9996 $\pm$ 0.0001 \\
  \cmidrule[.5pt]{2-9}
 & \multirow[c]{4}{*}{VGG16} & 1 & \bfseries 0.0775 $\pm$ \bfseries 0.0113 & 0.0845 $\pm$ 0.0127 & 0.8586 $\pm$ 0.0550 & \bfseries 0.8746 $\pm$ \bfseries 0.0497 & 0.9992 $\pm$ 0.0002 & \bfseries 0.9993 $\pm$ \bfseries 0.0002 \\
 &  & {$2 \leq x < 5$}  & \bfseries 0.0850 $\pm$ \bfseries 0.0127 & 0.0933 $\pm$ 0.0142 & 0.8187 $\pm$ 0.0740 & \bfseries 0.8302 $\pm$ \bfseries 0.0605 & 0.9993 $\pm$ 0.0002 & \bfseries 0.9994 $\pm$ \bfseries 0.0002 \\
 &  & {$\geq 5$} & \bfseries 0.1146 $\pm$ \bfseries 0.0168 & 0.1246 $\pm$ 0.0180 & 0.7241 $\pm$ 0.1071 & \bfseries 0.7408 $\pm$ \bfseries 0.1054 & 0.9993 $\pm$ 0.0002 & \bfseries 0.9994 $\pm$ \bfseries 0.0001 \\
 \cmidrule[.5pt]{3-9}
 & & all & \bfseries 0.0807 $\pm$ \bfseries 0.0118 & 0.0880 $\pm$ 0.0132 & 0.8447 $\pm$ 0.0610 & \bfseries 0.8599 $\pm$ \bfseries 0.0544 & 0.9992 $\pm$ 0.0002 & \bfseries 0.9994 $\pm$ \bfseries 0.0002 \\
 \midrule
\multirow[c]{8}{*}{sharpen} & \multirow[c]{4}{*}{ResNet50} & 1 & \bfseries 0.6742 $\pm$ \bfseries 0.0144 & 0.6669 $\pm$ 0.0157 & 0.8739 $\pm$ 0.0502 & \bfseries 0.8870 $\pm$ \bfseries 0.0394 & \bfseries 0.9996 $\pm$ \bfseries 0.0001 & 0.9996 $\pm$ 0.0001 \\
 &  & {$2 \leq x < 5$} & \bfseries 0.6460 $\pm$ \bfseries 0.0153 & 0.6375 $\pm$ 0.0167 & 0.8347 $\pm$ 0.0586 & \bfseries 0.8462 $\pm$ \bfseries 0.0482 & \bfseries 0.9996 $\pm$ \bfseries 0.0001 & 0.9996 $\pm$ 0.0001 \\
 &  & {$\geq 5$} & \bfseries 0.6099 $\pm$ \bfseries 0.0171 & 0.5996 $\pm$ 0.0194 & 0.7389 $\pm$ 0.0978 & \bfseries 0.7588 $\pm$ \bfseries 0.0842 & \bfseries 0.9996 $\pm$ \bfseries 0.0001 & 0.9996 $\pm$ 0.0001 \\
 \cmidrule[.5pt]{3-9}
 & & all & \bfseries 0.6659 $\pm$ \bfseries 0.0147 & 0.6582 $\pm$ 0.0161 & 0.8602 $\pm$ 0.0541 & \bfseries 0.8732 $\pm$ \bfseries 0.0432 & \bfseries 0.9996 $\pm$ \bfseries 0.0001 & 0.9996 $\pm$ 0.0001 \\
 \cmidrule[.5pt]{2-9}
 & \multirow[c]{4}{*}{VGG16} & 1 & \bfseries 0.6131 $\pm$ \bfseries 0.0176 & 0.6062 $\pm$ 0.0189 & 0.8586 $\pm$ 0.0550 & \bfseries 0.8746 $\pm$ \bfseries 0.0497 & 0.9992 $\pm$ 0.0002 & \bfseries 0.9993 $\pm$ \bfseries 0.0002 \\
 &  & {$2 \leq x < 5$} & \bfseries 0.5824 $\pm$ \bfseries 0.0189 & 0.5748 $\pm$ 0.0201 & 0.8187 $\pm$ 0.0740 & \bfseries 0.8302 $\pm$ \bfseries 0.0605 & 0.9993 $\pm$ 0.0002 & \bfseries 0.9994 $\pm$ \bfseries 0.0002 \\
 &  & {$\geq 5$} & \bfseries 0.5484 $\pm$ \bfseries 0.0208 & 0.5390 $\pm$ 0.0216 & 0.7241 $\pm$ 0.1071 & \bfseries 0.7408 $\pm$ \bfseries 0.1054 & 0.9993 $\pm$ 0.0002 & \bfseries 0.9994 $\pm$ \bfseries 0.0001 \\
 \cmidrule[.5pt]{3-9}
 & & all & \bfseries 0.6043 $\pm$ \bfseries 0.0180 & 0.5971 $\pm$ 0.0192 & 0.8447 $\pm$ 0.0610 & \bfseries 0.8599 $\pm$ \bfseries 0.0544 & 0.9992 $\pm$ 0.0002 & \bfseries 0.9994 $\pm$ \bfseries 0.0002 \\
\bottomrule
\end{tabular}
}
\end{table}

%% file: tbl/imgnet/raw_all_methods.tex
\begin{table}
  \caption[Comparison of AltGame results - VRISE against other algorithms]{Comparison of AltGame results - VRISE against other algorithms. For "remove" lower is better, for "sharpen" higher is better. Results for other methods were carried over from \cite{rise}. Grad-CAM is the only white-box method in this table. Best results are marked in bold. Italic font marks the best result obtained by us. Dataset: ILSVRC2012 Val split. Results for GradCAM, LIME, RISE and Sliding Window are quoted from \cite{rise}}
  \label{tbl:imgnet.all_methods}
  \begin{tabular}{l|r|cc|cc}
    \toprule
    \multirow[c]{2}{*}{Method} & {Model >} & \multicolumn{2}{c|}{ResNet50} & \multicolumn{2}{c}{VGG16}  \\
    {} & {AltGame type >} & {remove} & {sharpen} & {remove} & {sharpen} \\
    \midrule
    \multicolumn{2}{l|}{Grad-CAM \cite{gradcam}} & {0.1232} & {0.6766} & {0.1087} & {0.6149} \\
    \midrule
    \multicolumn{2}{l|}{Sliding window \cite{slidingWindow}} & {0.1421} & {0.6618} & {0.1158} & {0.5917} \\
    \multicolumn{2}{l|}{LIME \cite{lime}} & {0.1217} & {0.6940} & {0.1014} & {0.6167} \\
    \multicolumn{2}{l|}{RISE \cite{rise}} & {0.1076 $\pm$ 0.0005} & {\bfseries 0.7267 $\pm$ 0.0006} & {0.0980 $\pm$ 0.0025} & {\bfseries 0.6663 $\pm$ 0.0014} \\
    \midrule
    \multicolumn{2}{l|}{RISE (reproduced)} & {0.0988 $\pm$ 0.0145} & {0.6582 $\pm$ 0.0161} & {0.0880 $\pm$ 0.0132} & {0.5971 $\pm$ 0.0192} \\
    \multicolumn{2}{l|}{VRISE (ours)} & {\itshape \bfseries 0.0917 $\pm$ 0.0126} & {\itshape 0.6659 $\pm$ 0.0147} & {\itshape \bfseries 0.0807 $\pm$ 0.0192} & {\itshape 0.6043 $\pm$ 0.0180} \\
    \midrule
    \midrule
    \multicolumn{2}{l|}{abs$\Delta$(reproduced, original)} & {-0.0088} &	{-0.0685} &	{-0.0100} &	{-0.0692} \\
    \multicolumn{2}{l|}{rel$\Delta$(reproduced, original)} & {-0.0818} &	{-0.0943} &	{-0.1020} &	{-0.1038} \\

    \midrule
    \multicolumn{2}{l|}{VRISE abs$\Delta$ adjusted} & {0.1005} & {0.7344} &	{0.0907} & {0.6735} \\
    \multicolumn{2}{l|}{VRISE rel$\Delta$ adjusted} & {0.0997} & {0.7352} &	{0.0899} & {0.6743} \\
    \bottomrule
  \end{tabular}
\end{table}

%% file: tex/evaluation/ssim_eval.tex
\section{Map consistency analysis}
\label{sec:ssim_analysis}

In this section, we use data from GridSearch to take a closer look at the influence of VRISE parameters on similarity of maps generated using the same sets of parameters (intra-set consistency), as well as the parameters' influence on the rate of convergence.

\begin{figure}[H]
  \centering
  \includegraphics[width=\textwidth]{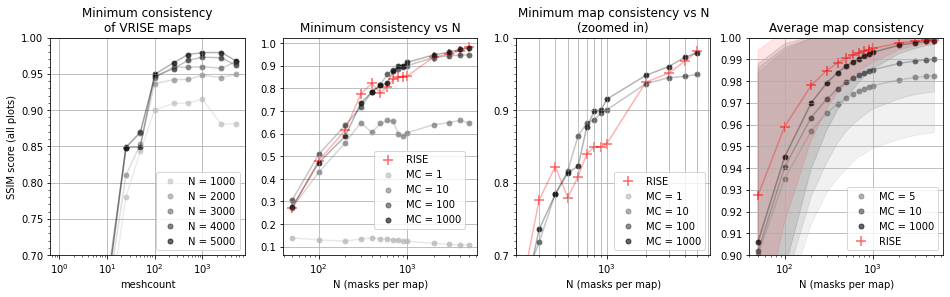}
  \caption[Map consistency vs meshcount]{Each datapoint aggregates all of SSIM scores measured between each pair of maps within each of corresponding paramsets. MC stands for \texttt{meshcount}. Gray series visualize results for VRISE. Tinted areas in the rightmost plot visualize standard deviation of each series. Note that the base of the Y-axis is different for each plot, to improve readability.}
  \label{fig:ssim_eval.gs_consistency}
\end{figure}

\figref{fig:ssim_eval.gs_consistency} shows relationships between the number of meshes and masks used per map, for RISE and VRISE. Each datapoint aggregates all of SSIM scores measured between every pair of maps within each of corresponding paramsets.

The leftmost plot shows the \textbf{lowest} value of SSIM across all results for each datapoint. \texttt{meshcount >= 100} yields very consistent maps of average minimal intra-set similarity of about 0.95. In other words, regardless of the choice of other parameters, no matter how good or bad the map might be in terms of AltGame and Pointing Game scores, when compared to other maps generated using the same set of parameters, it will be at least 95\% similar.

The two plots in the center of~\figref{fig:ssim_eval.gs_consistency} visualize the same aggregation as the leftmost plot but swap the X axis with data series. It's worth noting how incredibly poor is the consistency of maps generated from a single random mesh, as well as how much this metric improves when only a few additional meshes are used. At $\texttt{meshcount} > 100$, VRISE maps catch up to those generated by RISE in terms of consistency. The fact that multiple random Voronoi meshes can achieve consistency of results close to that of a grid mesh confirms the expected regularization properties of the multi-mesh approach.

The rightmost plot of~\figref{fig:ssim_eval.gs_consistency} shows \textbf{averages} of data series in central plots against the number of masks coloured bands visualize standard deviation of results. Although RISE scores worse in terms of minimum consistency, it is consistently better on average.

\begin{figure}[H]
  \centering
  \includegraphics[width=\textwidth]{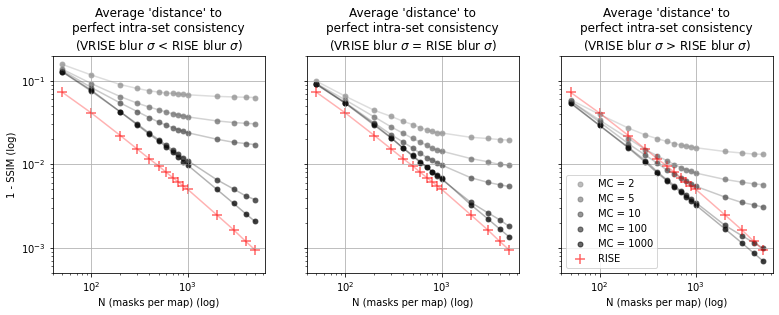}
  \caption[Distance to perfect similarity for all VRISE paramsets and only those matching RISE]{Both plots show the "distance" to perfect intra-set consistency for VRISE and RISE. $\sigma$.}
  \label{fig:ssim_eval.gs_consistency_better_subset}
\end{figure}

Fig.~\ref{fig:ssim_eval.gs_consistency_better_subset} is a modified view of the rightmost plot from fig.~\ref{fig:ssim_eval.gs_consistency}, and focuses on differences between VRISE and RISE at higher values of $N_{masks}$, which were squashed by the linear scale of the Y-axis. Additionally, it splits the result dataset into three groups - those using weaker, comparable, and stronger blur than RISE, showing how this additional parameter influences consistency. It clearly shows that higher \texttt{meshcount} increases the limit of $N_{masks}$ beyond which the intra-set consistency stops improving, while blur $\sigma$ improves consistency of all results on average, regardless of $N_{masks}$.

\begin{figure}[H]
  \centering
  \includegraphics[width=\textwidth]{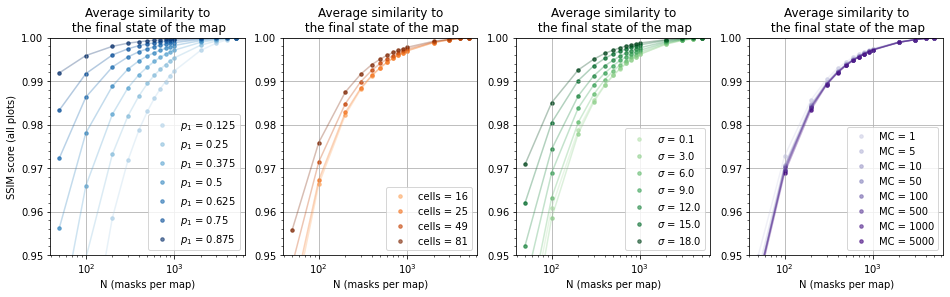}
    \caption[Intra-set consistency characteristic for configurable VRISE parameters]{Intra-set map consistency characteristics for each configurable parameter, plotted against the number of evaluated occlusion masks. Results for $N=5000$ masks per map.}
  \label{fig:ssim_eval.convergence_vs_param}
\end{figure}

\begin{figure}[H]
  \centering
  \includegraphics[width=0.6\textwidth]{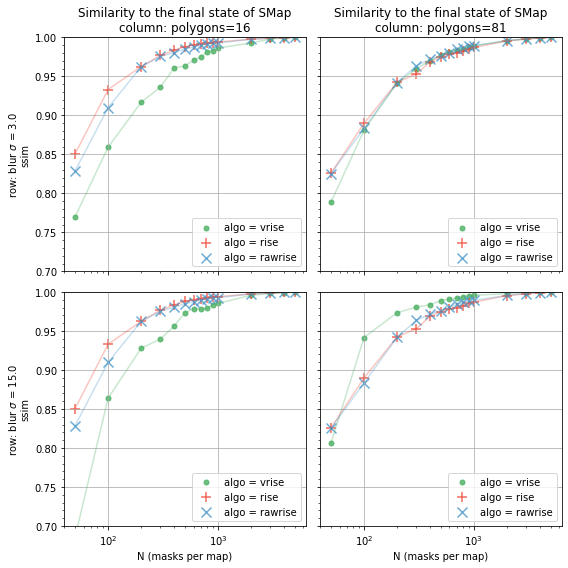}
  \caption[Convergence rate comparison for VRISE and RISE on selected parameter sets]{Each series shows average SSIM similarity between saliency maps at X masks and their ultimate state consisting of 5000 masks. "rawrise" series contain results for original RISE, while "rise" series are based on RISE using a Hybrid Grid Generator with mask informativeness guarantee. RISE series are identical within columns. $p_1$ = 0.125 for all series and \texttt{meshcount} = 1000 for all VRISE series.}
  \label{fig:ssim_eval.gs_conv_rawrise}
\end{figure}

\figref{fig:ssim_eval.gs_conv_rawrise} compares convergence rate of VRISE and RISE maps. A higher polygon count is shown to positively impact the convergence rate and allows VRISE to achieve a level of intra-set map consistency equal to that of RISE. With increased blur strength, VRISE can actually converge even faster than RISE (bottom-right plot). However, GridSearch results show that VRISE performs worse in terms of AltGame and Pointing game under these conditions (fig.~\ref{fig:grid_search.catdog_aucdiff_heatmap}).

Another interesting occurrence, captured in the upper-left plot of fig.~\ref{fig:ssim_eval.gs_conv_rawrise}, is the difference in convergence rates of RISE and "raw RISE". A mask generated from 16-cell grid with 12.5\% chance of cell visibility has 11\%~chance of being fully occluded and uninformative. This causes the map generated by unmodified RISE ("rawrise" series) to converge slower than RISE equipped with a HGG ("rise" series) and provides an example of the effectiveness of changes proposed in sec.~\ref{sec:guaranteed_grid_gens}.

\figref{fig:ssim_eval.convergence_vs_param} shows convergence rate is influenced primarily by $p_1$, followed by blur $\sigma$, both parameters also influencing the total area of occlusions.
$p_1$ directly determines what percentage of the input image will be occluded, while blur $\sigma$ effectively acts as a multiplier of $p_1$.
This indicates that the less of the original image is occluded, the quicker the map will converge, and maps from independent runs will be more similar to one another, although these properties will come at the expense of other quality metrics, as shown in fig.~\ref{fig:grid_search.rawauc_origame} and \ref{fig:grid_search.pointing_game_rawscore}.
An extreme but intuitive example of rapid convergence at high $p_1$ would involve $p_1 = 100\%$. The very first mask would produce the ultimate state of this saliency map, although it would also contain no information about the location of salient features, because not a single pixel would have been occluded during evaluation.
\texttt{meshcount} is shown to have almost no effect on the rate of convergence.

\begin{figure}[H]
  \centering
  \includegraphics[width=\textwidth]{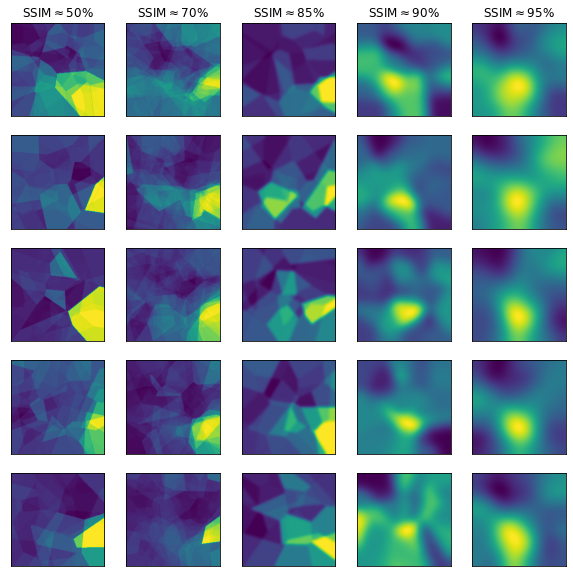}
  \caption[Demonstration of map sets of varying similarity]{Demonstration of visual saliency maps of varying levels of consistency. Each column contains a set of VRISE maps achieving intra-set mutual similarity (consistency) stated in the header.}
  \label{fig:ssim_eval.consistency_map_demo}
\end{figure}

\begin{figure}[H]
  \centering
  \includegraphics[width=0.9\textwidth]{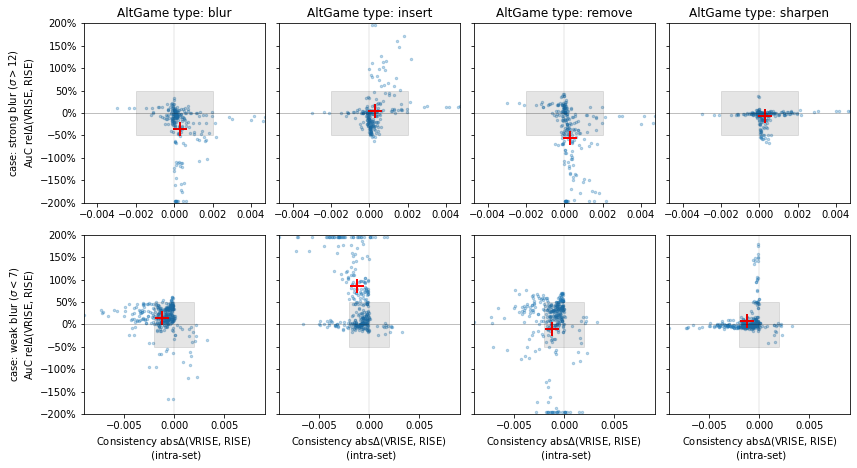}
  \caption[rel$\Delta$AuC plotted against abs$\Delta$SSIM]{rel$\Delta$AuC plotted against abs$\Delta$SSIM. Y-axis is truncated to improve readability and exceeding datapoints are shifted to horizontal edge. Red cross marks the mean of all points (pre-clipping). \texttt{meshcount} = 1000, $N$ > 3000.}
  \label{fig:ssim_eval.ssim_vs_auc}
\end{figure}

\figref{fig:ssim_eval.ssim_vs_auc} visualizes relationships between improvement in terms of SSIM and AuC (or lack thereof).
The bottom row of plots, for maps using weak occlusion blur, shows an almost complete absence of results in the first quadrant, which would improve both AuC and consistency.

\begin{figure}[H]
  \centering
  \includegraphics[width=0.9\textwidth]{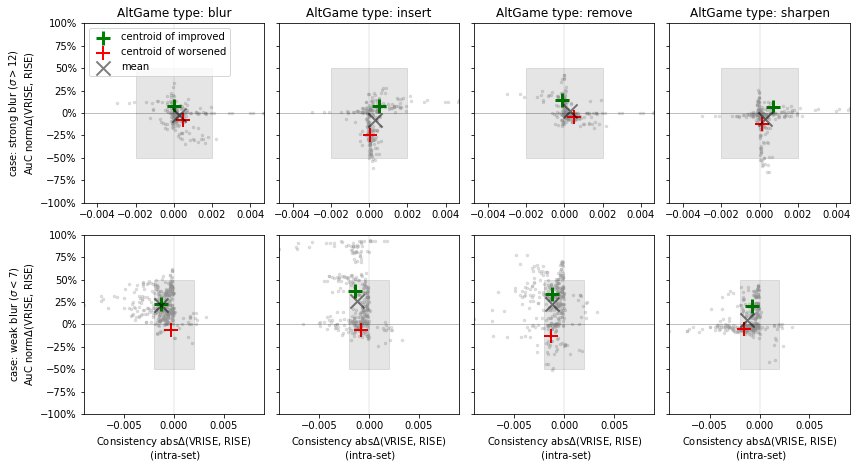}
  \caption[norm$\Delta$AuC plotted against rel$\Delta$SSIM]{norm$\Delta$AuC plotted against rel$\Delta$SSIM. The "mean" centroid is the mean of all points. \texttt{meshcount} = 1000, $N$ > 3000.}
  \label{fig:ssim_eval.ssim_vs_auc_normdiff}
\end{figure}

\figref{fig:ssim_eval.ssim_vs_auc_normdiff} presents the same set of data as \figref{fig:ssim_eval.ssim_vs_auc}, using norm$\Delta$ instead of rel$\Delta$ on the Y-axis (see sub-section~\ref{subsec:norm_diff}). Due to the characteristic of this metric, the plot displays separate median centroids for subsets of results which improve and which worsen the AuC, in addition to the mean centroid of the entire dataset.

Both of these scatterplot sets reveal a tradeoff between AuC score and consistency, dependent on blur strength. Maps using stronger blur are not significantly more consistent on average, but those which improve consistency usually do not significantly improve AuC. Maps using weaker blur attain greater AuC improvement on average, but hardly any of them score higher in terms of consistency.

%% file: tex/theory/feature_slicing.tex
\subsection{Feature slicing}
\textit{Feature slicing} occurs when an edge of the occlusion mesh splits a salient feature, and reduces its recognizability. This negatively impacts the saliency score of cells containing parts of such split feature, as they are less likely to be considered salient on their own and are also less likely to be sampled together.

\begin{figure}[H]
  \centering
  \includegraphics[width=\textwidth]{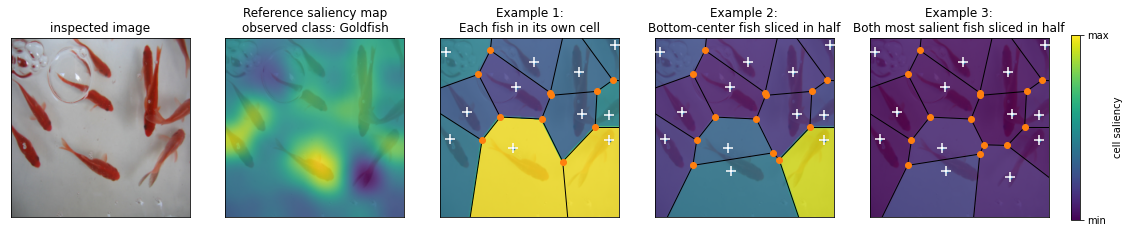}
  \caption[A demonstration of feature slicing, absolute value range]{An example of the feature slicing effect. Notice how the saliency of the fish instances on the bottom drops as they are split by the diagram's edges. (White crosses mark Voronoi seeds. All maps use $p_1=0.125$. $N_{masks}=1000$. Color scale is absolute and common to all maps, highlighting how confidence scores dwindle as the features are sliced)}
  \label{fig:vrise_description.feature_slicing_absolute}
\end{figure}

\begin{figure}[H]
  \centering
  \includegraphics[width=\textwidth]{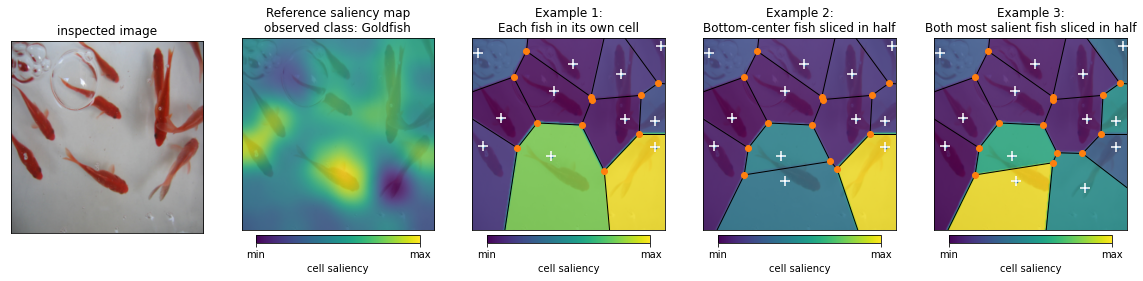}
  \caption[A demonstration of feature slicing, relative value ranges]{An example of the feature slicing effect. Notice how the location of most salient cell changes with each additional slice. (White crosses mark Voronoi seeds. All maps use $p_1=0.125$. $N_{masks}=1000$. Each map uses a separate color scale, highlighting saliency relative to rest of the map)}
  \label{fig:vrise_description.feature_slicing_relative}
\end{figure}

Figures \ref{fig:vrise_description.feature_slicing_absolute} and \ref{fig:vrise_description.feature_slicing_relative} demonstrate the impact of feature slicing on reported saliency of individual instances of the "goldfish" class. The Voronoi mesh gains an extra seed in each example, which is deliberately positioned in a way which will cause a Voronoi edge to appear over the currently most salient class instance.

In \figref{fig:vrise_description.feature_slicing_relative}, example \#1 shows a handcrafted Voronoi mesh, in which every instance of a goldfish is contained in its own separate cell, ensuring that each fish is always presented in its entirety to the classifier.
It also clearly shows which of the instances are recognized with greatest confidence, as cell saliency score is proportional to the classifier's confidence score.
Notice how slicing the most salient instance in example \#1 causes the map in example \#2 to report the previously second most salient fish instance as the now most salient, while at the same time the split instance appears to be less salient than it was previously.
Another slice applied to the bottom-right instance shifts the focus back to one of the parts of the central instance, as it appears to contain the most convincing representation of a goldfish out of all other cells.

Figure \ref{fig:vrise_description.feature_slicing_absolute} compares saliency maps from the previous example to one another. The most striking effect is the apparent lack of salient features in example \#3, even though both example \#1 and the reference map clearly show that the image contains at least two very prominent features.

Random shifts used by RISE as well as multiple Voronoi meshes in VRISE reduce the severity of this effect by relocating edges of the occlusion mesh. RISE accomplishes that through random shifts of the occlusion mask, while VRISE aims to mitigate this effect through use of multiple meshes.

The primary adverse effect of feature slicing, especially in multi-instance case, is that secondary features may be indicated as the most salient only because the "actual" primary feature was handicapped during the map generation process.
The secondary negative effect becomes apparent only when maps are compared to each other, as shown in \figref{fig:vrise_description.feature_slicing_absolute}. Despite the exact same set of parameters and numerous evaluations, an unfortunate layout of occlusions in example \#3 yields a misleading saliency map.

%% file: tex/theory/saliency_misattribution.tex
\subsection{Saliency misattribution}
\label{ch:saliency_misattribution}

Saliency misattribution occurs when both salient and non-salient areas co-occur in a masked image. The saliency score contributed by the relevant areas is attributed to the entirety of the visible area, including regions which did not contain salient features, unduly increasing saliency of the latter.
This effect is unavoidable in random sampling methods, although it diminishes as the number of samples approaches infinity and the saliency map converges to its final state. Avoiding it entirely would require prior knowledge of placement of the salient features, in which case an explanation algorithm would not be needed at all.

\begin{figure}[H]
  \centering
  \includegraphics[width=\textwidth]{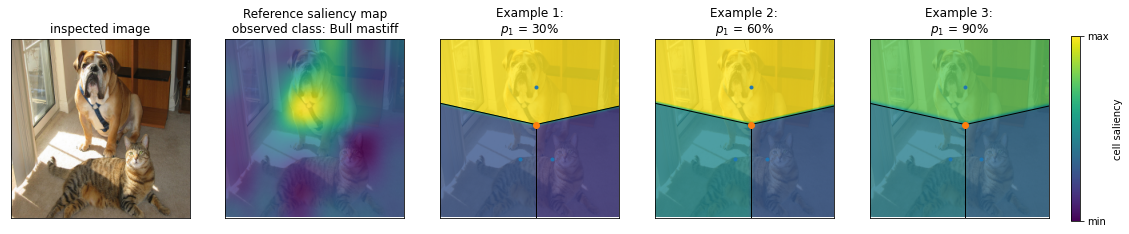}
  \caption[A demonstration of saliency misattribution (absolute)]{An example of saliency misattribution. All maps use a common, \textbf{absolute} color scale. The difference in saliency values between relevant and irrelevant cells shrinks as the frequency of their co-occurrence increases.}
  \label{fig:vrise_description.saliency_misattribution_demo_absolute}
\end{figure}

The toy example presented in \figref{fig:vrise_description.saliency_misattribution_demo_absolute} induces this effect by gradually increasing $p_1$, which increases the frequency of co-occurrence of the relevant top cell and irrelevant bottom cells in occlusion masks. At $p_1=0.9$, each image cell has a 90\% chance of being visible during evaluation. High $p_1$ is not required for misattribution to occur, but it conveniently intensifies this effect for the purpose of this demonstration.
The bottom-left cell is an example of irrelevant cell containing no features salient to the observed class, according to the reference saliency map. However, as the frequency of its co-occurrence with the top cell increases, it becomes increasingly difficult for the algorithm to determine which of these cells is actually contributing to confidence score and causing misattribution.

The bottom-right cell resists misattribution of saliency for the "dog" class because it contains features salient to "cat" and because the model under evaluation uses a softmax output layer. When the bottom-right cell is visible together with the top one, the features it introduces "steal" some of the confidence score from the "dog" class because the vector of confidence scores must always sum to one.
In consequence, whenever the bottom-right cell is not masked, the dog class scores lower than usual and that is reflected in low saliency score of this cell.

This example shows that "neutral" regions of the image which don't contain salient features of any of recognized classes are the most prone to saliency misattribution.
\figref{fig:vrise_description.saliency_misattribution_demo_absolute} compares saliency maps to one another by using an absolute color scale. The inverse relationship between saliency of the top cell and values of $p_1$ isn't an effect of saliency misattribution, but rather a combination of frequent co-occurrence of cat and dog cells (which decreases confidence scores for both) and the final step of map generation process, which adjusts the map's values by $p_1$, by performing an elementwise multiplication with $\frac{1}{p_1}$. The decreasing relative difference between cells, however, is a result of misattribution.

\figref{fig:vrise_description.feature_slicing_absolute} happens to demonstrate the saliency misattribution effect as well. The examples in this figure contain 11, 12 and 13 cells, respectively. With $p_1 = 0.125$, the majority of images presented to the classifier contain from 1 to 3 cells. As recognizability of the most prominent features is reduced by slicing, the maps dim, implying that cells in the upper half of the saliency map actually owe their score to co-occurrence with the two most salient cells on the bottom.
If any of the cells in the upper half of the map were highly salient on its own, it would retain its initial saliency value, regardless of the state of other cells.

In maps using a higher number of cells ($>100$), misattribution contributes to visual noise in maps.

\begin{figure}[H]
  \centering
  \includegraphics[width=\textwidth]{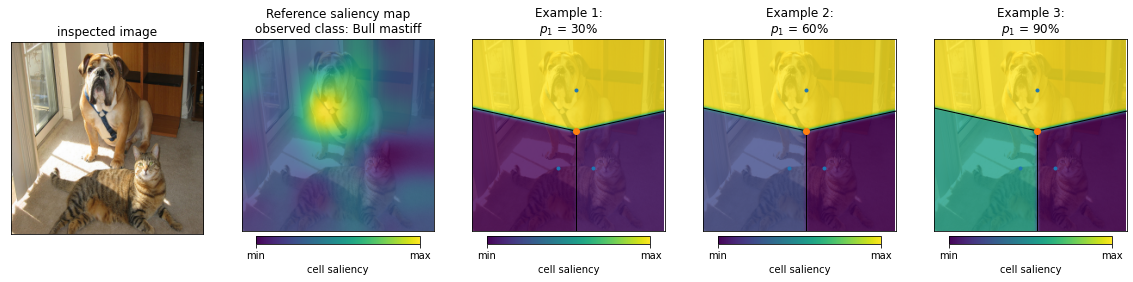}
  \caption[A demonstration of saliency misattribution (relative)]{An example of saliency misattribution. Each map uses a \textbf{relative} color scale adjusted to its own $min$ and $max$. Saliency of the top cell is misattributed to the bottom-left cell as increasing $p_1$ causes them to co-occur within masks more frequently.}
  \label{fig:vrise_description.saliency_misattribution_demo_relative}
\end{figure}

%% file: tex/evaluation/_outro.tex
\section{Discussion of results}
\label{sec:eval.outro}

\subsection{Discussion of observed effects}
\label{subsec:eval.outro.effects}

Results of GridSearch show that the relationship between blur $\sigma$, $p_1$ and its influence on quality of maps is quite complex compared to blur's interactions with other parameters, which tend to be monotonic and positively influence maps as blur weakens.
Figure~\ref{fig:grid_search.rawscore_sigma_VS_p1_narrowed}, in particular the subplots for "sharpen" variant of AltGame, presents an example of blur $\sigma$ and $p_1$ being at odds with one another.
Weak blur performs slightly worse when fill rate is low and significantly better when fill rate is high.
When blur strength is high, this relationship is inverted.
The optimal configuration appears to be a product of both parameters, and achieve a very similar quality of results over a broad range of parameter values as long as the pairing as a whole is optimal.
This suggests that the percentage of \textit{usable image area} (i.e. partially occluded (dimmed) and still bright enough to provide additional information to the classifier) has more influence on the final shape of the saliency map than either of these two parameters alone.

 \figref{fig:gs_appendix.catdog_demo_p1_sigma} provides a suitable example.
 Saliency maps along diagonals in the lower half of the map grid tend to be visually similar in terms of average background saliency and prominence of low-saliency areas, even though values of $p_1$ and blur $\sigma$ vary significantly.

\begin{figure}[H]
   \centering
   \includegraphics[width=\textwidth]{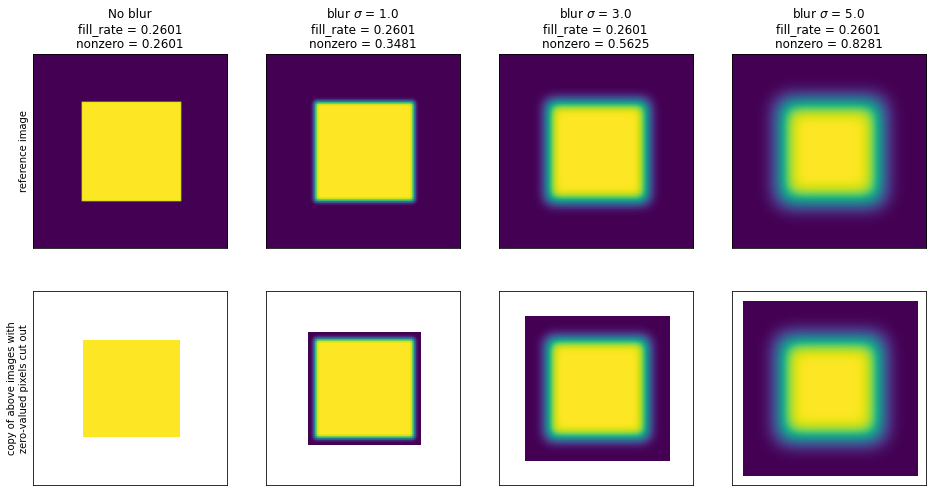}
   \caption[Demonstration of various blur strengths (blur $\sigma$)]{Demonstration of various blur strengths, fill rate and reach of the non-zero values after application of Gaussian blur. Images in the bottom row are copies of images from the top row, with zero-valued pixels left transparent, showing the actual area affected by Gaussian blur.}
   \label{fig:blur_strength_demo_cropped}
 \end{figure}

 For context, application of blur does not affect fill rate measured as mean value of the masking array, but it does increase the percentage of non-zero cells (demonstrated in fig.~\ref{fig:blur_strength_demo_cropped}). Some of these "dim" pixels might still be bright enough to inform the classifier, effectively increasing the percentage of total image area visible after occlusion, beyond what $p_1$ would indicate.

 \begin{minipage}[r]{0.5\textwidth}

  An ad-hoc evaluation - performed to substantiate this claim - used 50 random images from ILSVRC2012 and dimmed them via multiplication with a value in range $[0, 1]$.
  Full confidence score vectors were gathered for each brightness percentage point between 1\% and 100\%.
  The rank of the top1 class (according to dataset annotations) was tracked across results for all brightness levels.
  The results show that even 40\% of original brightness is enough to restore the rank of the top1 class in 90\% of cases, and as little as 20\% put the observed class among the top10 classes with highest confidence score.

 \end{minipage}
 \hspace{0.05\textwidth}
 \begin{minipage}[r]{0.3\textwidth}
     \raggedright\footnotesize{Minimum image brightness required to restore the confidence score of top1 class to top10/top1 results, for at least $q$\% of examples:}

     \vspace{3mm}
     \resizebox{\textwidth}{!}{
       \begin{tabular}{c|cc}
         \toprule
         \multirow[c]{2}{*}{$k$} & \multicolumn{2}{c}{Brightness $P_{k\%}$} \\
         {} & {top10} & {top1} \\
         \midrule
         {90\%} & {23\%} & {40\%} \\
         {95\%} & {30\%} & {59\%} \\
         {99\%} & {62\%} & {78\%} \\
         {100\%} & {85\%} & {91\%} \\
         \bottomrule
       \end{tabular}
     }
 \end{minipage}

 However, the actual \textit{Total usable area} is difficult to predict ahead of time, as it is influenced by more factors than just $p_1$ and blur $\sigma$. The shape and size of individual occlusions, as well as how they are distributed across the image, also influence the percentage of image area which remains informative during evaluation.
 For instance, small, scattered occlusions require much weaker blur to make the entirety of the image usable, than a huge occlusion cluster would (fig.~\ref{fig:blur_strength_demo_cropped}).
 Additionally, the evaluated classifier's capacity to fill in gaps in the image also influences how all other factors contribute to the percentage of total usable area. For instance, in case of very small occlusions, like 1px random noise, the entirety of the image would become usable even without blurring at $p_1 \approx 50$.

\begin{figure}[H]
   \centering
   \includegraphics[width=0.75\textwidth]{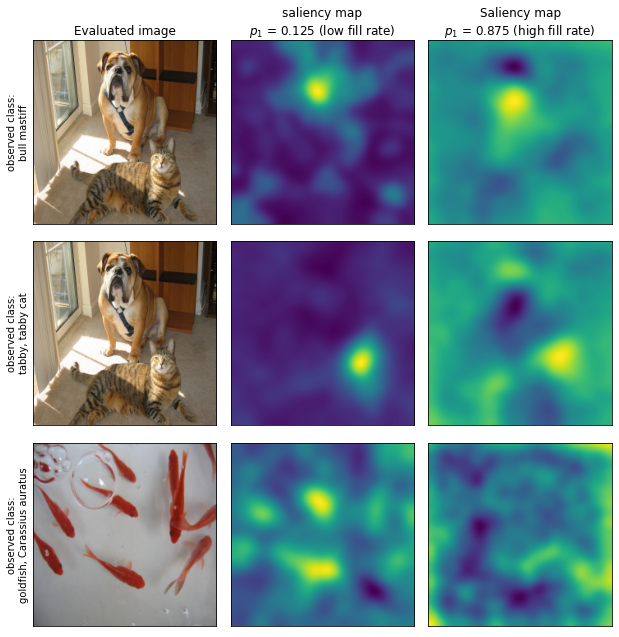}
   \caption[Comparison of high- and low-$p_1$ SMaps]{Note that high $p_1$ maps mark features accurately but not as precisely as low-$p_1$ maps. The dark areas in high $p_1$ maps tend to contain features of other classes, and (due to softmax) reduce confidence of the observed class. In "bull mastiff" map, the eyes and snout increase confidence for "Shih-Tzu" and "Pug", at the expense of "Bull Mastiff".}
   \label{fig:low_p1_high_p1_map_demo}
 \end{figure}

  Another effect associated with the combination of high values of $p_1$ and blur $\sigma$ is occurrence of highly concentrated low-saliency areas, sometimes right next to highly-salient ones.
 \figref{fig:low_p1_high_p1_map_demo} provides an example of this effect and offers a potential explanation of low quality scores of high $p_1$ maps. In the "bull mastiff" example, the high $p_1$ map contains a significant depression in the area where the dog's eyes and snout are located. Since this part of the image has been identified as the least salient area, Sharpen AltGame will restore it at the very end. Additionally, GridSearch has shown that low-$p_1$ maps, which do not exclude the eyes and snout area, do not suffer the same negative impact on scores as their high-$p_1$ counterparts. These two pieces of information suggest that the eyes and snout are in fact necessary to confidently recognize the Bull Mastiff class.
 The scores of Remove Game might be impaired for the same reason. Again, the face area will be the last to be removed from the image, upholding the confidence score until the very end and negatively affecting the AuC.

 An analysis of top-K classes in this area has revealed that the eyes and snout are also indicative of other short-faced dog breeds like Shih-Tzu and Pug, which leads us to the following conclusion: if the fill rate is high, and the classifier uses a softmax layer (as it did in our study), both RISE and VRISE become very sensitive to \textit{distinctive} features of the observed class, even if they are not sufficient for confident recognition of the object. Saliency score of these features is inflated at the expense of features shared with other classes, inadvertently impairing the quality of saliency maps.

\subsection{Parameters versus map quality}
\label{sec:outro.params_vs_quality}

Conclusions regarding relationships between parameters, across experiments:
The number of random Voronoi meshes used during mask generation (\texttt{meshcount}) has a positive, monotonic and asymptotic influence on map quality, although increasing the value of this parameter also linearly increases the duration of the occlusion generation step, slowing down the entire map generation process. This parameter could be excluded from future evaluations, or limited to 2 or 3 values within the range $[0, 1000]$ to make space for denser sampling of other parameters or a larger set of images. At least 100 meshes per map are required to make up for the lack of random shifts and close the quality gap between the no-shift multi-mesh approach of VRISE and the shifted single-mesh of RISE, as discussed further in sec.~\ref{sec:ssim_analysis}. We recommend 500 as default.

The performance with respect to occlusion mesh density (\texttt{polygons}) is the one aspect where VRISE does improve map quality, in particular when the mesh consists of a few large occlusions or the evaluated image contains multiple instances of the observed class. Results presented in sec.~\ref{sec:targeted_s} indicate that this particular trait is a consequence of multi-mesh approach and randomized occlusion shapes.

The percentage of occlusions left unused in each mask ($p_1$) should be kept in range 20\%-50\%. When too much of the image is visible, the saliency attribution algorithm has difficulties identifying the location of salient features. On the other end of the spectrum, very low $p_1$ values slow down convergence, decrease consistency, and negatively affect constructive AltGame, but not Pointing Game.

The relationship between the strength of blur applied to occlusions (blur $\sigma$) and map quality is parabolic - too much or too little will negatively impact measurement by both Alteration Game and Pointing Game. Our solution achieves higher result quality when blur is configured to be weaker than the default used by RISE for the same mesh density. On this end of the spectrum, only very low values ($<1$) decrease result quality.

\subsection{Cross-experiment summaries}

The assumptions made in sec.~\ref{ch:grid_search}, regarding improved performance on multi-instance images and in low-polygon configurations, have held up in the fixed blur experiment (sec.~\ref{sec:targeted_s}).
VRISE does perform better on multi-instance images than on single-instance ones, without the aid of decreased blur strength, as shown in tables~\ref{tbl:targeted_s.rawscores.chk}, \ref{tbl:targeted_s.reldiff.chk} and \ref{tbl:targeted_s.reldiff.params_instances}.

The trend for $p_1$ from sec.~\ref{ch:grid_search} was not observed in sec.~\ref{sec:targeted_s}, but on the other hand, two sampling points might have been too few to determine to what extent these two trends would align.

A comparison of quality scores provided by AltGame and Pointing Game with those yielded by Consistency metric shows that similarity metrics do provide additional information about map quality, but they should not be used as the only metric. For instance, high blur $\sigma$ positively correlates with similarity metrics, as shown in figures~\ref{fig:ssim_eval.gs_consistency_better_subset} and \ref{fig:ssim_eval.convergence_vs_param}, but at the same time it correlates negatively with the other two metrics (fig.~\ref{fig:grid_search.rawauc_origame} and \ref{fig:grid_search.pointing_game_rawscore}).

    The side-by-side comparison of results for GridSearch and FixedSigma in tbl.~\ref{tbl:eval.2way} indicates that images used in GridSearch evaluation are not as representative of their respective groups (of single/multi-instance examples) as we expected.
    However, some trends from the GridSearch experiment, regarding performance w.r.t. to mesh density reappear results of FixedSigma (tbl.~\ref{tbl:eval.2way}). For instance, the improved performance in destructive game or negative correlation between mesh density and improvement in Pointing Game
    (fig.~\ref{fig:grid_search.pointing_game_rawscore}, \ref{fig:grid_search.pointing_game_rawscore_better_subset}, \ref{fig:grid_search.rawscore_sigma_VS_polygons_narrowed}).
    Some observations pertaining to trends present in results of GridSearch might be generalizable, but an evaluation on a larger dataset would be required to state them with confidence.

      Zeros in Pointing Game column for GridSearch, in tables~\ref{tbl:eval.2way} and~\ref{tbl:eval.3way_fewmany} are caused by both algorithms having achieved 100\% accuracy in this particular parameter configuration (see PG results for $\texttt{polygons} = 49$ in fig.~\ref{fig:grid_search.rawscore_sigma_VS_p1_narrowed_fixedblur} and~\ref{fig:grid_search.rawscore_sigma_VS_polygons_narrowed_fixedblur})

    \input{tbl/eval_generic/2way_rdiff_of_avg.tex}
    \input{tbl/eval_generic/3way_rdiff_of_avg_fewmany.tex}

    Both FixedSigma and ImgNet evaluated the effectiveness of Voronoi meshes alone, in isolation from the influence of freely adjustable blur strength. Their results are consistent in terms of presence or lack of improvement (tbl.~\ref{tbl:eval.3way_fewmany}) and even though VRISE did not succeed in improving results of Pointing Game, it did increase the average result of both variants of AltGame by up to 10\%.
    It's likely that the general trend of performance w.r.t. \texttt{polygons} + blur~$\sigma$, observed in the FixedSigma experiment, would hold if these particular configurations were evaluated on the entirety of the ILSVRC2012 Val split.
    Moreover, since the results gathered in the ImageNet evaluation tend to be better than their counterparts from FixedSigma, a re-evaluation of those parameter configurations on the full dataset could yield even higher improvement values.

    Finally, the results of ImageNet (sec.~\ref{sec:imgnet}) allow us to confidently assert whether our solution offers better performance than RISE. Table~\ref{tbl:imgnet.all_methods} shows that VRISE outperformed RISE in both variants of AlterationGame, in all instance quantity subgroups and for both classifiers, with average improvement ranging from 5\% to 8\% in destructive game and 1\% to 2\% in constructive game.
    However, our solution just as broadly underperformed in Pointing Game, delivering up to 2\% worse results in every configuration.
    Differences in inter-run map consistency were negligible, primarily due to high $N_{masks}$ setting used both in the original article and here.

\subsection{Reflection on goals}
\label{sec:reflection_on_goals}
We now move on to reassess our results from the perspective of our goals:
\begin{enumerate}
  \item Voronoi occlusions (sec.~\ref{ch:vrise_description}):
  \begin{itemize}
    \item Improved map quality for low-density meshes (low \texttt{polygons}) - \textbf{successful}. Results in sections~\ref{ch:grid_search} and \ref{sec:targeted_s} have shown that VRISE indeed increases the accuracy of maps in low-polygon configurations. Map consistency is decreased, but only slightly.

    \item Improved map quality for multi-instance images - \textbf{successful}. Results in tbl.~\ref{tbl:eval.3way_fewmany} show that improvement in FixedSigma and ImgNet is higher for subsets of the dataset which consist of multi-instance images ("few" and "many" rows). Although PointingGame scores are impaired in both cases, the impairment is also smaller in multi-instance cases.

    \item Synergistic improvement in low-\texttt{polygon}, multi-instance cases - \textbf{successful}. Comparison of FixedSigma results in tbl.~\ref{tbl:eval.2way}, between subsets of results for single- and multi-instance images shows that: first, improvement of AltGame scores is even greater for multi-instance subset; second, PointingGame scores improve as well, but only in low-\texttt{polygon} configurations. These results demonstrate that Voronoi occlusions can generate more accurate saliency maps than square occlusions when average area of an individual occlusion is large.

    \item No decrease in inter-run saliency map consistency when each mask is generated from an unique Voronoi mesh - \textbf{unsuccessful}. The results presented in sec.~\ref{sec:ssim_analysis} have shown that our multi-mesh approach comes very close to random shifts in terms of consistency of results, but it is not capable of outperforming shifts. Fig.~\ref{fig:ssim_eval.gs_consistency_better_subset} shows that increasing the size of Voronoi mesh pool does allow our solution to reach higher levels of result consistency, but even using a separate unique mesh for every single occlusion pattern is not enough to exceed random shifts in this aspect. What has proven to be capable of directly increasing the consistency of maps, however, is blur strength. When VRISE is configured to exceed the intensity of blur used by RISE, the consistency of VRISE maps is greater than that of RISE maps. However, excessive blur strength was shown to have adverse affects on map accuracy measured by AltGame and PointingGame in sec.~\ref{ch:grid_search}. Conversely, reducing blur strength in order to improve map accuracy would increase variance of results. Therefore, blur needs to be used in moderation.

    \item Improved performance on a large dataset - \textbf{partial success}. Results of ImgNet (sec.~\ref{sec:imgnet}) have shown a modest improvement in both variants of AltGame, but also a slight decrease of accuracy in PointingGame. Although there are differences in magnitude of relative difference between both methods, they are consistent in terms of presence (or lack) of improvement.
  \end{itemize}
  \item Mask informativeness guarantee
  \begin{itemize}
    \item Increased initial map convergence rate (sec.~\ref{sec:similarity_metrics}) in cases where values of both \texttt{polygons} and $p_1$ are low - success.
    This outcome has been observed both in synthetic examples in (sec.~\ref{sec:guaranteed_grid_gens}) as well as in results gathered for the purposes of another experiment (fig.~\ref{fig:ssim_eval.gs_conv_rawrise}).
  \end{itemize}

\end{enumerate}

%% file: tbl/eval_generic/2way_rdiff_of_avg.tex
\begin{table}[h]
  \caption[Comparison of results between GridSearch and Fixed~$\sigma$, for common paramsets]{
    Side-by-side comparison of metrics rel$\Delta$(avg(VRISE), avg(RISE)) for GridSearch and Fixed~$\sigma$.
    Remove game scores are negated to match the "higher is better" convention.
    $p_1=0.5; \texttt{meshcount}=1000; N_{masks}=max$
  }
  \label{tbl:eval.2way}
  \resizebox{\textwidth}{!}{
    \begin{tabular}{cc|cccc|cccc}

      \toprule
      \multicolumn{2}{c|}{experiment >} & \multicolumn{4}{c|}{GridSearch} & \multicolumn{4}{c}{Fixed $\sigma$} \\
      \multicolumn{2}{c|}{\texttt{polygons} >} & \multirow[c]{2}{*}{16} & \multirow[c]{2}{*}{25} & \multirow[c]{2}{*}{49} & \multirow[c]{2}{*}{81} & \multirow[c]{2}{*}{16} & \multirow[c]{2}{*}{25} & \multirow[c]{2}{*}{49} & \multirow[c]{2}{*}{81} \\

      {\#instances} & {AltGame} & { } & { } & { } & { } & { } & { } & { } & { } \\

      \midrule
      \multirow[c]{3}{*}{single} & {remove*} & {26.66\%} & {10.28\%} & {\phantom{-}7.13\%} & {\phantom{-}8.86\%} & {12.47\%} & {\phantom{-}7.78\%} & {\phantom{-}2.30\%} & {\phantom{-}1.89\%} \\
      {} & {sharpen} & {-1.00\%} & {-0.64\%} & {-2.70\%} & {-10.64\%} & {\phantom{-}1.63\%} & {\phantom{-}1.58\%} & {\phantom{-}0.75\%} & {\phantom{-}1.04\%} \\
      {} & {pointing} & {\phantom{-}0.00\%} & {\phantom{-}0.00\%} & {\phantom{-}0.00\%} & {\phantom{-}0.00\%} & {\phantom{-}6.36\%} & {\phantom{-}0.53\%} & {-6.67\%} & {-4.23\%} \\
      \midrule
      \multirow[c]{3}{*}{multiple} & {remove*} & {-68.73\%} & {\phantom{-}1.54\%} & {-35.87\%} & {13.07\%} & {14.49\%} & {\phantom{-}9.30\%} & {\phantom{-}4.37\%} & {\phantom{-}3.59\%} \\
      {} & {sharpen} & {\phantom{-}2.09\%} & {\phantom{-}2.54\%} & {-0.10\%} & {\phantom{-}1.75\%} & {\phantom{-}2.35\%} & {\phantom{-}1.72\%} & {\phantom{-}0.73\%} & {\phantom{-}0.45\%} \\
      {} & {pointing} & {\phantom{-}0.00\%} & {\phantom{-}0.00\%} & {\phantom{-}0.00\%} & {\phantom{-}0.00\%} & {\phantom{-}6.84\%} & {\phantom{-}8.01\%} & {-2.00\%} & {\phantom{-}1.30\%} \\
      \bottomrule

    \end{tabular}
  }
\end{table}

%% file: tbl/eval_generic/3way_rdiff_of_avg_fewmany.tex
\begin{table}[h]
  \caption[Comparison of results between all 3 experiments]{
    Side-by-side comparison of metrics rel$\Delta$(avg(VRISE), avg(RISE)) for all 3 experiments.
    Remove game scores are negated to match the "higher is better" convention.
    No data for GS in the "few" category because its multi-instance example contains more than 4 object instances.
    $\texttt{polygons}=49; blur \sigma=9.0; p_1=0.5; \texttt{meshcount}=1000; N_{masks}=max$
  }
  \label{tbl:eval.3way_fewmany}
  \begin{tabular}{cc|ccc}
    \toprule
    \multicolumn{2}{c|}{experiment >} & \multirow{2}{*}{GS} & \multirow{2}{*}{F$\sigma$} & \multirow{2}{*}{ImgNet} \\
    {\#instances} & {game} & {} & {} & {} \\

    \midrule
    \multirow{3}{*}{single} & {remove*} & {\phantom{-}7.13\%} & {\phantom{-}2.30\%} & {\phantom{-}6.88\%} \\
    {} & {sharpen} & {-2.70\%} & {\phantom{-}0.75\%} & {\phantom{-}1.09\%} \\
    {} & {pointing} & {\phantom{-}0.00\%} & {-6.67\%} & {-1.47\%} \\
    \midrule
    \multirow{3}{*}{few} & {remove*} & {-} & {\phantom{-}9.50\%} & {\phantom{-}8.28\%} \\
    {} & {sharpen} & {-} & {\phantom{-}2.48\%} & {\phantom{-}1.72\%} \\
    {} & {pointing} & {-} & {-6.67\%} & {-2.62\%} \\
    \midrule
    \multirow{3}{*}{many} & {remove*} & {-35.87\%} & {\phantom{-}2.27\%} & {\phantom{-}7.78\%} \\
    {} & {sharpen} & {-0.10\%} & {\phantom{-}0.12\%} & {\phantom{-}1.34\%} \\
    {} & {pointing} & {\phantom{-}0.00\%} & {-0.09\%} & {-1.36\%} \\
    \midrule

    \multicolumn{2}{c|}{VRISE sample size} & {15} & {1000} & {150000} \\
    \multicolumn{2}{c|}{RISE sample size} & {51} & {1400} & {150000}  \\
    \bottomrule
  \end{tabular}
\end{table}

%% file: tex/fluff/summary.tex
Overall, the application of Voronoi occlusions has improved the accuracy of saliency maps generated using large occlusions, especially in cases where multiple class instances were present in the explained image. From the perspective of our original objectives, it is mostly successful.
However, for other parameter configurations, we have observed a tradeoff between accuracy measured by Alteration Game and Pointing Game ("Fixed $\sigma$" col. of tbl.~\ref{tbl:eval.2way}). A negligible decrease in consistency of results was also present in the majority of cases. (tbl.~\ref{tbl:targeted_s.rawscores.params}).
What we have not managed to confidently determine is whether VRISE-specific parameters can be configured in a way that causes VRISE saliency maps to improve accuracy on both accuracy metrics \textit{regardless} of the configuration of RISE-specific parameters.
Blur strength ($\sigma$) and the percentage of visible image area ($p_1$) were found to have the greatest influence on the quality of results. (sec.~\ref{sec:outro.params_vs_quality})

Furthermore, considering that our modified occlusion generation step is more computationally expensive than the original, the current implementation of our method is clearly not competitive against RISE in scenarios where speed is an important factor.

The second proposed improvement: mask informativeness guarantee (sec.~\ref{sec:guaranteed_grid_gens}), has achieved the goal of increasing the convergence rate for low-\texttt{polygon}, low-$p_1$ configurations.
This outcome has been observed both in synthetic examples in (sec.~\ref{sec:guaranteed_grid_gens}) as well as in the results of an unrelated experiment (fig.~\ref{fig:ssim_eval.gs_conv_rawrise}).
For the parameter configuration used in the latter case, the convergence rate of maps using masks with guarantee was $2.5\%$ greater on average than that of maps using the original occlusion selector.
Additionally, the proposed Hybrid approach was shown to successfully eliminate the issue of increased mask generation time in configurations where uninformative masks are highly unlikely, while providing all benefits of slower grid generators with an informativeness guarantee.

We will now briefly summarize the remaining results.

Our results gathered on a large dataset (sec.~\ref{sec:imgnet}) show that VRISE improves the accuracy of saliency maps by $5\%$ but only when measured by Alteration Game. Accuracy measured by Pointing Game was decreased by $2\%$ (tbl.~\ref{tbl:imgnet.by_instance.reldiff})
We also compared our solution to other algorithms \cite{lime, gradcam, slidingWindow}, using results reported by authors of RISE (tbl.~\ref{tbl:imgnet.all_methods}).
The results obtained for both ResNet50 \cite{resnet50} and VGG16 \cite{vgg16} were consistent.

Another experiment (sec.~\ref{sec:targeted_s}) evaluated performance on a few select configurations, while controlling for influence of blur on results by configuring both algorithms to use identical blur strength. Its results ("Fixed $\sigma$" col. of tbl.~\ref{tbl:eval.2way}) can be grouped into two categories:
\begin{itemize}
  \item Configurations with small occlusions - only Alteration Game scores have improved (+1.89\% AltGame, -2.9\% Pointing Game)
  \item Configurations with large occlusions - both metrics have improved (+6.4\% AltGame, +5.43\% Pointing Game).
\end{itemize}
Furthermore, if images contained multiple instances of the observed class the average improvement values were higher (+6.97\% AltGame, +7.43\% Pointing Game).
These results show that multi-mesh Voronoi occlusions have an advantage in configurations which use a small number of large occlusions and for input images containing multiple instances of a single class.

In section \ref{ch:grid_search}, we have explored the relationships between parameter configuration and quality of results for both algorithms. However, due to the size of parameter space and significant cost of these computations (mentioned in the introduction of Chapter~\ref{ch:evaluation}) we had to limit our dataset to only two images, expecting them to be representative of the single-instance and many-instances image subclasses. Although trends seen in aggregate results (fig.~\ref{fig:grid_search.rawscores_vs_N_better_subset}) were similar to those observed on a larger dataset ("Fixed $\sigma$" col. of tbl.~\ref{tbl:eval.2way}), a side-by-side comparison of results for matching configurations (tbl.~\ref{tbl:eval.2way}) has shown that the reliability of our observations might be low and an evaluation on a larger dataset is required. Some of our reported findings - such as the asymptotic relationship between the number of Voronoi meshes and the quality of results - can be used to limit the size of parameter space in future work.

The analysis of similarity metrics has shown that multi-mesh approach possesses result regularization properties matching, but not exceeding, those of random shifts used by RISE. (fig.~\ref{fig:ssim_eval.gs_consistency_better_subset})

%% file: tex/fluff/future_work.tex
\section{Future work}

We propose the following avenues of future work:

\begin{itemize}
  \item Leveraging the properties of Voronoi mesh in iterative approach to saliency attribution.
  \item Continuation of parameter space exploration on a larger dataset. Our sample dataset was very small (due to the size of parameter space) and the comparison with results of other evaluations has shown that the former might be dataset-specific.
  \item Evaluation of VRISE with restored random shifts (both single- and multi-mesh) aiming to determine to what extent can the mesh pool and the computational cost of the occlusion generation process be reduced without sacrificing result quality.
  \item Comparison between VRISE and a modified version of RISE with configurable blur strength. GridSearch has shown that decreased blur strength can significantly improve map quality, which raises the following question: to what extent could the performance of RISE grids be improved by only manipulating blur strength, and in which configurations would Voronoi meshes remain a better alternative. This study would help identify the extent of the parameter space where Voronoi meshes have an unique advantage, with greater confidence and precision.

  \item Verification of effects observed for high-$p_1$ maps (sec.~\ref{ch:saliency_misattribution}) if softmax layer is removed from the evaluated model. We suspect that the issues of high-$p_1$ saliency misattribution and prioritization of class-differentiating features over those salient but shared with other classes are caused by the influence of softmax on class confidence scores.

  \item Future research could also focus on improving map convergence rate and $\frac{\text{map quality}}{\text{mask count}}$ ratio, in order to make the occlusion-based methods more competitive against white-box explainability algorithms and therefore more viable in scenarios other than forensic analysis of black-box models.
\end{itemize}

%% file: tex/fluff/acknowledgements.tex
\section{Acknowledgements}

This work has been partially supported by Statutory Funds of Electronics, Telecommunications and Informatics Faculty, Gdansk University of Technology.

We thank dr inż. Paweł Rościszewski for taking an interest in this topic and his support over the course of this project.

%% file: appendix/B_gridsearch_map_samples.tex
\begin{landscape}
\chapter{Saliency map examples}
\label{appendix:vrise_map_examples}
\begin{figure}[H]
  \centering
  \includegraphics[width=0.97\textwidth]{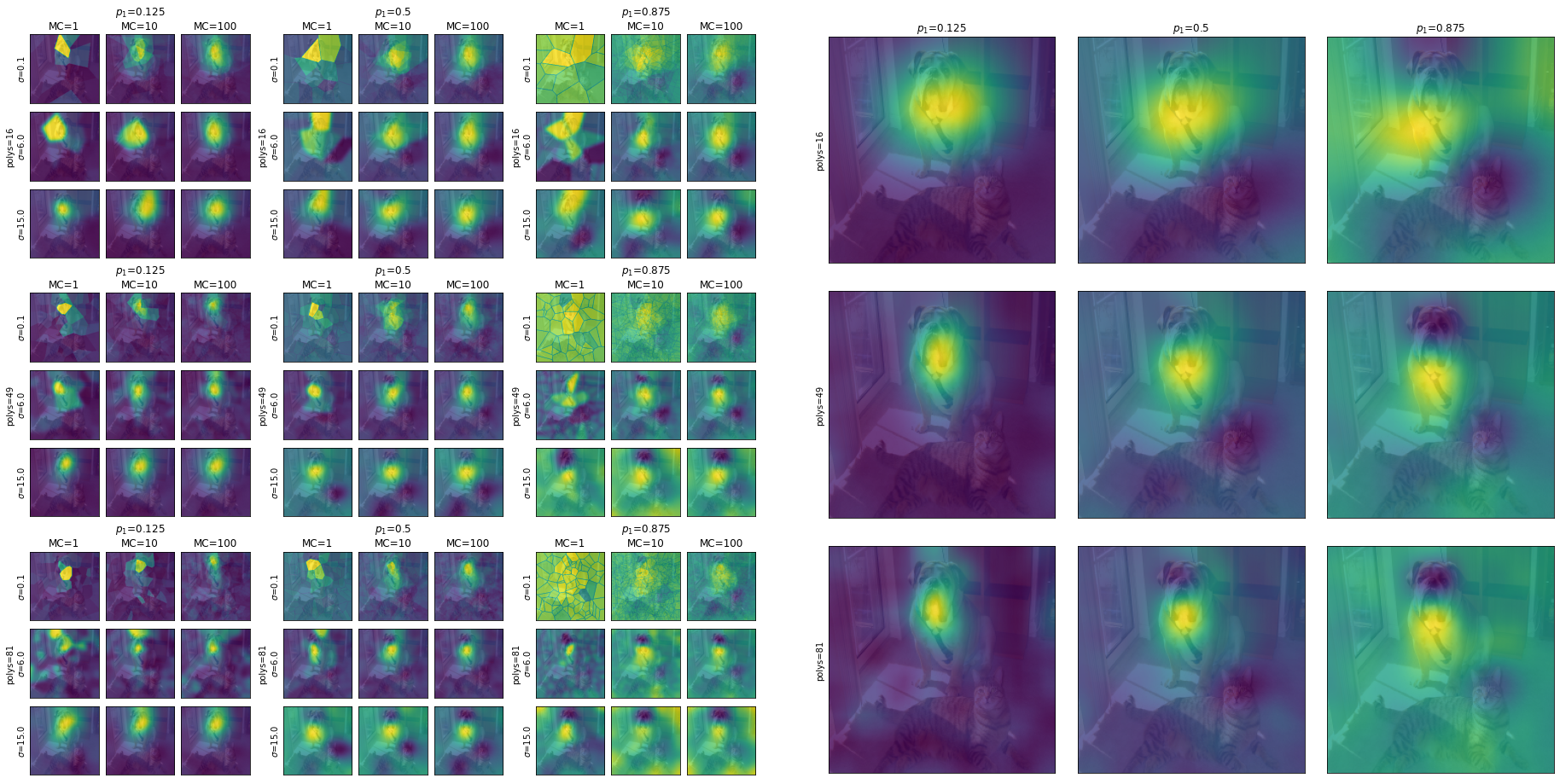}
  \caption[A demonstration of VRISE and RISE saliency maps for single-instance image]{A demonstration of VRISE and RISE saliency maps for single-instance image. 9x9 grid on the left contains VRISE maps, while the 3x3 grid on the right contains RISE maps. Each 3x3 sub-grid of VRISE maps corresponds to the similarly positioned RISE map in terms of parameters $p_1$ and \texttt{polygons}.}
  \label{fig:gs_appendix.catdog_demo}
\end{figure}
\begin{figure}[H]
  \centering
  \includegraphics[width=\textwidth]{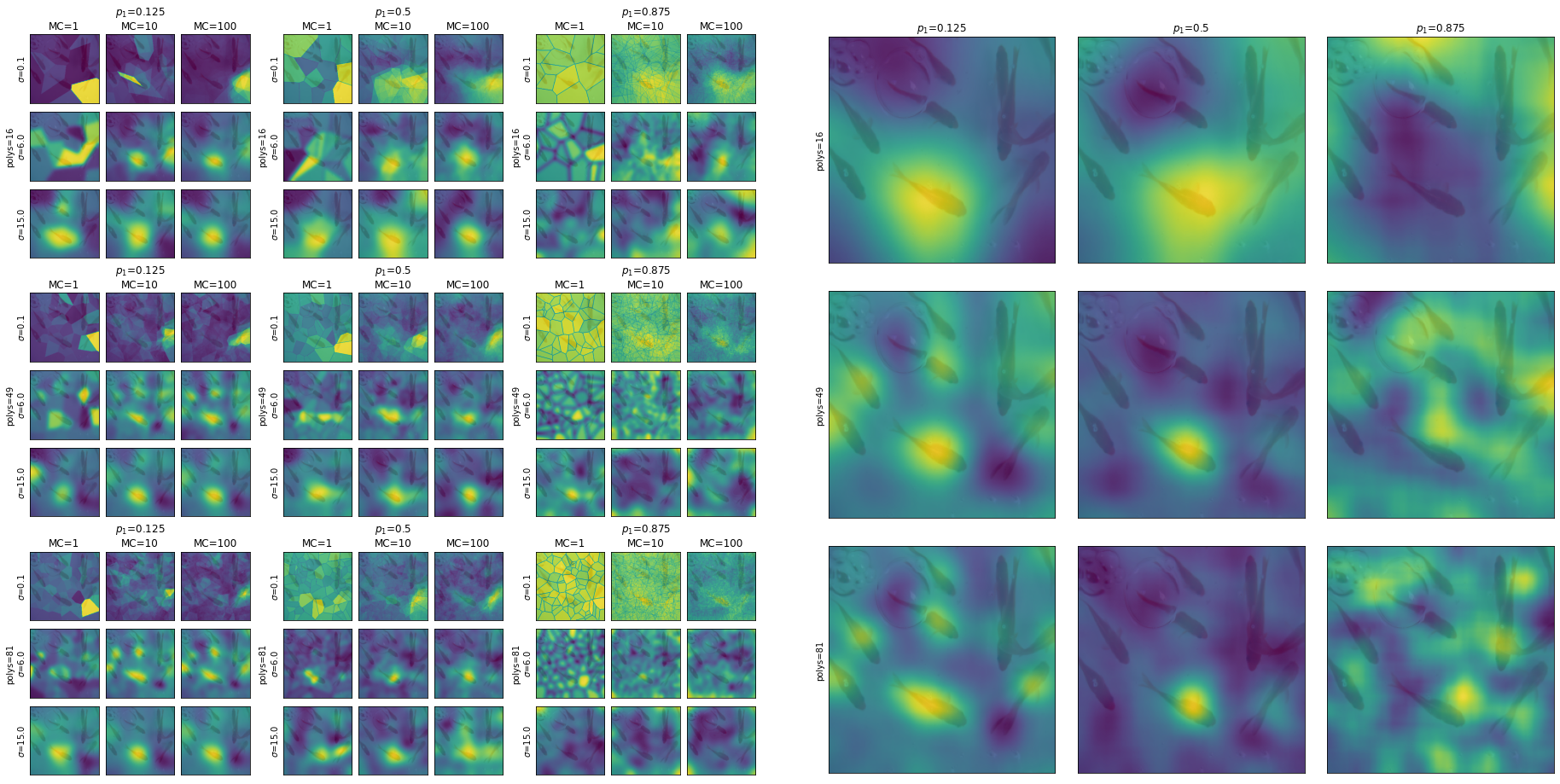}
  \caption[A demonstration of VRISE and RISE saliency maps for multi-instance image]{A demonstration of VRISE and RISE saliency maps for multi-instance image. 9x9 grid on the left contains VRISE maps, while the 3x3 grid on the right contains RISE maps. Each 3x3 sub-grid of VRISE maps corresponds to the similarly positioned RISE map in terms of parameters $p_1$ and \texttt{polygons}.}
  \label{fig:gs_appendix.goldfish_demo}
\end{figure}
\end{landscape}
\begin{figure}[H]
  \centering
  \includegraphics[width=0.87\textwidth]{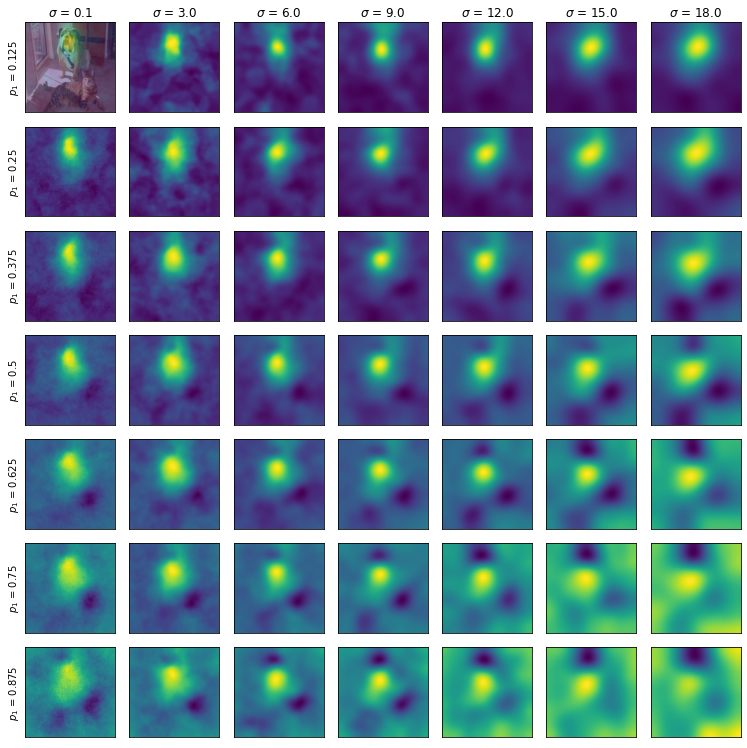}
  \caption[A demonstration of VRISE SMaps sampled across the entire spectrum of $p_1$ and blur $\sigma$ values]{A demonstration of VRISE SMaps sampled across the entire spectrum of $p_1$ and blur $\sigma$ values. Brightness indicates saliency. The map in upper left corner contains the input image, as reference. $\texttt{meshcount} = 500$, $\texttt{polygons} = 49$}
  \label{fig:gs_appendix.catdog_demo_p1_sigma}
\end{figure}
\begin{figure}[H]
  \centering
  \includegraphics[width=0.87\textwidth]{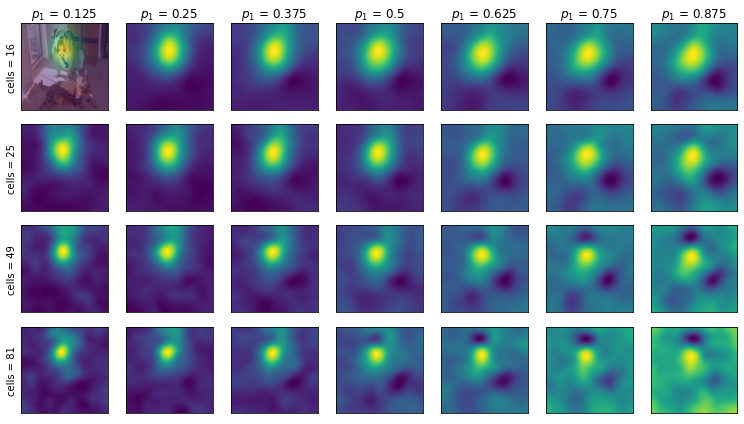}
  \caption[A demonstration of VRISE SMaps sampled across the entire spectrum of $p_1$ and \texttt{polygons} values]{A demonstration of VRISE saliency maps sampled across the entire spectrum of $p_1$ and \texttt{polygons} values. The map in upper left corner contains the input image, as reference. $\texttt{meshcount} = 500$, $blur \sigma = 9$}
  \label{fig:gs_appendix.catdog_demo_p1_polygons}
\end{figure}
\begin{figure}[H]
  \centering
  \includegraphics[width=0.87\textwidth]{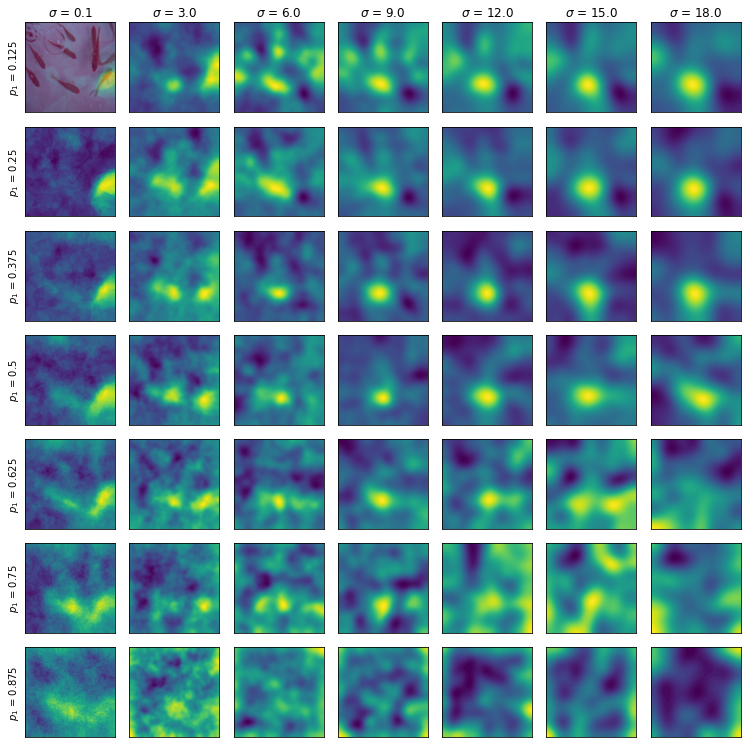}
  \caption[A demonstration of VRISE SMaps sampled across the entire spectrum of $p_1$ and blur $\sigma$ values]{A demonstration of VRISE SMaps sampled across the entire spectrum of $p_1$ and blur $\sigma$ values. Brightness indicates saliency. The map in upper left corner contains the input image, as reference. $\texttt{meshcount} = 500$, $\texttt{polygons} = 49$}
  \label{fig:gs_appendix.goldfish_demo_p1_sigma}
\end{figure}
\begin{figure}[H]
  \centering
  \includegraphics[width=0.87\textwidth]{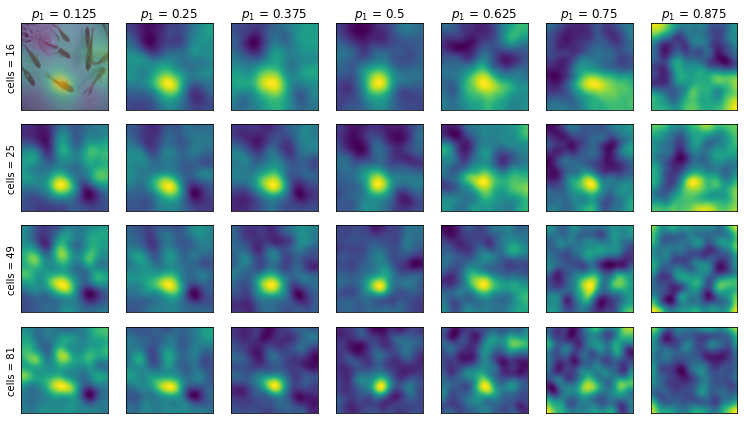}
  \caption[A demonstration of VRISE SMaps sampled across the entire spectrum of $p_1$ and \texttt{polygons} values]{A demonstration of VRISE saliency maps sampled across the entire spectrum of $p_1$ and \texttt{polygons} values. The map in upper left corner contains the input image, as reference. $\texttt{meshcount} = 500$, $blur \sigma = 9$}
  \label{fig:gs_appendix.goldfish_demo_p1_polygons}
\end{figure}

%% file: appendix/C_grid_search_heatmaps.tex
\chapter{Detailed breakdown of results of the "GridSearch" experiment}
\label{appendix:gridsearch.heatmaps}
\begin{figure}[H]
  \centering
  \includegraphics[width=0.95\textwidth]{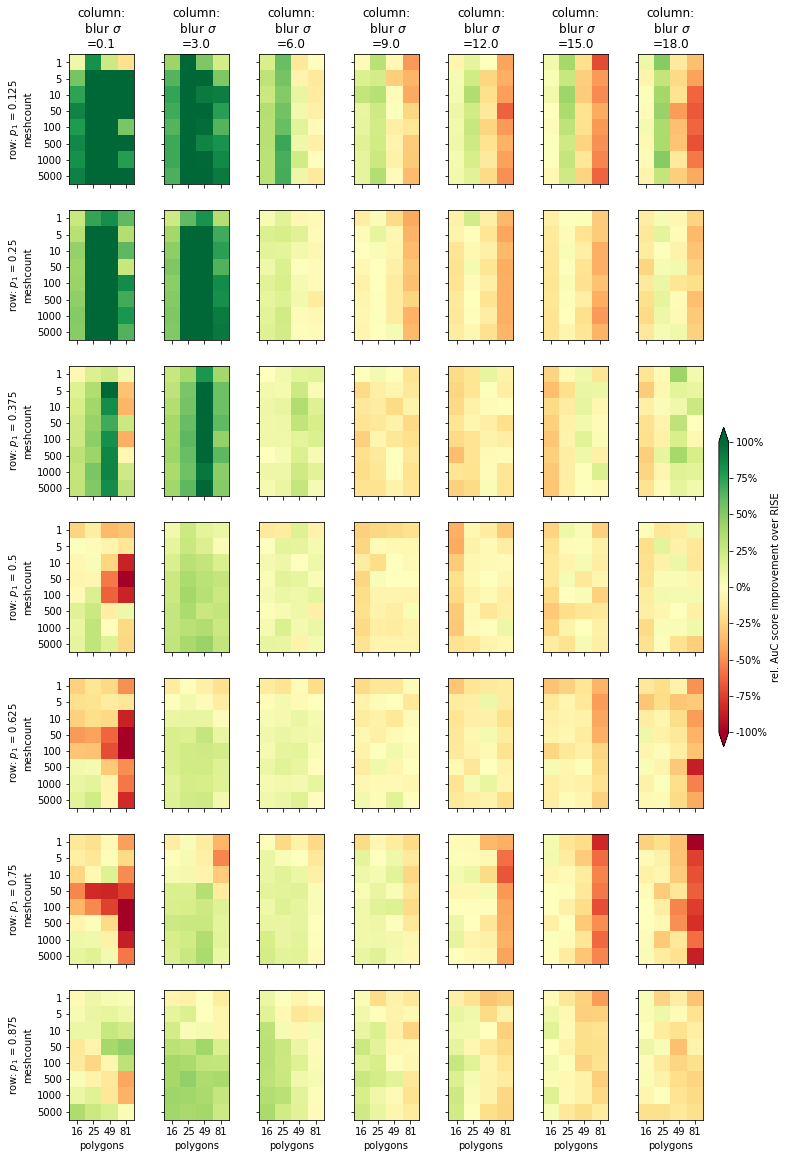}
  \caption[Heatmap of r$\Delta$(AuC) for each parameter set, averaged over all AltGame types (multi-instance case)]{Heatmap of r$\Delta$(AuC) for each parameter set, averaged over all AltGame types. 5k masks per map. \textbf{multi-instance} case.}
  \label{fig:grid_search.goldfish_aucdiff_heatmap}
\end{figure}
\begin{figure}[H]
  \centering
  \includegraphics[width=\textwidth]{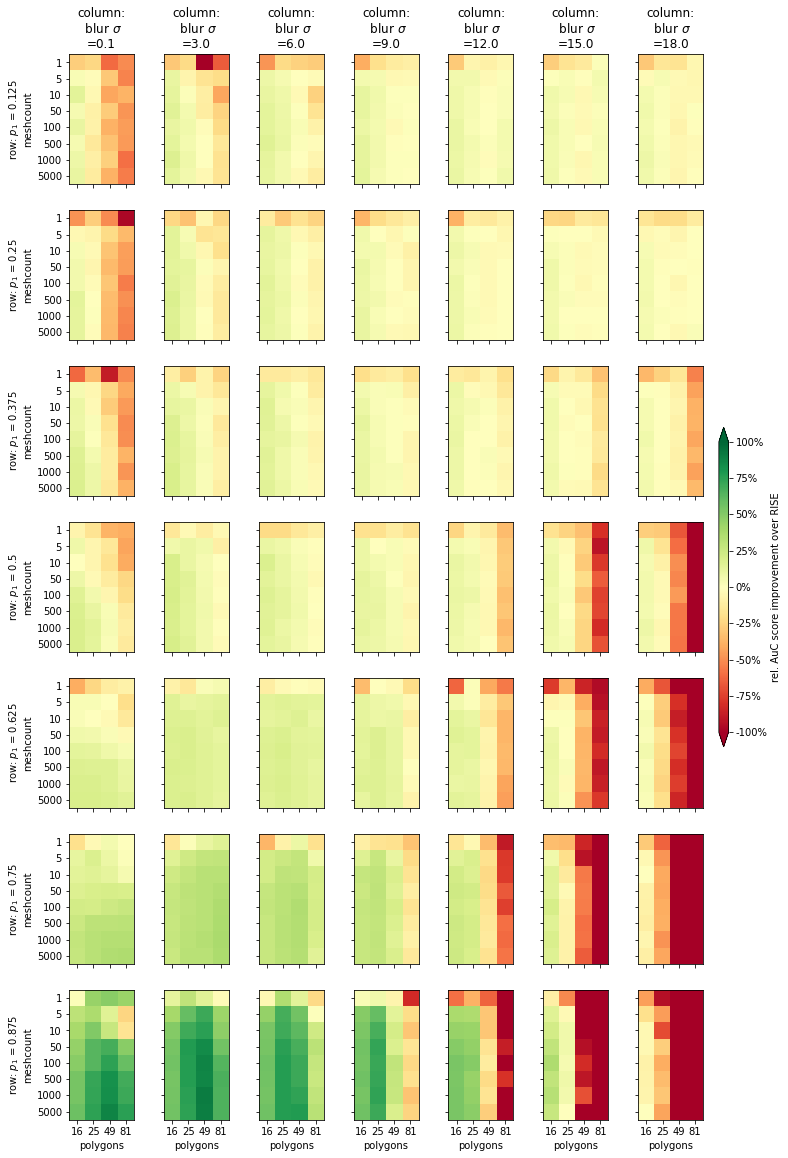}
  \caption[Heatmap of r$\Delta$(AuC) for each parameter set, averaged over all AltGame types (single-instance case)]{Heatmap of r$\Delta$(AuC) for each parameter set, averaged over all AltGame types. 5k masks per map. \textbf{single-instance} case.}
  \label{fig:grid_search.catdog_aucdiff_heatmap}
\end{figure}
\begin{figure}[H]
  \centering
  \includegraphics[width=\textwidth]{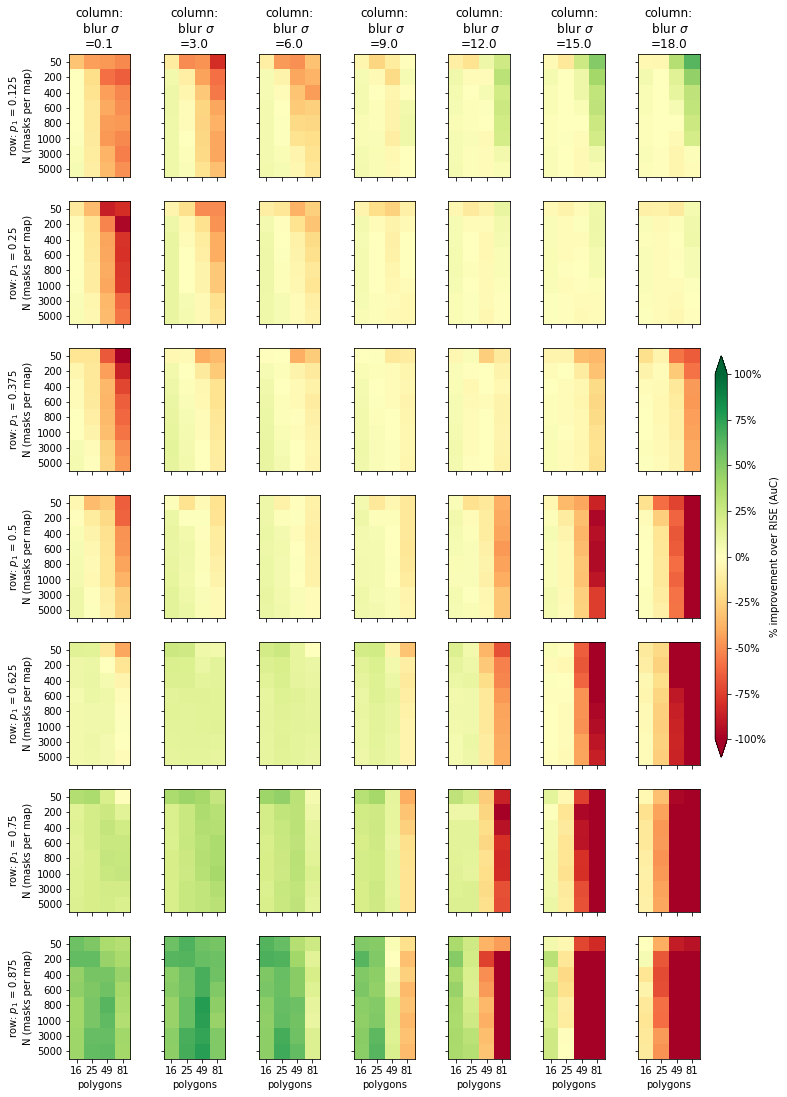}
  \caption[Heatmap of r$\Delta$(AuC) for values of $N_{mask}$, averaged over all AltGame types and $\texttt{meshcount}$ (single-instance case)]{Heatmap of r$\Delta$(AuC) for each parameter set, averaged over all AltGame types and $\texttt{meshcount}$. 5k masks per map. \textbf{single-instance} case.}
  \label{fig:grid_search.catdog_aucdiff_heatmap_checkpoints}
\end{figure}
\begin{figure}[H]
  \centering
  \includegraphics[width=\textwidth]{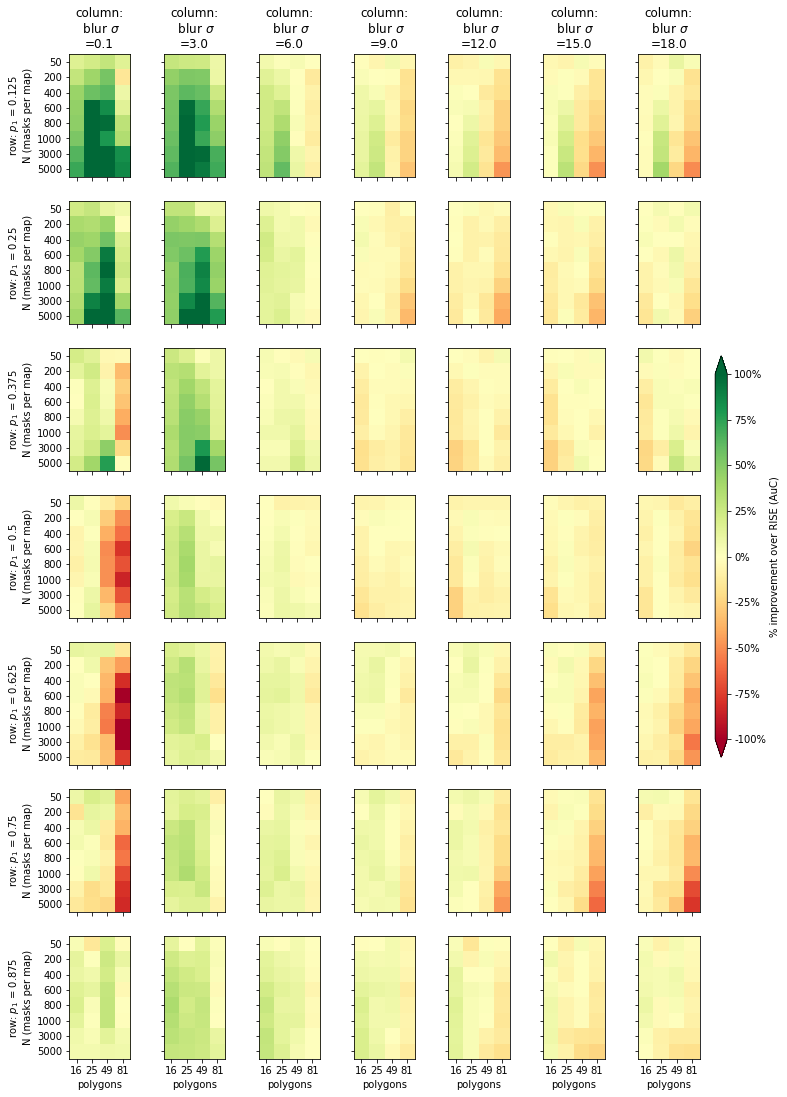}
  \caption[Heatmap of r$\Delta$(AuC) for various values of $N_{mask}$, averaged over all AltGame types and $\texttt{meshcount}$ (multi-instance case)]{Heatmap of r$\Delta$(AuC) for each parameter set, averaged over all AltGame types and $\texttt{meshcount}$. 5k masks per map. \textbf{multi-instance} case.}
  \label{fig:grid_search.goldfish_aucdiff_heatmap_checkpoints}
\end{figure}

%% file: appendix/D_grid_search_lineplots.tex
\chapter{Partially aggregated results used in the "GridSearch" experiment}
\label{appendix:gridsearch.lineplots}

\begin{figure}[H]
  \centering
  \includegraphics[width=0.76\textwidth]{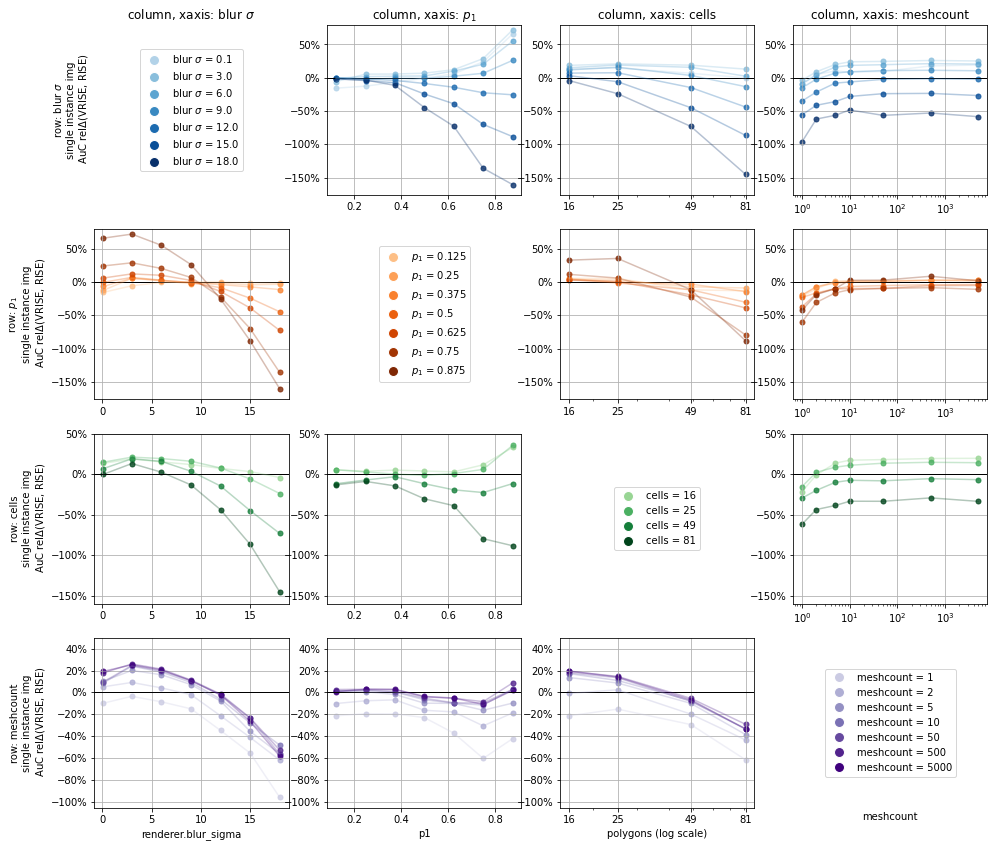}
  \caption[Relationships between pairs of parameters and rel$\Delta$(AuC) - single-instance case]{Relationships between pairs of parameters and rel$\Delta$(AuC) - \textbf{single-instance case}. Scores averaged over all game types, for maps in their final state.}
  \label{fig:grid_search.catdog_rauc_grid}
\end{figure}

\begin{figure}[H]
  \centering
  \includegraphics[width=0.76\textwidth]{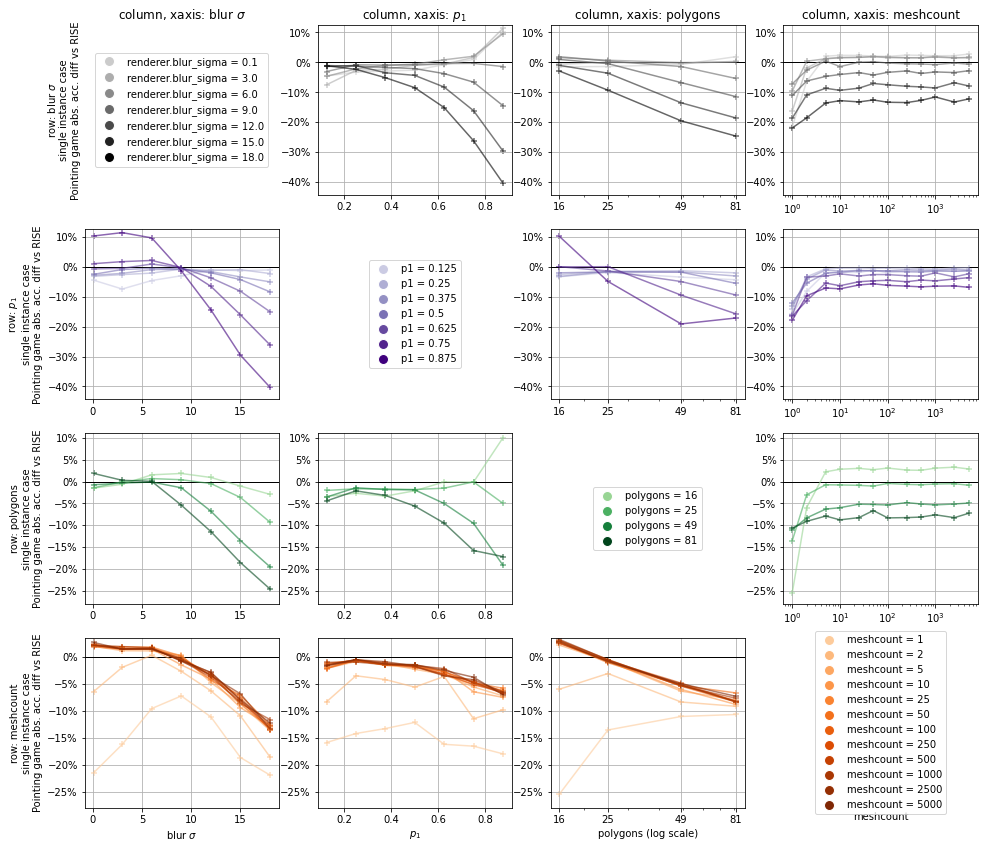}
  \caption[Abs$\Delta$ of Pointing Game accuracy for pairs of VRISE parameter values - single-instance case]{Abs$\Delta$ of Pointing Game accuracy for pairs of VRISE parameter values.  \textbf{single-instance} case.}
  \label{fig:grid_search.catdog_pg_grid}
\end{figure}

\begin{figure}[H]
  \centering
  \includegraphics[width=0.83\textwidth]{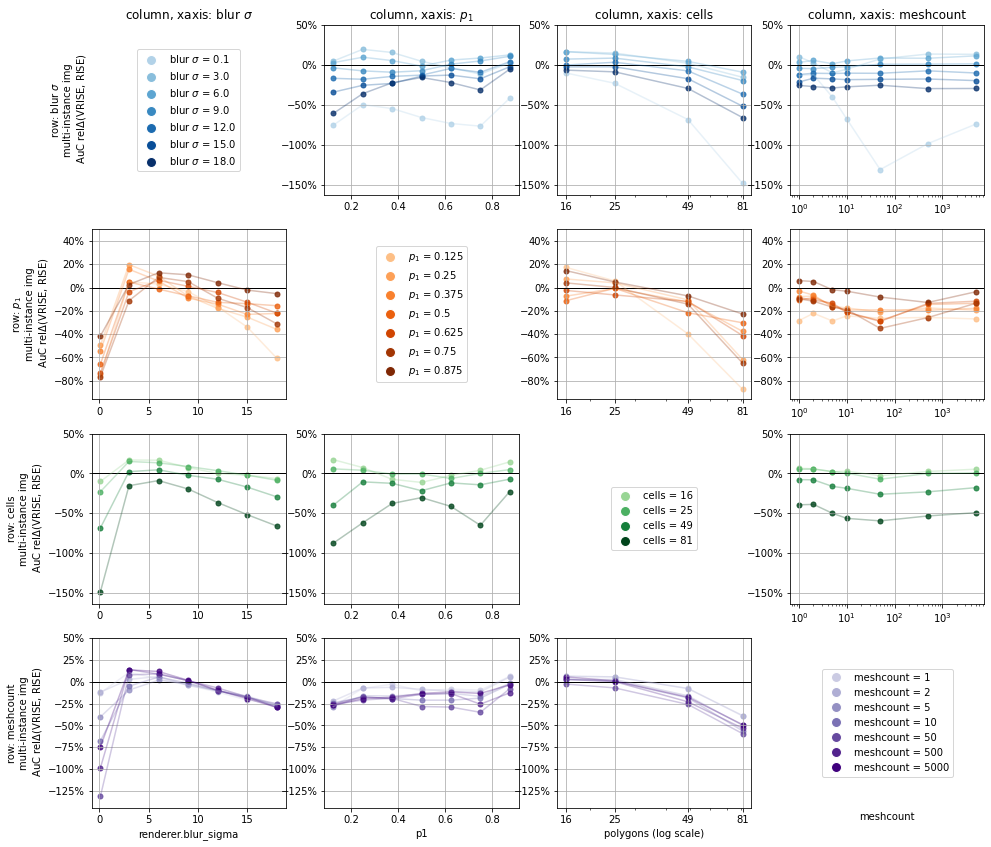}
  \caption[Relationships between pairs of parameters and rel$\Delta$(AuC) - multi-instance case]{Relationships between pairs of parameters and rel$\Delta$(AuC) - \textbf{multi-instance case}. Scores averaged over all game types, for maps in their final state.}
  \label{fig:grid_search.goldfish_rauc_grid}
\end{figure}

\begin{figure}[H]
  \centering
  \includegraphics[width=0.83\textwidth]{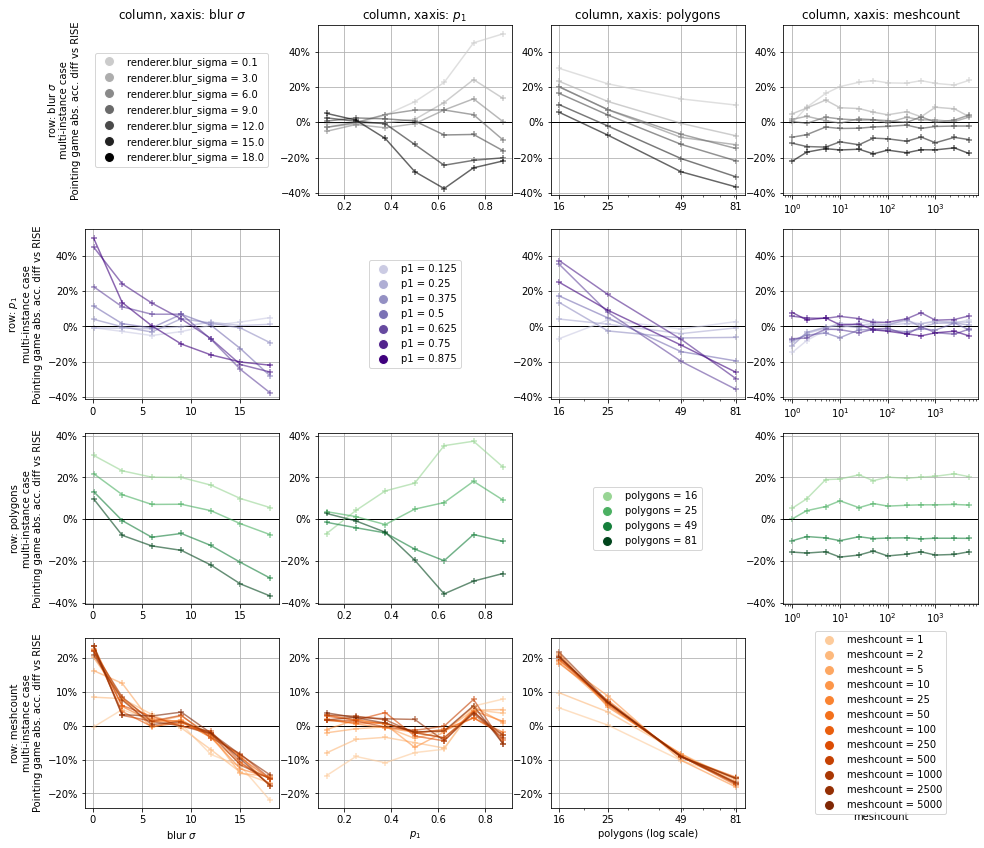}
  \caption[Abs$\Delta$ of Pointing Game accuracy for pairs of VRISE parameter values - multi-instance case]{Abs$\Delta$ of Pointing Game accuracy for pairs of VRISE parameter values. \textbf{multi-instance} case.}
  \label{fig:grid_search.goldfish_pg_grid}
\end{figure}